\documentclass[sigconf]{acmart}

\AtBeginDocument{%
  \providecommand\BibTeX{{%
    \normalfont B\kern-0.5em{\scshape i\kern-0.25em b}\kern-0.8em\TeX}}}

\setcopyright{acmlicensed}
\copyrightyear{2024}
\acmYear{2024}
\acmDOI{10.1145/3637528.3671551}

\acmConference[KDD '24]{Proceedings of the 30th ACM SIGKDD Conference on Knowledge Discovery and Data Mining}{August 25--29, 2024}{Barcelona, Spain}
\acmBooktitle{Proceedings of the 30th ACM SIGKDD Conference on Knowledge Discovery and Data Mining (KDD '24), August 25--29, 2024, Barcelona, Spain}
\acmISBN{979-8-4007-0490-1/24/08}

\usepackage{amsmath}
\DeclareMathOperator*{\argmin}{arg\,min} 

\usepackage{xspace}

\usepackage{booktabs}
\usepackage{longtable}
\usepackage{multirow}

\newcommand{\framework}{\textbf{\mbox{CG-CT}}\xspace}
\newcommand{\frameworklong}{counterfactual generator for continuous treatments\xspace}

\usepackage{hyperref}  
\hypersetup{
    colorlinks = true,
    citecolor  = blue,
    linkcolor  = blue
    }
    
\usepackage{enumerate}
\usepackage{enumitem}
\usepackage{float}

\usepackage{graphicx}
\usepackage{caption}
\usepackage{subcaption}
\usepackage{float}
\usepackage[percent]{overpic}

\newcommand{\figletter}[1]{{{\fontfamily{\sfdefault}\selectfont \textbf{(#1)}}}}

\usepackage{rotating}
\usepackage{tabularx}
\usepackage{dcolumn}
\usepackage{pdflscape}
\usepackage{rotating}

\usepackage{array}

\usepackage{algpseudocode}

\usepackage{algorithm}

\begin{document}

\title{Causal Machine Learning for Cost-Effective Allocation of Development Aid}

\author{Milan Kuzmanovic}
\email{mkuzma96@gmail.com}
\affiliation{
  \institution{ETH Zurich}
  \city{Zurich}
  \country{Switzerland}
}

\author{Dennis Frauen}
\email{frauen@lmu.de}
\affiliation{
  \institution{Munich Center for Machine Learning \& LMU Munich}
  \city{Munich}
  \country{Germany}
}

\author{Tobias Hatt}
\email{hatt.tob@gmail.com}
\affiliation{
  \institution{ETH Zurich}
  \city{Zurich}
  \country{Switzerland}
}

\author{Stefan Feuerriegel}
\email{feuerriegel@lmu.de}
\affiliation{
  \institution{Munich Center for Machine Learning \& LMU Munich}
  \city{Munich}
  \country{Germany}
}

\renewcommand{\shortauthors}{Kuzmanovic, et al.}

\begin{abstract}
The Sustainable Development Goals (SDGs) of the United Nations provide a blueprint of a better future by ``leaving no one behind'', and, to achieve the SDGs by 2030, poor countries require immense volumes of development aid. In this paper, we develop a causal machine learning framework for predicting heterogeneous treatment effects of aid disbursements to inform effective aid allocation. Specifically, our framework comprises three components: (i)~a balancing autoencoder that uses representation learning to embed high-dimensional country characteristics while addressing treatment selection bias; (ii)~a counterfactual generator to compute counterfactual outcomes for varying aid volumes to address small sample-size settings; and (iii)~an inference model that is used to predict heterogeneous treatment--response curves. We demonstrate the effectiveness of our framework using data with official development aid earmarked to end HIV/AIDS in 105 countries, amounting to more than USD 5.2 billion. For this, we first show that our framework successfully computes heterogeneous treatment--response curves using semi-synthetic data. Then, we demonstrate our framework using real-world HIV data. Our framework points to large opportunities for a more effective aid allocation, suggesting that the total number of new HIV infections could be reduced by up to 3.3\% ($\sim$50,000 cases) compared to the current allocation practice.
\end{abstract}




\maketitle

\section{Introduction}


The Sustainable Development Goals (SDGs) by the United Nations define a global framework to achieve a better and more sustainable future for the people and the planet \cite{UNsdg2015}. The SDGs list various targets that should be met by 2030 (e.g., ending the HIV/AIDS epidemic, eliminating hunger). Here, a central principle of the SDGs is that of \emph{``leaving no one behind''}: it is the unequivocal aim of the United Nations to improve conditions for all parts of society and across all countries, including developing countries \cite{UNreport2018}.


An important driver of progress towards the SDGs, especially for developing countries, is the provision of development aid \cite{OECDoda2018, UNsdg2014}. Development aid, also called official development assistance~(ODA), supports developing countries through various activities funded by donor institutions (e.g., foreign governments, national development agencies, development banks, philanthropic organizations). ODA pools immense financial resources from donor institutions worldwide. For instance, in 2021, ODA amounted to USD 178.9 billion \cite{OECDoda2021}. However, current aid allocation practices mostly rely on human judgment and decision heuristics \cite{OECD2009, OECDsdglab2022, Toetzke2022}, and can thus result in inefficient allocation \cite{GB2020}. Given that ODA budgets are limited, as well as current challenges (e.g., the Russian invasion of Ukraine, climate change, etc.), a cost-efficient use of development aid is of great need.


In this paper, we aim to predict the effectiveness of development aid, so that practitioners can then identify cost-effective allocations. For this, we develop a novel, causal machine learning framework for predicting the effect of aid disbursements on SDG outcomes called \framework (\emph{\frameworklong}). Specifically, we predict heterogeneous treatment effects of aid disbursements as a continuous treatment while additionally controlling for various confounders. Our framework is different from existing methods and carefully tailored to our setting. Specifically, our \framework is designed to simultaneously address continuous treatments, high-dimensional covariates (country characteristics), treatment selection bias, and small sample-size settings. This allows us to predict the heterogeneous aid effect across different countries, that is, individualized \emph{treatment--response curves}. In our setting, we also refer to the latter as \emph{aid--response curves}. Further, our framework is data-efficient: the different components operate on a low-dimensional covariate space and polynomial regression together with data augmentation. This is a crucial difference of our framework from common baselines that build upon neural networks (e.g., \cite{Bica2020,Schwab2020}).


We show the effectiveness of our \framework in various experiments. For this, we use data from a collaboration with the \emph{Organisation for Economic Co-operation and Development (OECD)}, an intergovernmental organization tasked with monitoring and coordinating the development aid activities of different donors. For our experiments, we focus on official development aid earmarked to end HIV/AIDS in 105 countries. This choice is motivated by the fact that (i)~the HIV epidemic is a major cause of death \cite{UNhiv2020, WHOhiv2020, UNAIDShiv2022}, and (ii)~ODA committed to ending HIV/AIDS pools extensive resources (e.g., more than USD 106 billion since 2002)  \cite{OECDsdgdata2022}. 
The results demonstrate the potential of our machine learning framework to inform more effective aid allocations for achieving progress towards the SDGs. 

Our main \textbf{contributions} are as follows:\footnote{Data and code that supports the findings of our study is made publicly available via a GitHub repository at: \url{https://github.com/mkuzma96/CG-CT}}
\begin{enumerate}[leftmargin=0.6cm] 
\item We study the problem of predicting heterogeneous treatment effects of development aid on SDG outcomes, thereby informing the effective allocation of development aid. 
\item We develop a novel causal machine learning framework (\framework) for this purpose. We further show that state-of-the-art methods are outperformed by our \framework.  
\item We demonstrate our \framework framework using real-world data from the HIV epidemic, where it pinpoints a large potential for reducing the number of new HIV infections over current practice. 
\end{enumerate}

\section{Related Work}

\textbf{Development aid:}
To inform aid allocations, the effectiveness of aid has been subject to extensive analyses (see supplements in our GitHub repository for details). One literature stream builds upon experiments (e.g., \cite{Duflo2003, Duflo2006}). However, experiments involving development aid are costly and limited to micro-level analyses (e.g., a small geographic area and not across multiple countries). A different literature stream uses observational data (e.g., \cite{Birchler2016, Munyanyi2020}), yet these works make strong modeling assumptions (e.g., the treatment effect is linear in the aid volume, the effectiveness of aid cannot vary across countries but the effectiveness is instead identical for all countries, etc.). Moreover, the true treatment effect is typically underestimated due to the nature of the modeling approach \cite{Birchler2016, Munyanyi2020}, and, therefore, estimates can merely serve as a lower bound. Importantly, all of the above works estimate the \emph{average} effect of aid across countries but without considering the \emph{heterogeneity} in the aid effect across countries. However, to inform aid allocations of decision-makers, methods are needed to predict the \emph{heterogeneous} treatment effect of aid for a specific country of interest.  


To predict the treatment effect of aid under between-country heterogeneity, existing research has also applied standard machine learning methods (e.g., LASSO regression \cite{Jakubik2022}). Here, the benefit of machine learning is that it should generalize well across multiple countries and thus also allow for accurate predictions of treatment effects for out-of-sample data points. Yet, standard machine learning methods can give unreliable predictions of treatment effects in the presence of treatment selection bias \cite{Shalit2017}. For example, if wealthy countries generally receive low volumes of development aid, predicting the treatment effect of large aid volumes for wealthy countries might be unreliable due to the limited number of data points. Hence, treatment selection bias can create covariate shifts in the observational data that preclude reliable estimation of treatment effects in some covariate domains. Such covariate shifts are also present in our data around development aid and macroeconomic variables (see supplements in our GitHub repository for an analysis of aid-dependent covariate shifts). To address this issue, we develop a tailored causal machine learning framework for predicting heterogeneous treatment effects of development aid under treatment selection bias.   

\textbf{Causal machine learning:}
Recent advances in machine learning allow for predicting heterogeneous treatment effects from observational data \cite{Alaa2018, Feuerriegel2024}. This has, among others, enabled new applications in health (e.g., \cite{Liu2021,Kraus2023}). Here, the prime focus is to reliably predict treatment effects despite different challenges (e.g., confounding bias, treatment selection bias). This is also the reason why na{\"i}ve methods (e.g., standard machine learning models) for predicting treatment effects can be unreliable \cite{Shalit2017}. 

As an example, let us assume that economic activity positively affects an SDG outcome, but negatively affects the volume of received aid. Then, failing to control for GDP (gross domestic product) per capita when predicting the effect of aid on the SDG outcome could lead to the wrong conclusion that a smaller volume of aid improves the SDG outcome (an instance of confounding bias). Moreover, even if we control for GDP per capita, observational data is likely to have a limited number of observations of large aid volumes in wealthy countries, thus precluding reliable estimation of treatment effects of large aid volumes for wealthy countries (an instance of treatment selection bias). To this end, it is crucial to address these problems when predicting treatment effects. 

To predict heterogeneous treatment effects, many custom methods based on machine learning have been proposed for binary treatments (such as, e.g., causal forests) \cite{Johansson2016, Shalit2017, Yoon2018, Athey2019, Hatt2021}. In contrast, we are interested in predicting the effect of continuous treatments, which is also known as predicting a heterogeneous dose-response curve (in our setting, we call it later treatment-response curve or aid-response curve). However, there are only a few works for continuous treatments and heterogeneous treatment effects (e.g., generalized propensity score (GPS), DRNet, SCIGAN) \cite{Hirano2004, Schwab2020, Bica2020}. Importantly, all of these methods have caveats when simultaneously addressing continuous treatments, high-dimensional covariates, treatment selection bias, and small sample-size settings (see the overview in supplements in our GitHub repository). 

\section{Methods}
\label{sec:methods}

\subsection{Problem setup}


We consider the following cross-sectional setting with $i = 1, \ldots, n$ countries. (1)~Let  $Y_i \in \mathcal{Y} \subseteq \mathbb{R}$ denote the outcome in country $i$. In our case, this is the relative reduction in the HIV infection rate compared to the previous year. However, without loss of generality, one could also use our framework for other SDG outcomes (e.g., prevalence of undernourishment, human poverty indices, tuberculosis incidence). (2)~Let $A_i \in \mathcal{A} \subseteq \mathbb{R}_{+}$ denote the continuous treatment, which, in our case, is the volume of development aid that is allocated to a country. (3)~Let $X_i \in \mathcal{X} \subseteq \mathbb{R}^{p}$ refer to the additional covariates that capture country characteristics and act as potential confounders. To make inferences, we have access to observational data on past aid disbursements earmarked to the SDG outcome (here: HIV infection rate). We refer to the observational data by $\mathcal{D} =  \{(y_i, a_i, x_i)\}_{i=1}^{n}$. For brevity, we omit the index $i$ unless explicitly needed.

\begin{figure}[htb]
\centering   
\includegraphics[width=0.25\textwidth]{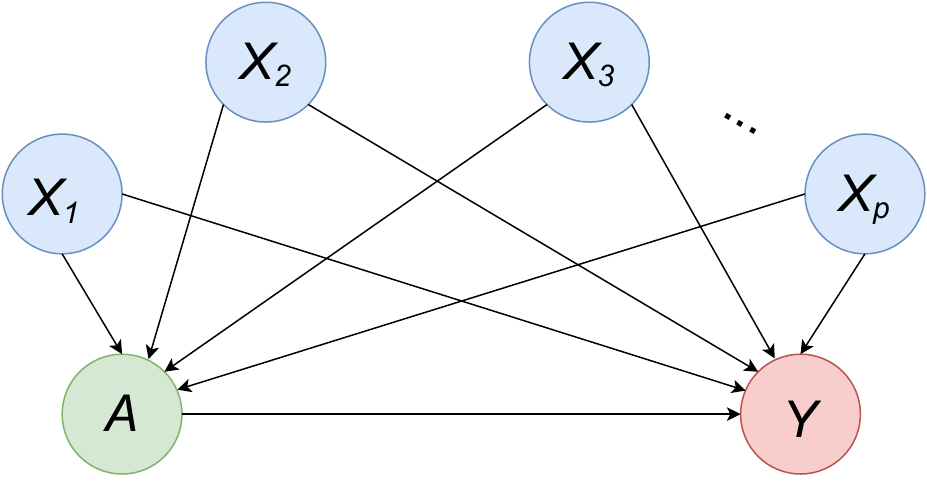}
\caption{\textbf{Causal structure of our problem setup. Our aim is to predict the effect of development aid ($A$) on the reduction in the HIV infection rate ($Y$), while controlling for various country characteristics such as socioeconomic, macroeconomic, and health-related covariates ($X_{1}, X_{2}, \ldots, X_{p}$) as potential confounders. We have a cross-sectional setting without time dependencies.}
}
\label{fig:cau_str}
\end{figure}

We formalize our task using the Rubin-Neyman potential outcomes framework \cite{Rubin2005}. Specifically, we define outcomes as follows. For every treatment $a \in \mathcal{A}$, we denote the potential outcome by $Y(a) \in \mathcal{Y}$. It gives the SDG outcome if, hypothetically, aid volume $a$ would have been allocated to a country. However, the fundamental problem of causal inference is that we observe only one potential outcome; that is, the observational data includes only one $Y(a)$ for a single value $A=a$. In terms of terminology, the observed outcome is also called the factual outcome, while all other unobserved potential outcomes are called counterfactual outcomes. Note that, in our setting with continuous treatments, we have infinitely many counterfactual outcomes. The causal structure of our problem setup is shown in Figure~\ref{fig:cau_str} (note that we do not assume that covariates are mutually independent, but rather use the illustration to show that covariates are potential confounders).


Our aim is to predict the heterogeneous treatment effect, i.e., the expected potential outcome for a given treatment value conditioned on covariates. Formally, the \emph{heterogeneous treatment effect} is defined by
\begin{equation}
\mu(a,x) := \mathbb{E} \, [ \, Y(a) \mid X = x \, ], 
\end{equation}
where $x$ is the given value for the covariates. We further introduce the so-called \emph{treatment--response curve}: it refers to the function $\mu(a,x)$ for varying volumes of development aid, $a$, while the country characteristics, $x$, are kept fixed. In our case, we also call it \emph{``aid--response curve''}. In order to unbiasedly estimate $\mu(a,x)$ from observational data $\mathcal{D}$, we make the following assumptions to ensure identifiability of the treatment effects.

\emph{Assumptions.} 
(i)~Consistency: {$Y = Y(a)$ if $A = a$};
(ii)~Positivity: {$0 < p(A = a \mid X = x) < 1$, $\forall a \in \mathcal{A}$, if $p(x) > 0 $};
(iii)~Ignorability: {$Y(a) \perp \!\!\! \perp A \mid X = x \, \, $, $\forall a \in \mathcal{A}$}.

The above assumptions are the standard for identifiability of treatment effects from observational data \cite{Pearl2009, Imbens2015} . Consistency from Assumption~(i) ensures that the observed outcomes are realizations of potential outcomes given observed treatment. Positivity from Assumption~(ii) ensures that all treatments are possible on the entire covariate space. Ignorability from Assumption~(iii) is often referred to as ``no hidden confounders'' assumption, meaning that all variables that affect both treatment $A$ and potential outcomes $Y(a)$ are measured in the covariates $X$. Under Assumptions (i)--(iii), we have that $\mu(a,x) := \mathbb{E} \, [ \, Y(a) \mid X = x \, ] = \mathbb{E} \, [ \, Y \mid A = a, X = x \, ]$. Hence, we can estimate $\mu(a,x)$ from observational data $\mathcal{D}$ by learning a function $f: \mathcal{A} \times \mathcal{X} \rightarrow \mathcal{Y}$ using machine learning. (In the supplements in our GitHub repository, we provide several secondary analyses where we provide empirical results to validate the above assumptions.) 

Our task is subject to three main challenges, which we briefly summarize in the following. (1)~There is a comparatively large set of country characteristics that we need to control for in order to adjust for confounding. However, as a result, having a high-dimensional covariate space leads to higher estimation variance (see Supplement~\ref{supp:control_vars} for an overview of all control variables). (2)~Treatment selection bias can be present in observational data, which means that treatment--response estimates in some covariate domains might be unreliable due to a limited number of observed data points (see supplements in our GitHub repository for an analysis of covariate distributions for different development aid volumes). (3)~There is only a comparatively small set of countries receiving aid because of which the sample size $n$ with observed outcomes is naturally small. To obtain reliable estimates despite the aforementioned challenges, we develop a tailored method in the following called \frameworklong (\framework).  

\subsection{Data} 




We use the following data: 
(1)~Outcome variable $Y$ represents the relative reduction in the HIV infection rate (i.e., the annual number of new infections per 1,000 uninfected in the population) compared to the previous year. The data are obtained from UN SDG indicator database \cite{UNsdgdata2022}.
(2)~Treatment $A$ represents the official development aid volume in USD millions per annum earmarked to end HIV in each country. The data are obtained from OECD CRS database \cite{OECDsdgdata2022}.
(3)~Covariates $X$ represent control variables that capture different country characteristics with regard to socioeconomic, macroeconomic, and health-related dimensions. Examples are: gross domestic product (GDP) per capita, GDP growth, population size, maternal mortality, infant mortality, school enrollment, etc. The complete list of covariates is given in Supplement~\ref{supp:control_vars}. The data are obtained from the World Bank database \cite{WBdata2022}.

We use the data from 105 countries that are recipients of development aid earmarked towards ending the HIV epidemic. The sample covers a population of more than 4.3 billion people (around half of the world's population). 
We predict the country-specific aid--response curves for the year 2017. The reason is that some of the data are made public by international organizations only with a time lag after the actual data collection (e.g., maternal mortality). For learning, we use data from the previous year, that is, 2016. This ensures that all performance evaluations are conducted using out-of-sample observations from 2017. Later, we also conduct robustness checks with other time frames (see supplements in our GitHub repository). 

\subsection{Causal machine learning framework}

To learn treatment--response curves for our HIV setting, we develop the following causal machine learning framework which we call \emph{\frameworklong}  (\framework). \framework is tailored so that it can handle continuous treatments and deal with the unique challenges of our problem setup (i.e., high-dimensional covariates in the form of country characteristics, treatment selection bias, small sample size). \framework comprises three components: (1)~a balancing autoencoder aimed at embedding high-dimensional country characteristics while simultaneously addressing treatment selection bias; (2)~a counterfactual generator to compute counterfactual outcomes for varying volumes of aid to address the small sample-size setting; and (3)~an inference model which is a (causal) machine learning model used to predict treatment--response curves from observational data. The idea behind components (1) and (2) is to perform a data augmentation that generates synthetic data points under our causal graph, which, in turn, helps in predicting the treatment--response curves in component~(3), despite the small sample-size setting. Below, we present the three components of our \framework (see Figure~\ref{fig:mthd_arch}). Pseudocode is in Supplement~\ref{supp:pseudocode}.

\begin{figure}[htb]
\centering   
\includegraphics[width=0.95\columnwidth]{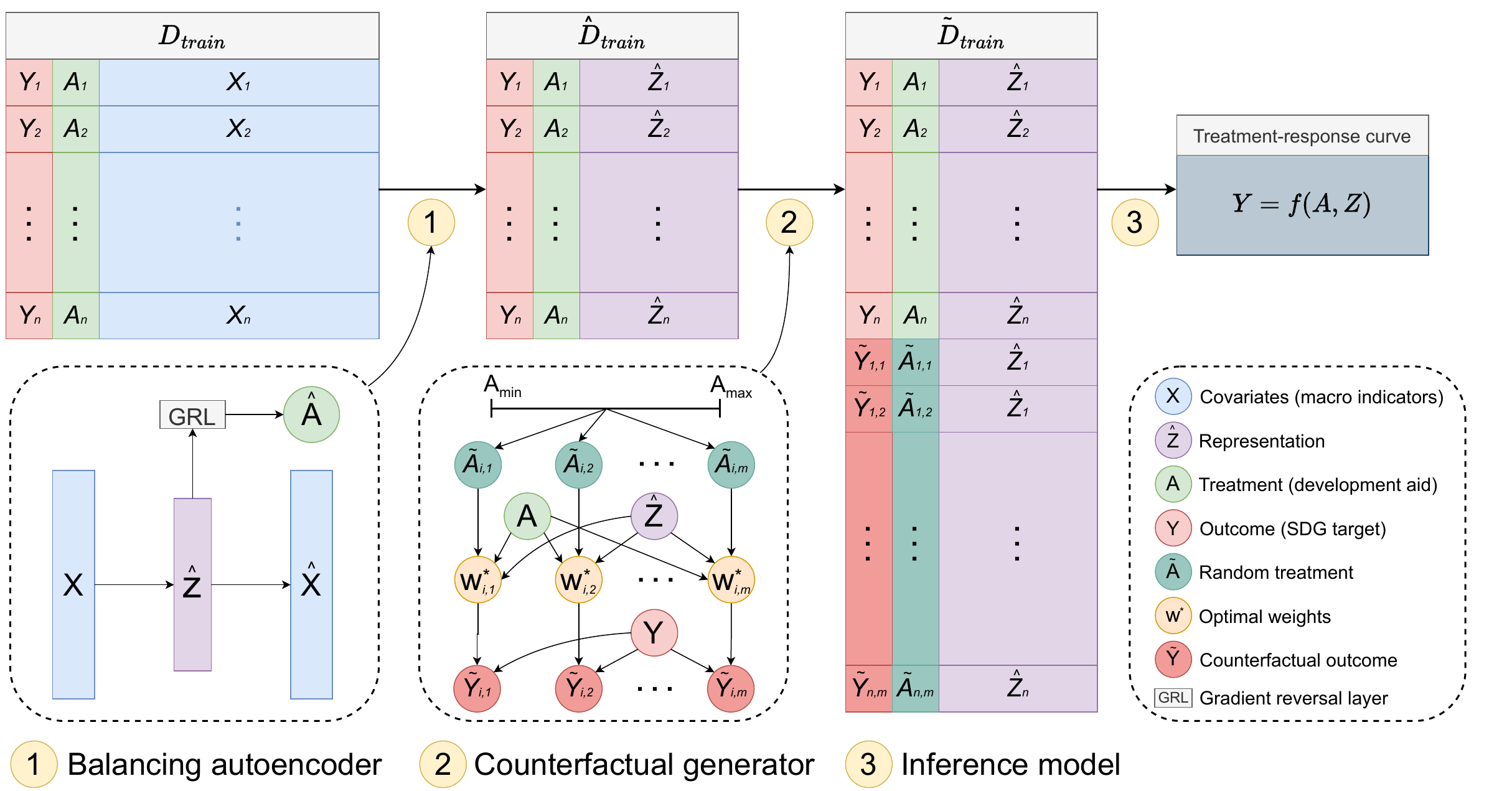}
\caption{\textbf{Overview of our machine learning framework. The aim of our \framework is to predict heterogeneous treatment effects of aid disbursements on SDG outcomes. For this, our \framework proceeds along three components. In the first component, country characteristics (e.g., socioeconomic, macroeconomic, or health-related controls) are embedded in a lower dimension using a balancing autoencoder. Here, we also address treatment selection bias by learning a representation of covariates that is not predictive of the treatment (implemented via gradient reversal layer). In the second component, counterfactual outcomes are generated for varying treatment values, and then combined with the observational data. In the third component, an inference model is used together with the previous data to learn a relationship between outcome ($Y$) and treatment ($A$) for a given representation of covariates ($Z$). This gives the treatment--response curve.}}
\label{fig:mthd_arch}
\end{figure}

\noindent
\underline{\emph{Component (1): Balancing autoencoder.}} The balancing autoencoder maps the original high-dimensional covariate space $\mathcal{X} \in \mathbb{R}^p$ onto a representation space $ \mathcal{Z} \in \mathbb{R}^r$ with $r < p$. The aim is to: (i)~reduce the dimension of the covariate space while preserving ignorability (i.e., confounder control), and (ii)~address treatment selection bias. For this, we use representation learning \cite{Bengio2013} to learn a representation $\phi: \mathcal{X} \rightarrow  \mathcal{Z}$  that preserves the original covariate information in a lower-dimensional space while being non-predictive of the treatment $A$. Therefore, we learn our balancing autoencoder by minimizing the loss function 
\begin{equation}
\label{eq:loss_bae}
\mathcal{L} = \underbrace{\frac{1}{n}  \sum_{i=1}^N \frac{1}{p} \big\| x_i - \hat{x}_i {\big\|}_2^2}_{ = \mathcal{L}_x} \, - \, \theta \, \underbrace{\frac{1}{n}  \sum_{i=1}^N \big\| a_i - \hat{a}_i {\big\|}_2^2}_{ = \mathcal{L}_a},
\end{equation}
where $\mathcal{L}_x$ is the reconstruction loss (i.e., the mean squared error in predicting covariates $X$ using the representation $Z$) with $\hat{x}_i = g_x(\hat{z}_i)$ and $g_x: \mathcal{Z} \rightarrow \mathcal{X}$ (i.e., $g_x$ is the inverse of $\phi$), $ \mathcal{L}_a$ is treatment prediction loss (i.e., the mean squared error in predicting the treatment $A$ using the representation $Z$) with $\hat{a}_i = g_a(\hat{z}_i)$ and $g_a: \mathcal{Z} \rightarrow \mathcal{A}$, $p$ is the dimension of the covariate space, and $\theta > 0$ is a trade-off parameter. We discuss both the reconstruction loss and the treatment prediction loss in the following. 

The reconstruction loss $\mathcal{L}_x$ enforces the invertibility of the representation that would allow to reconstruct the original covariates from the representation. When $\phi$ is an invertible mapping, then the ignorability assumption holds with respect to the representation as well, i.e., $Y(a) \perp \!\!\! \perp A \mid X \, \, \implies Y(a) \perp \!\!\! \perp A \mid Z = \phi(X) \, \, $ (see \cite{Zhang2020}). Hence, enforcing invertibility is important because it ensures that the original covariate information is preserved in a way that the representation can be used to control for confounding.

The treatment prediction loss $\mathcal{L}_a$ (with a negative sign) aims to reduce treatment selection bias by ensuring that the learned representation of covariates is not predictive of the treatment. In the presence of treatment selection bias, there can be a strong dependence between covariates and treatment. This often results in having covariate domains where certain levels of treatment are rarely observed in observational data, which means that heterogeneous treatment effects cannot be reliably estimated in these domains. One approach to deal with this problem is to learn a balanced representation of covariates that is not predictive of the treatment, thereby removing the association between treatment and the respective representation of covariates. This domain adaptation technique was applied for categorical treatments, with treatment prediction loss being cross-entropy loss (see \cite{BicaCRN2020}). While previous work was focused on binary and not continuous treatments, it has been shown that domain adaptation for continuously indexed domains can be implemented via an $L_2$-loss \cite{WangDA2020}. Hence, we implement a similar domain adaptation approach as in \cite{BicaCRN2020}, i.e., using adversarial learning with gradient reversal layer (GRL) \cite{Ganin2016}, but adapt it to our setting with continuous treatment where our $\mathcal{L}_a$ is set to the $L_2$-loss. 

Thus, the balancing autoencoder produces a low-rank representation of covariates $\hat{Z} = \hat{\phi}(X)$ (where $\hat{\phi}$ is the representation $\phi$ learned from data) which aims to (i)~preserve the original covariate information by minimizing the reconstruction loss $\mathcal{L}_x$, and (ii)~remove the association between covariates and treatment by maximizing treatment prediction loss $\mathcal{L}_a$. This allows us to use $\hat{Z}$ as control variables to adjust for confounding while simultaneously addressing treatment selection bias. 

\noindent
\underline{\emph{Component (2): Counterfactual generator.}} The counterfactual generator computes counterfactual outcomes for varying treatments and covariates (i.e., for varying aid volumes and varying country characteristics). Here, the aim is to address the missing counterfactual outcomes in observational data and improve the predictions by combining the observational data with \emph{additional} counterfactual outcomes. Our approach for counterfactual generation is motivated by results from earlier research in machine learning \cite{Bica2020}. However, different from \cite{Bica2020}, we refrain from using a generative adversarial network (GAN), due to the small sample size and the poor fit as a result; instead, we develop a custom approach. 

We compute the counterfactual outcomes by weighting data points from the observational data. Specifically, for a given treatment $\tilde{a}$, covariate representation $\hat{z}$, and data $\hat{\mathcal{D}} =  \{(y_i, a_i, \hat{z}_i)\}_{i=1}^{n}$ (i.e., the observational data $\mathcal{D}$ after applying the balancing autoencoder), we compute the optimal weights by solving the following optimization problem
\begin{equation}
\label{eq:w_optim}
w^*  = \argmin_{ w \in \mathbb{R}^{n}}  \big\| \hat{z} - \hat{Z}_{\mathrm{mat}}^T \, w {\big\|}_2^2  \, + \, \alpha \, \| w {\|}_s \;\; 
 \mathrm{s.t.} \;\; A_{\mathrm{vec}}^T \, w = \tilde{a}  ,   
\end{equation}
where $\hat{Z}_{\mathrm{mat}}$ is an $n \times r$ matrix with rows being covariate representations from the observational data $\hat{z}_1, \ldots, \hat{z}_n$,  $A_{\mathrm{vec}}$ is a vector of treatments $a_1, \ldots, a_n$ from the observational data, and $s$ is the order of the norm (here: $s=1$). After computing optimal weights $w^*$, these are then used to compute the counterfactual outcome $\tilde{y}$ for treatment $\tilde{a}$ and covariate representation $\hat{z}$ by weighting outcome values from the observational data, i.e., $\tilde{y} = Y_{\mathrm{vec}}^T \, w^*$. 

The idea behind the counterfactual generator is to reweight the observational data to create a ``synthetic twin'' of a data point $(y_i,a_i,\hat{z}_i)$ under our causal graph. In our \framework, a synthetic twin then has the following two properties: (i)~the treatment of the synthetic twin equals some desired treatment $\tilde{a}$ for which we want to generate the counterfactual outcome; and (ii)~the covariate representation of the synthetic twin $\tilde{z}$ should resemble the covariate representation $\hat{z}_i$ of the data point (while addressing treatment selection bias). By computing $w^*$ using the observational data and Eq.~\eqref{eq:w_optim}, we generate the synthetic twin $(\tilde{y}, \tilde{a}, \tilde{z})$. Then, since the synthetic twin and the data point ideally differ only with respect to the value of the treatment, we can use the outcome of the twin $\tilde{y}$ as a prediction for the counterfactual outcome for the data point with covariate representation $\hat{z}_i$ if the treatment value was $\tilde{a}$. As such, we can add the data point $(\tilde{y}, \tilde{a}, \hat{z}_i)$ and increase the size of our observational data. 

The benefit of generating counterfactual outcomes is that we gain access to additional (counterfactual) outcome--treatment pairs, which can improve precision in predicting treatment--response curves. Notwithstanding, counterfactual generation procedure cannot provide ground-truth counterfactual outcomes, but rather predictions of counterfactual outcomes based on the observational data. Hence, to achieve performance gains, the counterfactual generator has to be sufficiently precise such that the error reduction when predicting treatment--response curves with additional counterfactual data outweighs the error in predicting counterfactual outcomes during counterfactual generation. 

We implement the counterfactual generator by generating $m$ counterfactual outcome--treatment pairs for each data point in the observational data. Specifically, for each country $i$, with corresponding covariate representation $\hat{z}_i$, we first sample $m$ treatment values $\{ \tilde{a}_{ij}\}_{j=1}^m$ uniformly from the interval $[A_{\mathrm{min}}, A_{\mathrm{max}}]$. Then, for each treatment $\tilde{a}_{ij}$ and corresponding $\hat{z}_i$, we compute the optimal weights $w_{ij}^*$ using the observational data and Eq.~\eqref{eq:w_optim}, and, then, we generate the counterfactual outcome $\tilde{y}_{ij}$. As a result, the counterfactual generator is used to augment the observational data with $m \cdot n$ counterfactual outcome--treatment pairs to improve the precision in predicting treatment--response curves.

\noindent
\underline{\emph{Component (3): Inference model.}} The inference model is used to predict treatment--response curves from given data by learning a function $f: \mathcal{A} \times \mathcal{Z} \rightarrow \mathcal{Y}$. This is the final step in \framework after applying the balancing autoencoder and the counterfactual generator to the observational data. In principle, any machine learning regression method can be used as an inference model; however, our focus is on the prediction of treatment effects of a continuous treatment. Hence, as inference model, we use a model from the causal machine learning literature that is aimed at predicting the effect of a continuous treatment, namely, the generalized propensity score (GPS) \cite{Hirano2004}. We have chosen the GPS as we expect it to be more robust in small sample size settings due to its parsimonious structure as compared to deep learning-based causal machine learning methods (e.g., the dose-response network (DRNet) \cite{Schwab2020}). We later evaluate other inference models (see supplements in our GitHub repository) and thereby confirm empirically that our choice of the GPS is preferred.

The GPS involves a two-step estimation procedure. In the first step, the conditional distribution of the treatment given covariates (in our case, it is the covariate representation, as the inference model is applied after the balancing autoencoder) is modeled under the assumption of a Gaussian distribution, i.e., 
\begin{equation}
    A_i \mid Z_i \sim N(\beta_0 + \beta_1 Z_{1i} + \ldots + \beta_r Z_{ri}; \, \sigma^2),
\end{equation}
where $r$ is the size of the covariate representation space. Once the estimates $\hat{\beta}_0, \hat{\beta}_1, \ldots, \hat{\beta}_r, \hat{\sigma}$ are obtained (e.g., via maximum likelihood estimation), they are used to compute the estimates of GPS as follows:
\begin{equation}
    \hat{R}_i = \frac{1}{\sqrt{2\pi\hat{\sigma}^2}} \exp \bigg( - \frac{1}{2\hat{\sigma}^2} \big( A_i - \hat{\beta}_0 - \hat{\beta}_1 Z_{i1} - \ldots - \hat{\beta}_r Z_{ir} \big)^2 \bigg).
\end{equation}
In the second step, the heterogeneous treatment--response curves are estimated by using a 2nd-degree polynomial regression given by
\begin{equation}
    Y_i = \alpha_0 + \alpha_1 A_i + \alpha_2 A_i^2 + \alpha_3 R_i + \alpha_4 R_i^2 + \alpha_5 A_i \, R_i.
\end{equation}
Hence, the full GPS model is estimated by obtaining parameter estimates $\hat{\beta}_{\mathrm{GPS}} = (\hat{\beta}_0, \hat{\beta}_1, \ldots, \hat{\beta}_r) $, $\hat{\sigma}_{\mathrm{GPS}}$ and $\hat{\alpha}_{\mathrm{GPS}} = (\hat{\alpha}_0, \hat{\alpha}_1, \ldots, \hat{\alpha}_5) $.

Why is the GPS able to predict treatment--response curves in small sample-size settings? Recall that the GPS is highly parsimonious: the first step has $r+2$ parameters, and the second step has 6 parameters, which amounts to only $r+8$ parameters to be estimated. In our case, we have $p=14$ covariates, and the covariate representation space $r$ is lower than $p$ by construction (see the range for the representation layer size of \framework in supplements in our GitHub repository). Hence, the GPS can be estimated well in our setting. This is a crucial difference of our framework from common baselines that build upon neural networks or other machine learning models (e.g., \cite{Bica2020,Schwab2020}). 

\subsection{Experimental details} 

\noindent
\underline{\emph{Semi-synthetic data.}} To assess the effectiveness of our \framework in predicting treatment--response curves, we need to evaluate prediction performance on counterfactual outcomes (and not only on factual outcomes). Because counterfactual outcomes are unobservable in the real-world data, we follow common practice in machine learning \cite{Schwab2020, Bica2020} and make use of semi-synthetic data where ``ground-truth'' counterfactual outcomes are known. Formally, we simulate semi-synthetic data as given in Supplement~\ref{supp:semisynthetic_data}.

\noindent
\underline{\emph{Baselines.}} We use the following baselines: (i)~linear model (LM) of varying order with regularization term (i.e., linear regression, LASSO regression, or ridge regression); (ii)~artificial neural network (ANN) for non-linearities; (iii)~generalized propensity score (GPS) \cite{Hirano2004}; (iv)~dose-response networks (DRNet) \cite{Schwab2020}; and (v)~SCIGAN \cite{Bica2020}. 

Baselines (i)--(ii) are standard machine learning models. While they are applicable to our setting, these baselines are not tailored for treatment effect estimation and, therefore, do \underline{not} address causal aspects such as treatment selection bias. Baselines (iii)--(v) are from the causal machine learning literature representing state-of-the-art benchmarks focused on predicting the effect of continuous treatment. Of note, there are other methods that deal with treatment effect estimation (e.g., causal forests), yet with a focus on \emph{binary} and \underline{not} on \emph{continuous} treatments \cite{Johansson2016, Shalit2017, Yoon2018, Athey2019, Hatt2021}. Hence, these methods are \underline{not} applicable to our setting. 


\noindent
\underline{\emph{Performance metrics.}} In our experiments, we evaluate the performance of \framework in two ways: (i)~counterfactual outcome prediction using semi-synthetic data, i.e., performance in predicting treatment--response curves; and (ii)~factual outcome prediction using real-world data. 

For (i), we follow prior literature \cite{Schwab2020} and use the mean integrated squared error (MISE) given by
\begin{equation}
\label{eq:mise}
\mathrm{MISE} = \frac{1}{n}\sum_{i=1}^n \int_{a = A_{\mathrm{min}}}^{A_{\mathrm{max}}} \big( y_i(a) - \hat{y}_i(a) \big)^2 \, \mathrm{d}a ,
\end{equation}
where $y_i (a)$ is the respective outcome given aid volume $a$, $\hat{y}_i (a)$ is the predicted outcome given aid volume $a$, and the interval $[A_{\mathrm{min}}, A_{\mathrm{max}}]$ is the observed interval of development aid in real-world data. The MISE has the advantage that we benchmark the performance across a range of different aid volumes $[A_{\mathrm{min}}, A_{\mathrm{max}}]$. Formally, we compute the inner integral of MISE using Romberg integration with 64 equally spaced samples of development aid on the interval $[A_{\mathrm{min}}, A_{\mathrm{max}}]$. 

For (ii), we use the root mean squared error. Note that we use the RMSE for real-world data (as opposed to the MISE), since the performance must be computed using a single outcome (as opposed to the complete treatment--response curve).

\noindent
\underline{\emph{Implementation.}} We used Python 3.7 and PyTorch 1.7. The training is done in batches for a given number of epochs. We use Adam \cite{Kingma2015} with learning rate $\eta$ for optimization.  Hyperparameter tuning is performed via cross-validation with 80/20 split (see supplements in our GitHub repository for details). The performance is reported as mean $\pm$ standard deviation (averaged over 10 runs).

\noindent
\underline{\emph{Decision-making problem.}} In order to understand the potential downstream impact of our framework and how it can support aid allocation in practice, we introduce the following decision-making problem. We study a hypothetical setting where an aid allocation under budget constraints should be identified that minimizes the total annual number of new HIV infections across countries. The idea is that such an analysis should identify countries in need of development aid to not fall behind. Reassuringly, we emphasize that the suggested allocation should not be interpreted as an optimal allocation due to various idiosyncrasies in practice, but rather as illustrative for how our framework can help decision-making in identifying needs and funding gaps. 

Formally, we solve the following decision-making problem:
\begin{align}
\label{eq:opt_alloc}
[a_1, \ldots, a_n ]^*  = \argmin_{\substack{a_i \in [0, L]\\ i = 1, \ldots, n}} & \sum_{i=1}^n \big(1-\hat{y_i}(a_i, x_i)\big) \, r_i  \, p_i \;\;
 \mathrm{s.t.} \;\; \sum_{i=1}^n a_i \leq B    \nonumber, 
\end{align}
where $L$ is the bound for aid allocation to a single country, $\hat{y_i}(a_i, x_i)$ is the predicted relative reduction in the HIV infection rate (predicted by our \framework for given aid volume $a_i$ and given country characteristics $x_i$), $r_i$ is the HIV infection rate for a given country from the previous year, $p_i$ is the population size in a given country, and $B$ is the available budget. In our analysis, we set $B$ to the total observed volume of aid allocated in the current year. We bound the disbursement, $a_i$ for $i = 1, \ldots, n$, to a single country to the range $[0, L]$ with $L = A_{\mathrm{max}}+\hat{\sigma}_{A}$, where $A_{\mathrm{max}}$ is the maximal value of observed aid for a single country, and $\hat{\sigma}_{A}$ is the estimated standard deviation for the aid variable $A$. We do this to prevent overly large extrapolation when providing estimates of aid effects using $\hat{y_i}(a_i, x_i)$. In our implementation, we solve the optimization problem using sequential least squares programming \cite{Kraft1988}.

\section{Results}

\subsection{Results for semi-synthetic data}

We make a few important observations based on the results of our experiments with semi-synthetic data (see Figure~\ref{fig:results}a). First, \framework is the best-performing method in predicting treatment--response curves. We observe that the square root of MISE for the best baseline (GPS) is 0.205 $\pm$ 0.000. In contrast, for our \framework, it is 0.158 $\pm$ 0.024. Hence, our method reduces the prediction error by at least 23\% compared to the baselines. This shows that our \framework is a superior method for predicting heterogeneous treatment--response curves in our HIV setting. Second, we observe a large estimation error and a large variance for SCIGAN (i.e., 11.443 $\pm$ 9.303), which is a state-of-the-art method that has been shown to have superior performance over both DRNet and GPS \cite{Bica2020}. However, SCIGAN is based on generative adversarial networks (GANs), which may have poor fit in small sample size settings. Given that in our case, we have a sample of 105 countries (and thus much smaller than experiments in \cite{Bica2020}), it is not surprising that SCIGAN has challenges in our HIV setting.

\subsection{Results for real-world data }


We now apply our machine learning framework \framework to real-world HIV data. Here, the main purpose is to demonstrate the applicability of our method (rather than performance benchmarking). The reason is that ground-truth treatment effects are absent in real-world data, and, therefore, the performance in predicting treatment--response curves can no longer be evaluated. 

\begin{figure}[htb]
\begin{minipage}[b]{0.25\textwidth}
{\raggedright\quad\quad\figletter{a}}
\vspace{0.1cm}

\centering
    \includegraphics[width=0.9\textwidth]{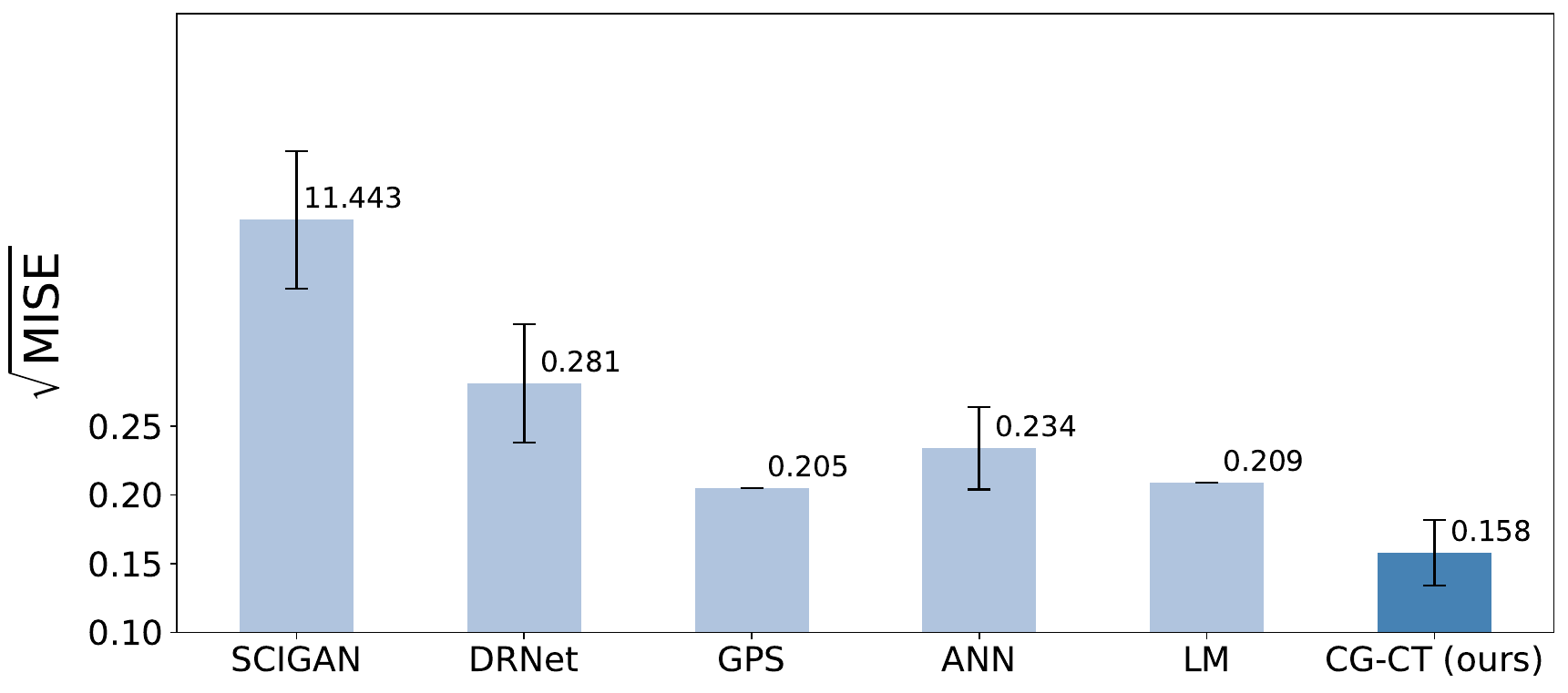}
    \label{fig:res_semi_syn}
\end{minipage} \hfill
\begin{minipage}[b]{0.2\textwidth}
 
{\raggedright\quad\quad\figletter{b}}
\vspace{0.1cm}

\centering
    \includegraphics[width=0.9\textwidth]{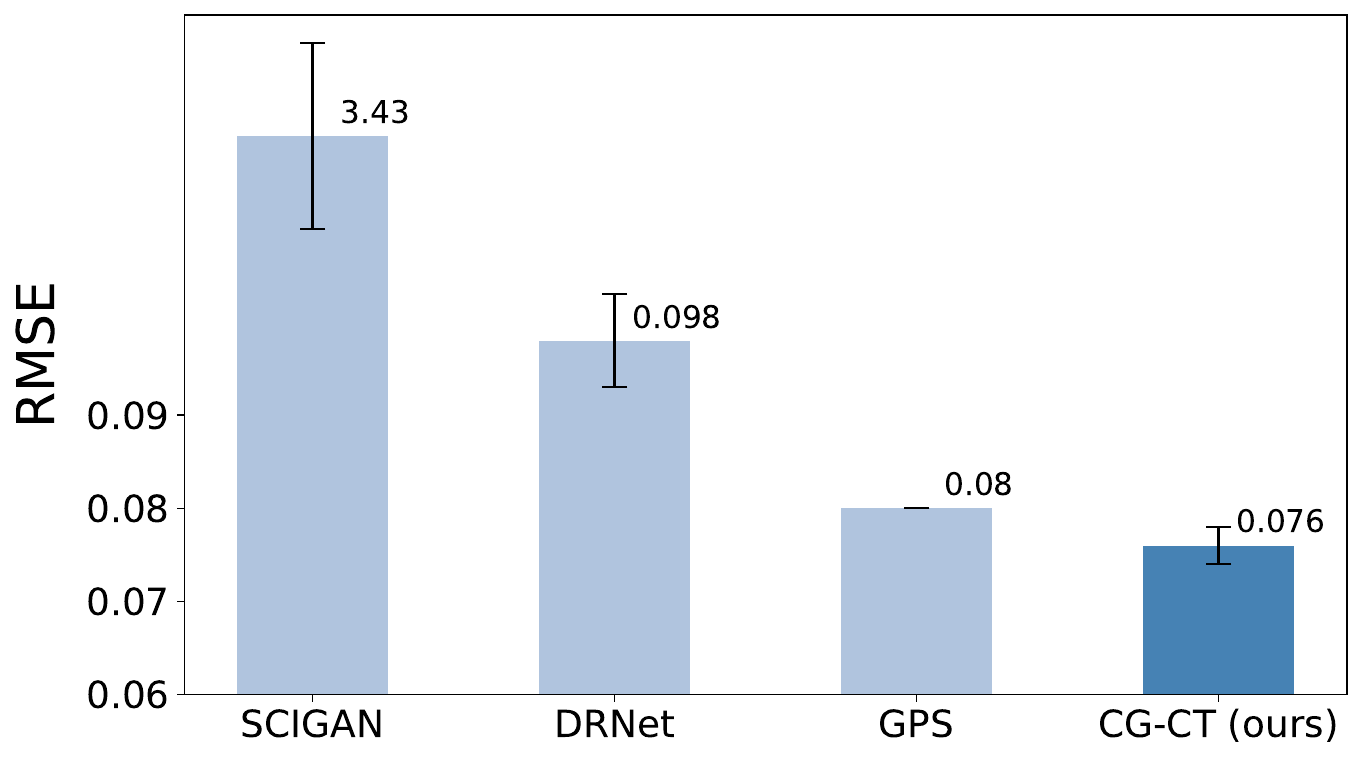}
    \label{fig:res_real_world}
\end{minipage}

\begin{minipage}[b]{0.5\textwidth}

{\raggedright\figletter{c}}
\vspace{0.1cm}

\centering
    \includegraphics[width=0.9\textwidth]{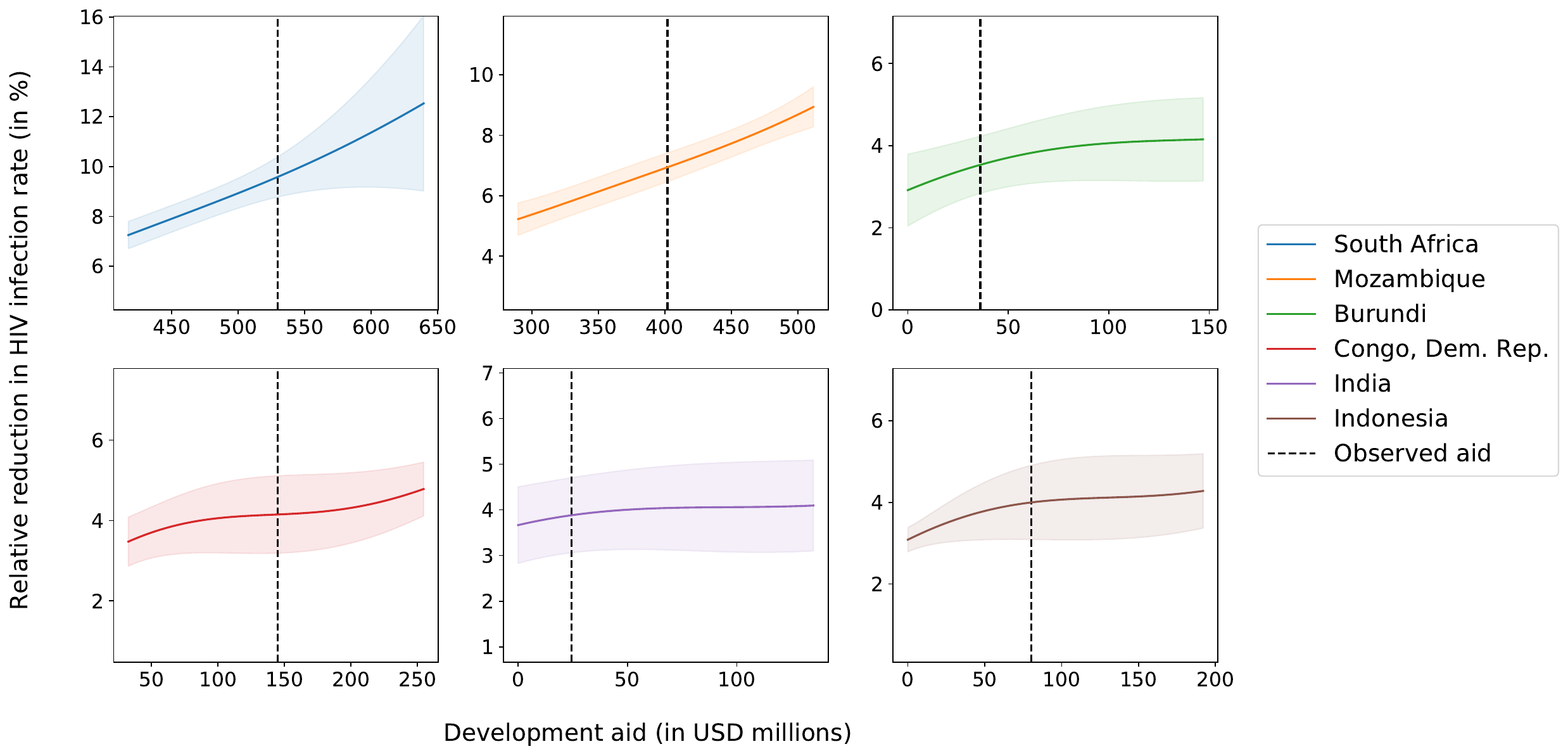}
    \label{fig:tr_curves}
\end{minipage}
\caption{\textbf{Results for predicted aid--response curves.} \figletter{a}~Results from the experiments with semi-synthetic data. 
\figletter{b}~Results from the experiments with real-world data. 
\figletter{c}~Predictions for aid--response curves in six example countries. 
The predicted aid--response curves for the remaining countries are provided in supplements in our GitHub repository. Vertical dashed line denotes the actual volume of development aid as observed in 2017. The $x$-axis is set to $A_\text{obs}$ $\pm$ $\hat{\sigma}_{A}$ (with a cut-off at zero to prevent negative values for Burundi and India), where $A_\text{obs}$ is the observed development aid and $\hat{\sigma}_{A}$ is the estimated standard deviation of development aid in 2017. 
}
\label{fig:results}
\end{figure}

We compare our \framework against baselines that produce treatment effects, i.e., GPS, DRNet, and SCIGAN  (see Figure~\ref{fig:results}b). This was done intentionally to prevent the advantages of more flexible methods that directly optimize their performance against factual outcome prediction and, thus, to ensure a fair comparison. Importantly, our results confirm that the baselines for predicting treatment-response curves are consistently outperformed by our \framework. In particular, we show an improvement of at least 5\% in RMSE. We again observe a large RMSE for SCIGAN along with a large variance (i.e., 3.430 $\pm$ 2.393), which can be attributed to challenges of fitting a GAN in a small sample setting. 

We also report the performance of standard machine learning methods in the following. However, these have the advantage of being designed for factual outcome prediction (whereas our method is designed for predicting treatment effects), and, therefore, we expect our \framework to achieve a somewhat similar performance. Here, the RMSE for the standard machine learning baselines is 0.078 $\pm$ 0.000 (for LM) and 0.083 $\pm$ 0.003 (for ANN). Our \framework achieves 0.076 $\pm$ 0.002, which is an improvement of 2.56\%. Hence, our framework outperforms the standard machine learning baselines as well. 



We use \framework to predict aid--response curves for different countries in 2017 (see Figure~\ref{fig:results}c). The aid--response curves can be used by public decision-makers to answer \emph{`what if?'} questions; that is, how potential changes in aid volumes would change the expected reduction in HIV infection rates. Overall, we see a clear pattern: development aid is predicted to have a positive effect on the reduction in HIV infection rates. This is expected as such a relation was empirically confirmed in retrospective studies \cite{Munyanyi2020}. Importantly, however, we observe considerable between-country heterogeneity in the predicted aid--response curves. For example, the observed aid for Mozambique (in orange) is USD 402 million, and the expected reduction in HIV infection rate predicted by our \framework is 6.94\%. Here, an increase in development aid of USD 50 million would result in expected reduction in the HIV infection rate by 7.76\%. In contrast, the observed aid for Congo (in red) is USD 145 million, and the expected reduction in the HIV infection rate predicted by our \framework is 4.15\%. Here, an increase in development aid of USD 50 million would result in expected reduction in the HIV infection rate by 4.29\%. Hence, an increase in development aid by USD 50 million is predicted to have a stronger impact in Mozambique than in Congo. Moreover, we note that our predicted aid--response curves predict a reduction in HIV infection rates even in the absence of aid, albeit of different magnitude across countries. This can be expected as reductions can also be driven by country characteristics and other trends (e.g., domestic financing to end the HIV epidemic, growing disease awareness, etc.). In sum, the aid--response curves may help public decision-makers when allocating aid by informing them about the expected effect of a particular aid allocation on HIV infection rates across countries. 

\subsection{Potential for more effective aid allocation} 

We now provide insights into how our framework can inform the decision-making around aid allocation. 
Our decision-making problem seeks to determine an aid allocation that minimizes the total annual number of new HIV infections across all countries under a budget constraint (see \nameref{sec:methods} for details). Here, we use the predicted aid--response curves as input, which provide predictions of the expected reduction in the HIV infection rate compared to the previous year for a given aid volume and given country characteristics. We use real-world HIV data from the year 2017. Specifically, to limit the overall available volume of development aid, we set the budget constraint to the observed total volume of aid allocated in the year 2017. For evaluation, we compare (i)~the expected annual number of new HIV infections under our suggested allocation against (ii)~the expected annual number of new HIV infections under current allocation practice. To account for variation, we report bootstrapped \cite{Efron1992} confidence intervals.

\begin{figure}[htb]
\begin{minipage}[t]{0.43\columnwidth}

{\raggedright\quad\quad\figletter{a}}
\vspace{0.1cm}

\centering
    \includegraphics[width=\textwidth]{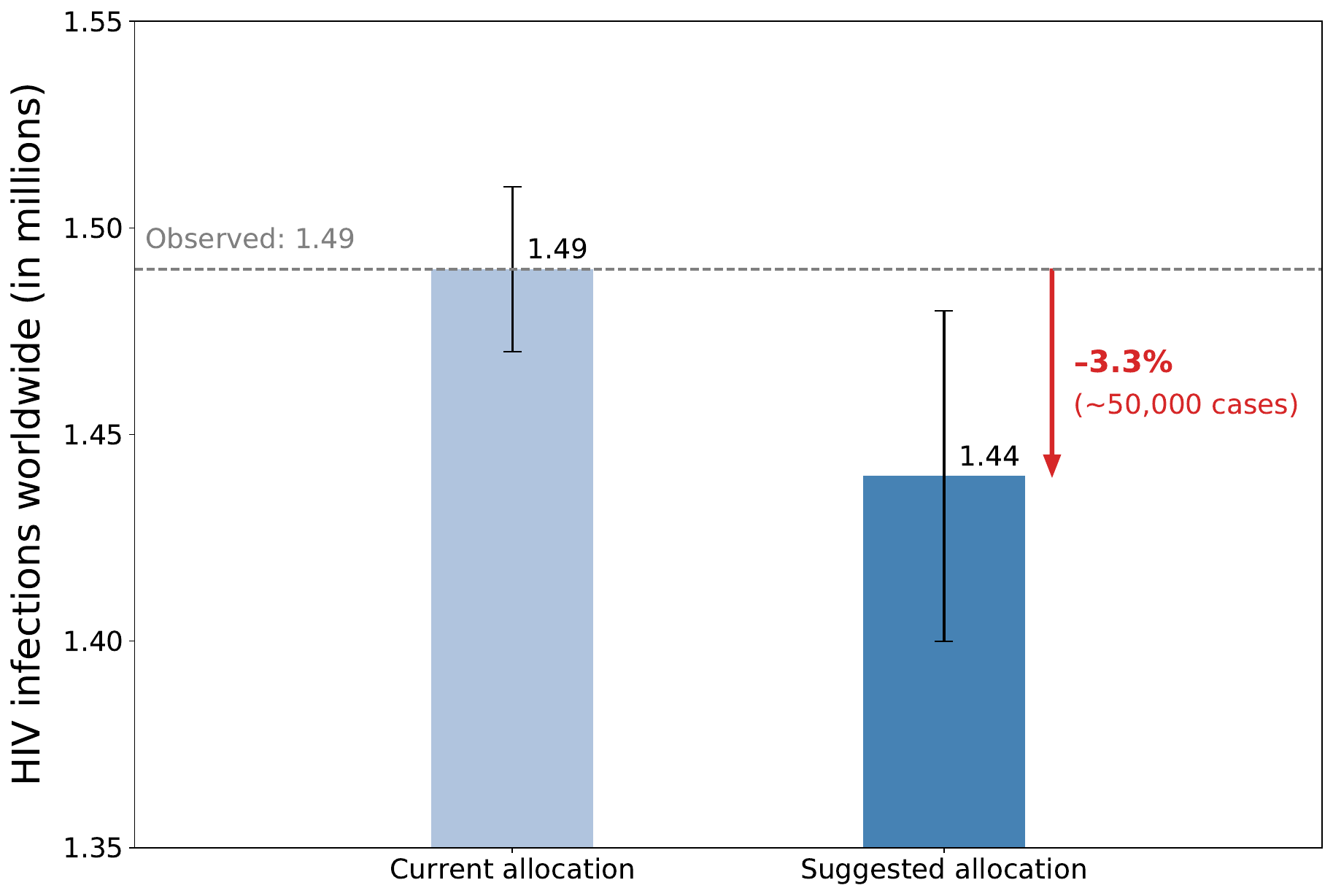}
    \label{fig:exp_HIV_cases}
\end{minipage}
\begin{minipage}[t]{0.55\columnwidth}
 
{\raggedright\quad\quad\quad\figletter{b}}
\vspace{0.1cm}

\centering
    \includegraphics[width=\textwidth]{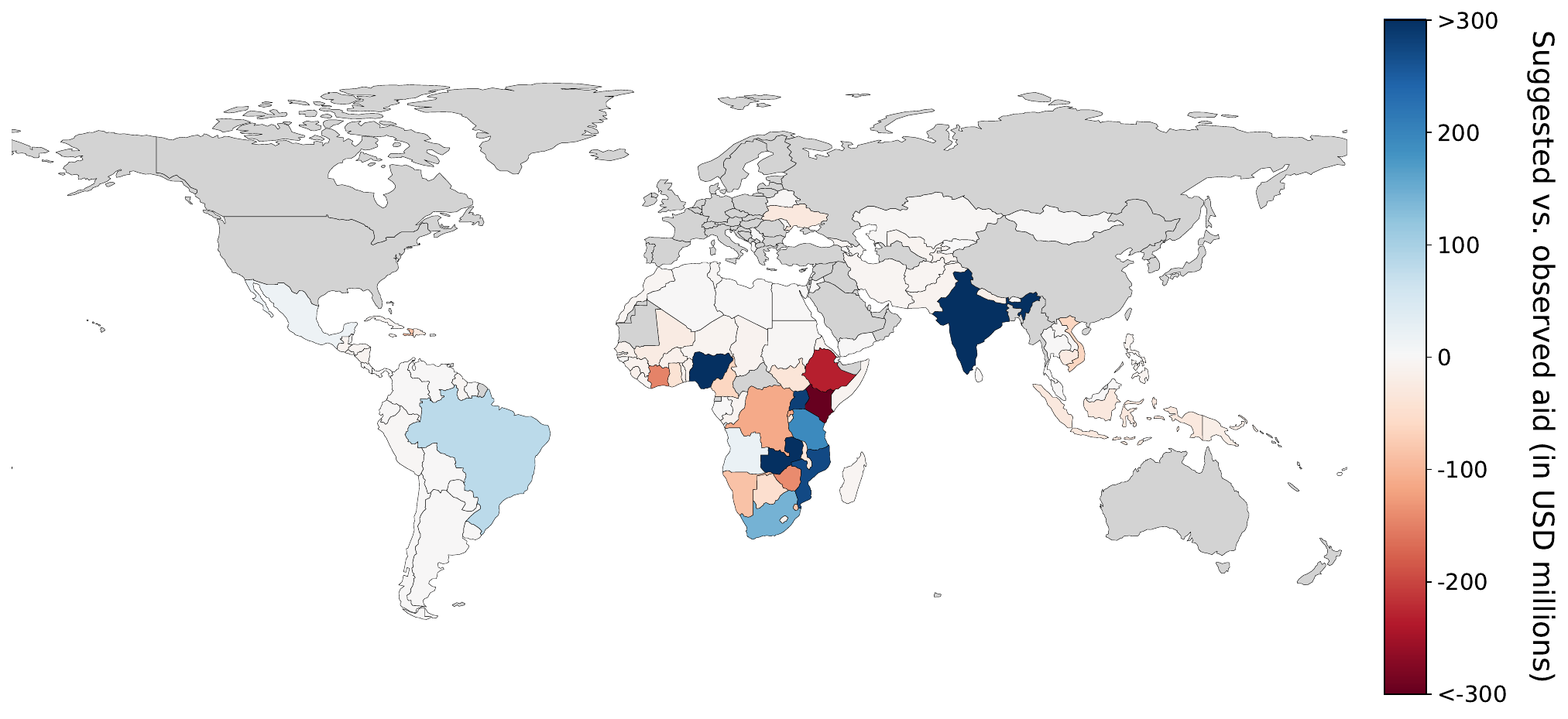}
    \label{fig:opt_alloc}
\end{minipage}
\caption{\textbf{Suggested vs. current aid allocation.} \figletter{a}~Reduction in the expected number of new HIV infections worldwide under the suggested allocation vs. the current allocation of development aid. 
\figletter{b}~Change in the aid volume between the suggested and the current aid allocation (in USD millions). Gray: countries that were not aid recipients in 2017.}
\label{fig:optimal_allocation}
\end{figure}

Our results (see Figure~\ref{fig:optimal_allocation}a) show that the expected number of new HIV infections under the current allocation is 1.49 $\pm$ 0.02 million. For our suggested allocation, the expected number of new HIV infections is 1.44 $\pm$ 0.04 million. The reduction amounts to 3.3\% $\pm$ 1.8\%. This corresponds to $\sim$50,000 fewer cases (i.e., 48,925 $\pm$ 26,779). As a robustness check for our modeling, we note that the expected number of new HIV infections predicted by our framework for the observed aid allocation (i.e., 1.49 million new HIV infections) matches the actual number of new HIV infections observed in the data (Figure~\ref{fig:optimal_allocation}a, gray dotted line). This again corroborates that our framework provides precise outcome predictions.

We further compare aid volumes under current aid allocation vs. the allocation suggested by our \framework (see Figure~\ref{fig:optimal_allocation}b). Importantly, our framework identifies countries with higher needs for aid than before. For example, our framework suggests increasing aid for Zambia by USD 366 million and for Nigeria by USD 329 million. Overall, our framework suggests increasing aid disbursements in several countries in Southeast Africa (e.g., South Africa, Mozambique, Tanzania, Zambia). Furthermore, our machine learning framework targets more aid towards non-African countries, which have previously not received large amounts of aid in the past despite having a comparatively large annual number of new HIV infections (e.g., India, Brazil). In sum, our findings demonstrate the effectiveness of our \framework to inform cost-efficient allocation of development aid that maximizes progress towards ending the HIV epidemic. 

\subsection{Robustness checks}

In order to further validate our \framework, we perform several additional analyses. (1)~We visually inspect the covariate representations learned by the balancing autoencoder using t-SNE \cite{Maaten2008} and isomap \cite{Tanenbaum2000}. The visualizations are shown in supplements in our GitHub repository. We observe that the balancing autoencoder induces homogeneity of the covariate representation with respect to treatments, thereby successfully addressing treatment selection bias. (2)~We perform a sensitivity analysis of our \framework with respect to the balancing parameter $\theta$ (from the balancing autoencoder) and the number of generated counterfactual outcomes per country $m$ (from the counterfactual generator). We observe that the results remain robust when varying these hyperparameters (see supplements in our GitHub repository). (3)~We examine how \framework behaves when other (causal) machine learning methods (i.e., LM, ANN, DRNet) are used as the inference model instead of GPS. Here, we find that our \framework leads to consistent performance improvements over the baselines across different inference models (see supplements in our GitHub repository). Moreover, the results show that the choice of GPS is preferred. (4)~We perform an ablation study where we toggle the balancing autoencoder and the counterfactual generator on/off. We observe that the performance gains by our \framework are a result of combining the balancing autoencoder with the counterfactual generator (see supplements in our GitHub repository). This demonstrates the importance of our methodological innovations. (5)~We repeat the ablation study with other inference models for which we now toggle the balancing autoencoder and the counterfactual generator on/off. Again, we see clear performance improvements when combining both components (see supplements in our GitHub repository). This confirms that balancing autoencoder and counterfactual generator are important for obtaining precise treatment--response curves in our setting.    

We perform several analyses to examine the validity of the underlying assumptions for identifiability of treatment effects, namely, positivity and ignorability, as well as the robustness to the potential violation of the latter. (1)~We first show the estimated probability of aid volumes for different countries, where we observe that the range of aid volumes over which we optimize allocation does not lead to severe extrapolation (see supplements in our GitHub repository). (2)~We examine the robustness of the predicted treatment effect of aid when adding other covariates that were not included in country characteristics. Here, we focus on past values of development aid, past values of HIV infection rate, past aid volumes of neighboring countries, and past HIV infection rates of neighboring countries, which allows us to control for additional temporal dynamics and spillover effects between countries. We find that the effect remains robust, indicating that our set of country characteristics is sufficient to address potential confounding effects of these covariates (see supplements in our GitHub repository). (3)~We perform a causal sensitivity analysis to check the robustness of our method regarding violations of the ignorability assumption, that is, unobserved confounding. We use a state-of-the-art method \cite{Jesson.2022,Frauen.2023,Frauen.2024} to check that our predicted treatment--response curves remain stable under small violations of the ignorability assumption (see supplements in our GitHub repository). In other words, even in a hypothetical case of unobserved confounding, our results remain robust, and one cannot explain away the predicted treatment effect.  

To show robustness of our results in a different time frame, we perform an analysis where we use HIV data from the year 2015 for learning and from the year 2016 for evaluation. We observe that the conclusions of our analysis remain consistent and that the baselines are outperformed by our \framework (see supplements in our GitHub repository). We also experimented with multi-year data for learning; however, this did not lead to significant out-of-sample performance improvements. Here, one explanation is that allocation practices are subject to changes over time (e.g., changes in priorities, changes in programs, funding mix) \cite{Avila2013}.

\section{Discussion}
\label{sec:discussion}

\textbf{Strengths:} Our data-efficient \framework is tailored to our setting where it offers several strengths for predicting heterogeneous treatment effects. While there are numerous methods in the literature for binary treatments (e.g., causal forests) \cite{Johansson2016, Shalit2017, Yoon2018, Athey2019, Hatt2021}, there are comparatively few for continuous treatments \cite{Hirano2004, Schwab2020, Bica2020}. Different from existing methods, our \framework is designed to simultaneously address continuous treatments, high-dimensional covariates (country characteristics), treatment selection bias, and a small sample-size setting. Here, our framework shows a superior performance in predicting aid--response curves. For example, in our experiments with semi-synthetic data, our \framework outperforms both standard machine learning methods (i.e., the linear model and the artificial neural network) and common baselines for predicting heterogeneous treatment--response curves \cite{Hirano2004, Schwab2020, Bica2020} by at least 23\%. Our ablation studies (see supplements in our GitHub repository) show that our performance improvements are largely attributed to our combination of the balancing autoencoder and the counterfactual generator as effective levers to address the above challenges. Finally, our framework is data-efficient: the different components operate on a low-dimensional covariate space and polynomial regression together with data augmentation. This is a crucial difference of our framework from common baselines that build upon neural networks (e.g., \cite{Bica2020,Schwab2020}). To this end, we foresee future applications of our \framework in other decision-making settings. 


\textbf{Limitations:} As any other work, ours is not free of limitations, which provide interesting opportunities for future research. First, the standard challenge when predicting treatment effects from observational data is that counterfactual outcomes are not observable \cite{Rubin2005}. This is not unique to our framework but is inherent to essentially all methods from the literature \cite{Schwab2020, Bica2020}, which then make mathematical assumptions to infer treatment effects. Notwithstanding, conducting a randomized controlled trial as the gold standard for estimating treatment effects is costly and may thus be considered infeasible for our problem setup. Here, our proposed framework based on machine learning using observational data offers a scalable approach that relies on the standard assumptions for identifiability of treatment effects from the literature. While relying on these assumptions is a limitation, we have conducted several analyses to examine their validity in our setting and demonstrate robustness of our results under potential violations (see supplements in our GitHub repository).   
Second, there are other factors that may drive aid allocations in practice. Examples are sanctions (e.g., as in Afghanistan in 2021) or political aims (e.g., prioritizing specific regions or activities). Nevertheless, our framework is useful in such situations as it promotes transparency by informing about needs and funding gaps. This is especially relevant as development financing transitions towards bilateral negotiations about aid volumes between donors and recipients. Finally, we emphasize that our objective is not to coordinate aid disbursements at the level of individual funding bodies. This would require a tactical model to account for operational constraints, expertise, coordination, and other idiosyncratic aspects of individual aid activities, because of which such a model would not be meaningful in practice. Instead, we fulfill a direct need in practice \cite{OECDsdglab2022} by informing allocations of development aid on a global scale.


\textbf{Impact:} Our framework shows large untapped potential for making progress towards ending the HIV epidemic through cost-efficient aid allocation. Of note, our framework is not only applicable to the HIV epidemic but essentially to all other SDG targets. Thereby, our work supports public decision-makers in making important progress towards the SDGs. This is especially relevant due to ongoing challenges (e.g., the Russian invasion of Ukraine, global warming, etc.), necessitating a more cost-efficient use of resources. Finally, our framework has the potential for even more general applications beyond the SDGs, since it can be relevant for many decision-making problems from public policy, where data availability is a common challenge. Examples include financial policy where interest rates must be set to balance different economic outcomes, or public health measures from the COVID-19 pandemic with the aim to reduce infection rates. Such applications of our \framework would present both a strong potential for social impact, as well as promising areas for further research.

\textbf{Deployment:} The current framework presented in this paper is the result of a collaboration with the \emph{Organisation for Economic Co-operation and Development (OECD)}, an intergovernmental organization coordinating development aid activities of different donors. The \emph{SDG Financing Lab} of the \emph{OECD} has been tasked with developing dashboards around machine learning models, such as ours, that make data-driven insights available to decision-makers. Here, we report performance based on computational experiments (rather than from post-launch) due to two reasons. (i)~Aid allocation follows a multi-year process from goal setting to shifting funding, so that real-world evidence on SDG outcomes is only available with a significant delay. (ii)~Aid allocation follows a human-in-the-loop process, making it difficult to use post-launch data to isolate human judgment from the recommendations of our framework. Hence, we opted for a numerical evaluation to offer timely results and demonstrate broad applicability.


\clearpage

\bibliographystyle{ACM-Reference-Format-no-doi}
\bibliography{literature.bib}

\appendix

\renewcommand{\thetable}{S\arabic{table}}
\renewcommand{\thefigure}{S\arabic{figure}}
\setcounter{figure}{0}
\setcounter{table}{0}

\section{Overview of model variables}
\label{supp:control_vars} 

\begin{figure}[H]
\centering
\begin{minipage}[b]{.4\textwidth}
{\raggedright\figletter{a}}
\vspace{0.1cm}

    \centering
    \begin{overpic}[scale=0.202]{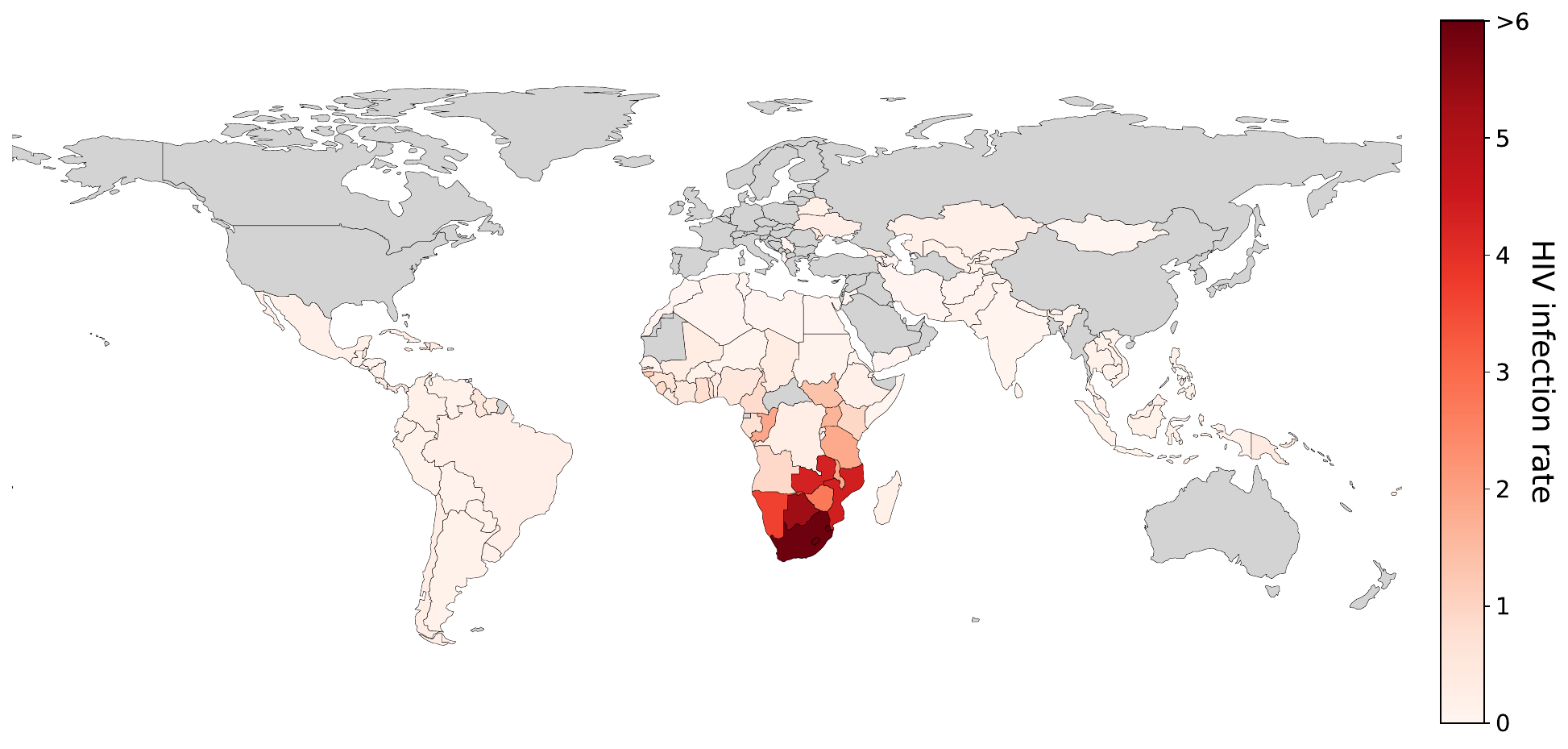}
     \put(3,3){\includegraphics[scale=0.072]{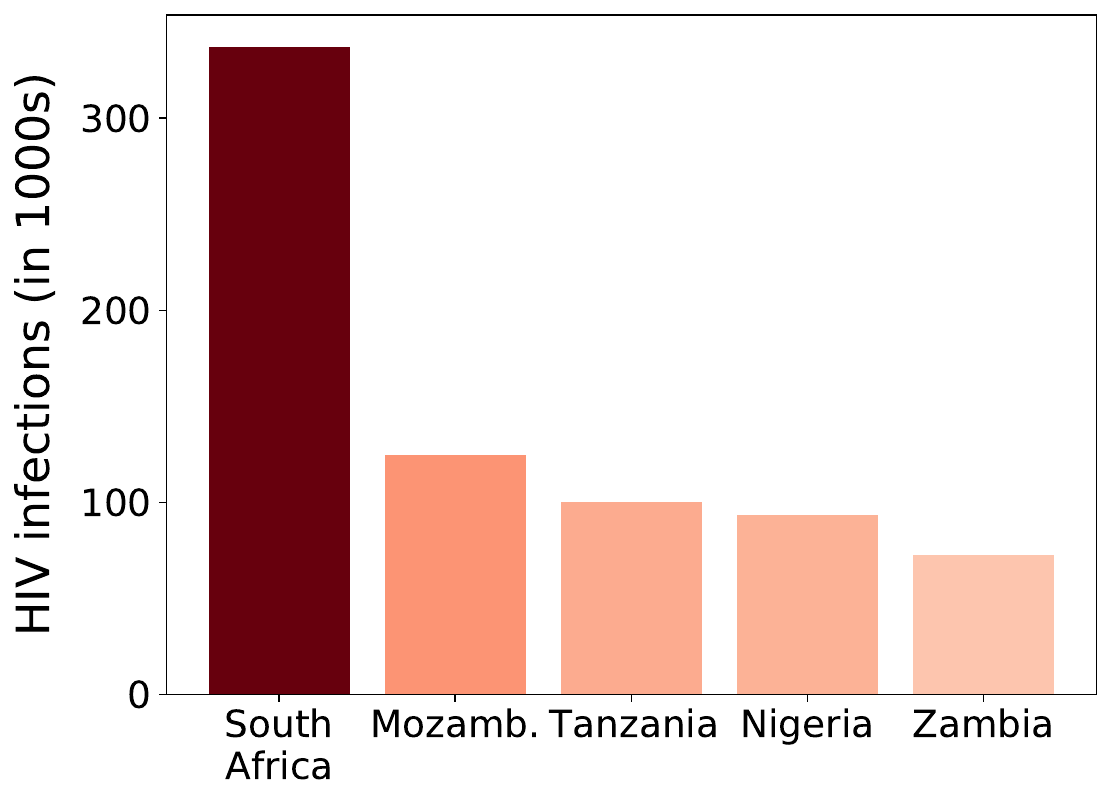}}  
     \label{fig:hiv_cases}
    \end{overpic}
    \label{fig:hiv_infection_map}
\end{minipage}

\vspace{0.5cm}

\begin{minipage}[b]{.4\textwidth}
{\raggedright\figletter{b}}
\vspace{0.1cm}

    \centering
    \begin{overpic}[scale=0.202]{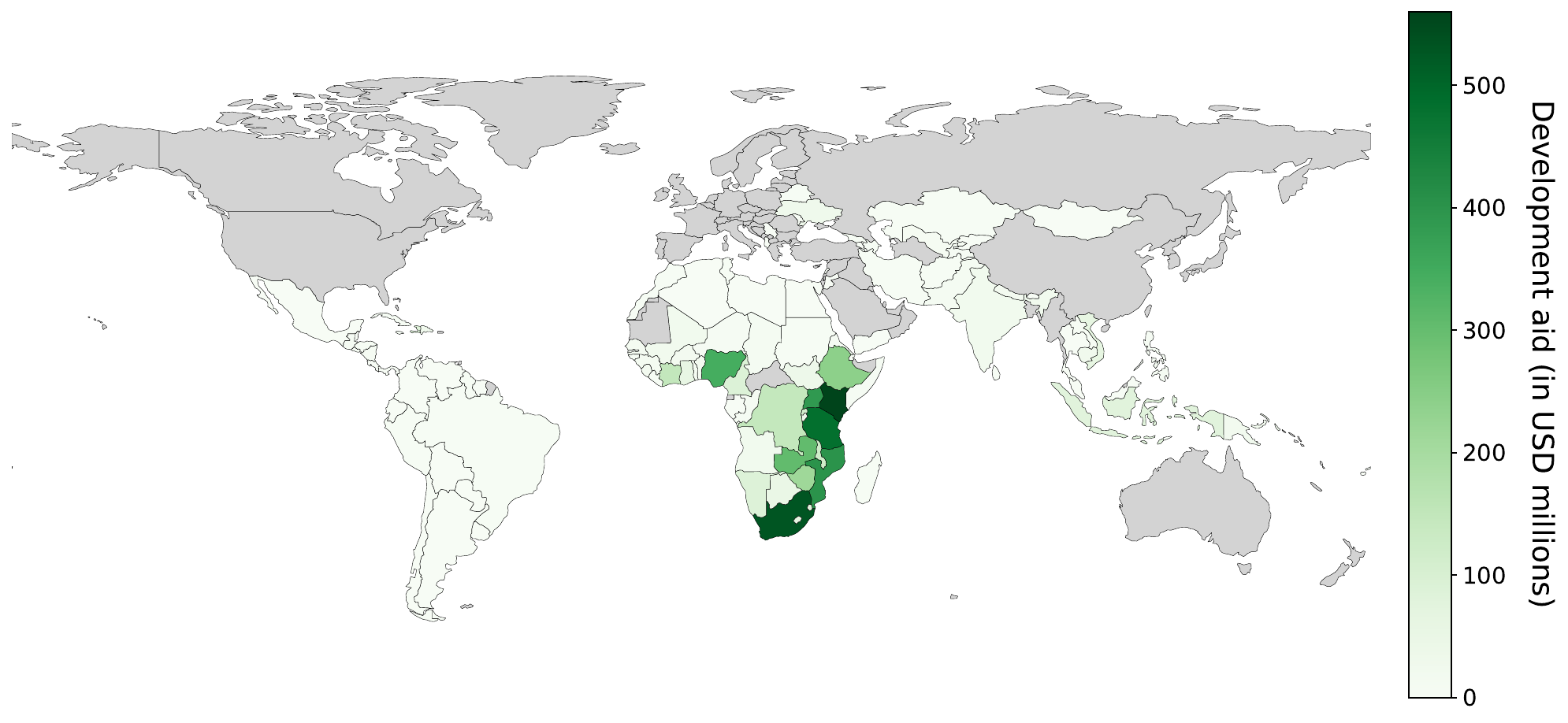}
     \put(3,3){\includegraphics[scale=0.072]{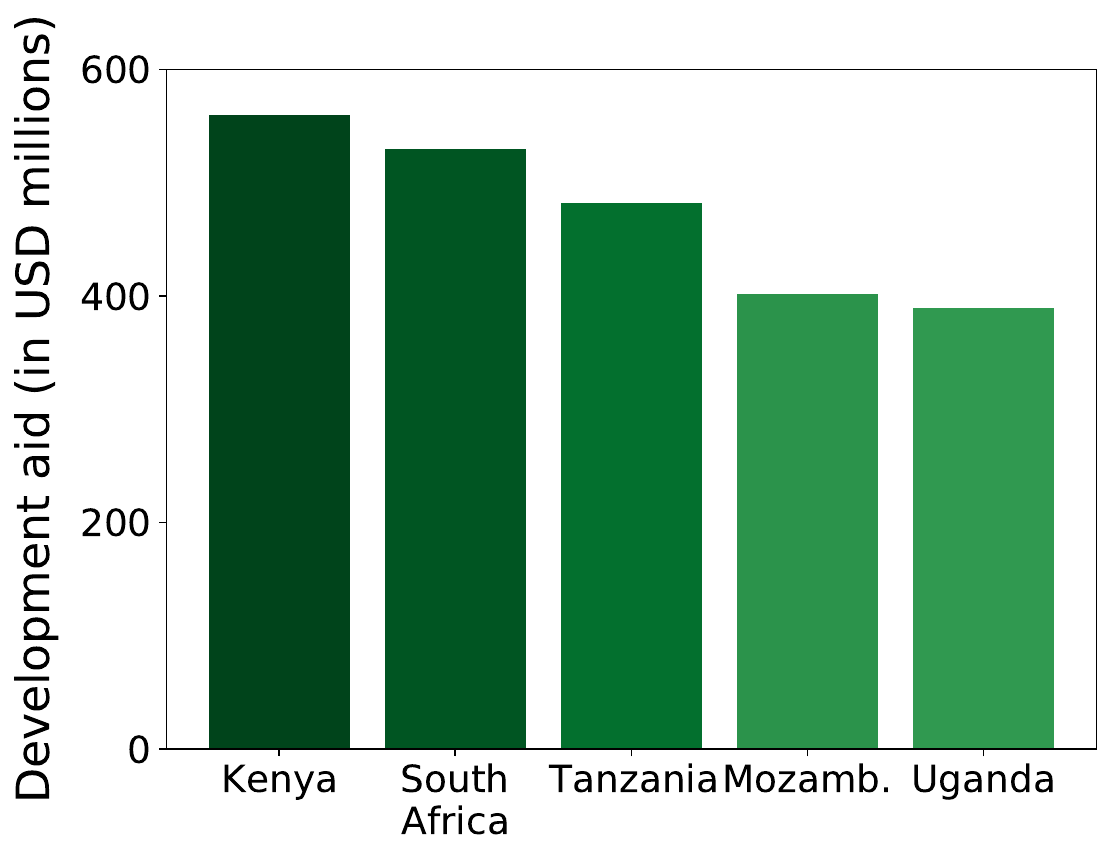}}  
     \label{fig:aid_recipients}
    \end{overpic}
    \label{fig:hiv_aid_map}
\end{minipage}

\caption{\textbf{Overview of HIV infections and aid disbursements.} \figletter{a}~HIV infection rate (i.e., the annual number of new HIV infections per 1,000 uninfected people in the population) per country in 2017. The subplot in the bottom-left corner shows five countries with the largest number of new HIV infections in 2017. \figletter{b}~Volume of HIV development aid in USD millions, in 2017. The subplot in the bottom-left corner shows five countries that were the largest recipients of development aid earmarked to end the HIV epidemic in 2017. Gray: countries that were not aid recipients in 2017.}
\vspace{-0.3cm}
\label{fig:data_overview}
\end{figure}

Ending the HIV epidemic is an important target of the SDG framework (Goal~3, Target~3.3.1) \cite{UNindicators2017}. In 2017, there were 35.9 million people living with HIV globally, with 1.7 million new infections and 800,000 HIV-related deaths \cite{UNAIDShiv2022}. Moreover, HIV has caused more than 40 million deaths since the beginning of the epidemic \cite{UNAIDShiv2022}, and it continues to be one of the leading causes of death, especially in low-income countries \cite{WHOhiv2020}. Overall, large HIV infection rates (i.e., the annual number of new HIV infections per 1,000 uninfected people in the population) are observed in Southern and Eastern Africa (see Figure~\ref{fig:data_overview}a). Countries with a particularly large number of new HIV infections per year are, for example, South Africa ($\sim$337,000), Mozambique ($\sim$125,000), Tanzania ($\sim$100,000), Nigeria ($\sim$94,000) and Zambia ($\sim$73,000) (see Figure~\ref{fig:data_overview}a, subplot). To this end, the HIV epidemic presents a major threat to global public health. 

Development aid is important for achieving the goal of ending the HIV epidemic, especially in developing countries that have limited domestic financing. In 2017, development aid committed to ending the HIV epidemic amounted to more than USD 5.2 billion. Large volumes of aid were allocated to African countries (see Figure~\ref{fig:data_overview}b). In particular, countries such as Kenya (USD 560 million), South Africa (USD 530 million), and Tanzania (USD 482 million) have received particularly large aid volumes (see Figure~\ref{fig:data_overview}b, subplot). 

Table~\ref{tab:vars} lists all variables that are used in our analysis: (i)~the outcome variable, (ii)~the treatment variable, and (iii)~the country characteristics as covariates. Here, we have intentionally included a wide range of socioeconomic, macroeconomic, and health-related country characteristics in order to control for potential confounding when predicting the aid--response curves. The choice for most of the covariates (e.g., GDP per capita, population, school enrollment, etc.) was motivated by previous work on aid effectiveness \cite{Birchler2016, Munyanyi2020, Jakubik2022}.

\begin{table}[h]
\begin{center}
\caption{Overview of model variables. The variables are grouped by: relative reduction in HIV infection rate (\emph{outcome}), development aid (\emph{treatment}), and country characteristics (\emph{covariates}). Reported are also summary statistics (SD: standard deviation). GDP per capita is reported as purchasing power parity (PPP) adjusted values.}
\label{tab:vars}
\resizebox{0.5\textwidth}{!}{
\renewcommand{\thinmuskip}{\,}
\begin{tabular}{l rrrrr}
\hline \hline
 \textbf{Variable} &  \textbf{Mean} & \textbf{SD} & \textbf{Median} & \textbf{Min} & \textbf{Max} \\
 \hline
\emph{Outcome variable} & & & & & \\
\thinmuskip Relative reduction in HIV infection rate (in \%) & $3.26$ & $7.85$ & $2.94$ & $-50.00$ & $17.30$ \\
\hline
\emph{Treatment variable} & & & & & \\
\thinmuskip Development aid (in USD millions) & $50.121$ & $112.858$ & $6.348$  & $0.003$ & $559.897$ \\
\hline
\emph{Country characteristics} & & & & & \\
\thinmuskip GDP per capita PPP (in USD thousands)  & $8.87$ & $6.79$ & $7.26$ & $0.77$ & $30.45$ \\
\thinmuskip GDP growth (in \%)  & $3.81$ & $3.53$ & $3.82$ & $-5.07$ & $26.68$\\
\thinmuskip Foreign direct investment (in USD billions) & $3.29$ & $8.75$ & $0.89$ & $-7.40$ & $68.89$\\
\thinmuskip Consumer price index - inflation (in \%)  & $7.75$ & $18.72$ & $4.38$ & $-1.54$ & $187.85$ \\
\thinmuskip Unemployment (in \%)  & $7.83$ & $6.23$ & $5.65$  &  $0.14$ &  $27.04$\\
\thinmuskip Population (in millions)  & $41.66$ & $135.75$ & $11.98$ & $0.38$ &  $1338.68$\\
\thinmuskip Fertility (number of births per woman)  & $3.25$ & $1.33$ & $2.85$ & $1.26$ & $7.00$ \\
\thinmuskip Maternal mortality (number of deaths per 100,000 live births)  & $241.21$ & $259.08$ & $144.00$ & $2.00$ & $1150.00$ \\
\thinmuskip Infant mortality (number of deaths per 1,000 live births)  & $30.95$ & $20.36$ & $26.60$ & $2.70$ &  $87.30$\\
\thinmuskip Life expectancy (in years)  & $68.50$ & $6.90$ & $69.51$ & $52.95$ & $79.91$ \\
\thinmuskip School enrolment ratio (primary education)  & $1.04$ & $0.13$ & $1.03$ & $0.68$ & $1.45$ \\
\thinmuskip Prevalence of undernourishment (in \% of population)  & $13.91$ & $11.65$ & $9.10$ & $2.50$ & $58.70$ \\
\thinmuskip Access to electricity (in \% of population)  & $73.83$ & $29.83$ & $90.30$ & $4.20$ & $100.00$\\
\thinmuskip Incidence of tuberculosis (number of cases per 100,000 people)  & $162.29$ & $158.96$ & $99.00$ & $5.20$ & $738.00$\\
\hline \hline
\end{tabular}
}
\end{center}
\end{table}
The above data have a small ratio of missing values (around 3.5\%), mostly present in a few country characteristics (e.g., school enrollment, inflation). We imputed the missing values using the $k$-nearest neighbors algorithm. Furthermore, due to varying scales in country characteristics, we standardize the covariate data prior to the analysis using a min-max scaler (i.e., scale the data to $[0, 1]$ interval by deducting the minimal value and, subsequently, dividing by the difference between the maximal and the minimal value). 

\section{Semi-synthetic data}
\label{supp:semisynthetic_data}


In order to assess the effectiveness of our \framework in predicting treatment--response curves, we need to evaluate prediction performance on counterfactual outcomes (and not only on factual outcomes). Because counterfactual outcomes are unobservable in the real-world data, we follow common practice in machine learning \cite{Schwab2020, Bica2020} and make use of semi-synthetic data where ``ground-truth'' counterfactual outcomes are known. Formally, we simulate semi-synthetic data where covariates, treatment, and relationships between variables are based on the real-world data in order to reflect our HIV setting. We simulate: (i)~semi-synthetic data with simulated outcomes based on real-world HIV data in year 2016 (used for learning); and (ii)~semi-synthetic data with simulated outcomes based on real-world HIV data in year 2017 (used for evaluation).

Formally, we proceed as follows. First, using real-world HIV data in year 2016, we compute the parameters in a linear regression of relative reduction in HIV infection rate (our outcome) on development aid (our treatment) and country characteristics (our covariates), including interaction terms between development aid and country characteristics. The estimated parameters (intercept $\hat{c}$, treatment effect $\hat{\alpha}$, covariates effects $\hat{\beta}_1, \ldots, \hat{\beta}_p $, and interaction effects $\hat{\gamma}_1, \ldots, \hat{\gamma}_p $) are used as ground-truth effects. We then leverage the respective ground-truth effects to compute the ``pseudo-mean'' of an outcome for each country $i=1, \ldots, n$ via 
\begin{equation}
\label{eq:gt_efffect}
    \mu_i = \hat{c} + \hat{\alpha} \, a_i +  \sum_{j=1}^p \big( \hat{\beta}_j \, x_{ij} + \hat{\gamma}_j \, a_{ij} \,  x_{ij} \big) .
\end{equation}
We normalize the pseudo-means $\{ \mu_i \}_{i=1}^n$ using a min-max scaler and thereby obtain $ \{ \hat{\mu}_i  \}_{i=1}^n $. Afterwards, we induce a non-linearity by taking the square root, i.e., $\tilde{\mu}_i = \sqrt{\hat{\mu}_i}$. Finally, we simulate the outcomes via  
\begin{equation}
    \tilde{y}_i = \tilde{\mu}_i + \epsilon_i ,
\end{equation}
where $\epsilon_i \sim N(0, 0.01^2)$ is Gaussian noise. 

Depending on whether we use the semi-synthetic data for learning or evaluation, we make the following adaptations. For semi-synthetic data (i) that we use for learning, we simulate 105 data points (for 105 countries with respective covariates/country characteristics) with factual outcomes that are simulated as above by using the observed volumes of development aid (treatment). For semi-synthetic data (ii) that we use for evaluation, we simulate counterfactual outcomes by using varying volumes of development aid on the observed interval $[A_{\mathrm{min}}, A_{\mathrm{max}}]$ instead of observed treatment $a_i$ in Eq.~\eqref{eq:gt_efffect} and without noise $\epsilon_i$. Because there are infinitely many counterfactual outcomes that constitute a treatment--response curve for a continuous treatment, we simulate outcomes for 64 equally spaced aid volumes on the interval $[A_{\mathrm{min}}, A_{\mathrm{max}}]$, which are then used for evaluation (see below). As a result, the dataset for evaluation counts $64 \cdot 105$ data points. In sum, the aforementioned procedure allows us to simulate ground-truth treatment--response curves that we use to evaluate the performance of our \framework.  

The above procedure for obtaining semi-synthetic data is designed to reflect the real-world setting of allocating aid earmarked to end the HIV epidemic. Importantly, we retain the following main characteristics of the real-world data: (i)~the same covariates; (ii)~the same sample size; (iii)~the same treatment interval; and (iv)~ground-truth effects that are based on estimates of treatment effects from real-world HIV data and thus reflect observed relationships (here, the linear regression coefficients are used as proxies). We add non-linearity as we expect that ground-truth effects of aid in real-world HIV settings are characterized by non-linearities \cite{Jakubik2022}. 

\newpage 


\section{Pseudocode}
\label{supp:pseudocode}

\framework combines three components -- i.e., the balancing autoencoder, the counterfactual generator, and the inference model -- in order to provide predictions of treatment--response curves. The learning algorithm for our \framework is given in Algorithm~\ref{alg:CG-CT}. First, the balancing autoencoder is applied to the observational data $\mathcal{D} =  \{(y_i, a_i, x_i)\}_{i=1}^{n}$ to obtain $\hat{\mathcal{D}} =  \{(y_i, a_i, \hat{z}_i)\}_{i=1}^{n}$. Then, by using the counterfactual generator with $\hat{\mathcal{D}}$, we generate additional $m$ outcome--treatment pairs for each $\hat{z}_i$, thus adding $m \cdot n$ new data points that represent counterfactual outcomes (i.e., the synthetic twins). This provides the final data that is used for predicting treatment--response curves, i.e.,  $\tilde{\mathcal{D}} = \big \{(y_i, a_i, \hat{z}_i) \cup (\{ \tilde{y}_{ij}, \tilde{a}_{ij} \}_{j=1}^m , \hat{z}_i) \big \}_{i=1}^{n}$. In the third step, we use the inference model (i.e., GPS) to predict treatment--response curves from data $\tilde{\mathcal{D}}$. For computational reasons, Algorithm~\ref{alg:CG-CT} makes use of batch learning for a given number of epochs.


\begin{algorithm}[htbp]
\caption{Learning algorithm for \framework}
\label{alg:CG-CT}

\vspace{0.25cm}
\begin{algorithmic}[1]
\scriptsize
\Require{Data $\mathcal{D} =  \{(y_i, a_i, x_i)\}_{i=1}^{n} $; loss functions $\mathcal{L}$, $\mathcal{L}_x$, and $ \mathcal{L}_a$; hyperparameters $\eta$, $\theta$, $r$, $m$, $\alpha$, and $s$; network architecture with weights $W_{\phi}$, $W_{g_x}$, $W_{g_a}$}
\Ensure{Optimal representation $\phi^*$ with weights $W_{\phi}^*$, and estimated weights of the GPS model $\hat{\beta}_{\mathrm{GPS}}$, $\hat{\alpha}_{\mathrm{GPS}}$, and $\hat{\sigma}_{\mathrm{GPS}}$} 
\vspace{0.3cm}
\Repeat  \Comment{\emph{Component 1: Balancing autoencoder}}
\State Randomly sample mini-batch of size $b$ from  $\mathcal{D}$
\State \mbox{Compute $g_1 = \nabla_{W_{\phi}}  \frac{1}{b} \sum \limits_{i=1}^b \mathcal{L}_x(g_x(\phi(x_i)),x_i) $}
\State \mbox{Compute $g_2 = \nabla_{W_{\phi}}  \frac{1}{b} \sum \limits_{i=1}^b \mathcal{L}_a(g_a(\phi(x_i)),a_i) $}
\State \mbox{Compute $g_3 = \nabla_{W_{g_x}}  \frac{1}{b} \sum \limits_{i=1}^b \mathcal{L}_x(g_x(\phi(x_i)),x_i) $}
\State \mbox{Compute $g_4 = \nabla_{W_{g_a}}  \frac{1}{b} \sum \limits_{i=1}^b \mathcal{L}_a(g_a(\phi(x_i)),a_i) $}
\State Update weights 
\vspace{-0.2cm}
\begin{align*}
W_{\phi} &\leftarrow W_{\phi} - \eta \, (g_1 - \theta g_2) , \\
W_{g_x} &\leftarrow W_{g_x} - \eta \, g_3 , \\
W_{g_a} &\leftarrow W_{g_a} - \eta \, \theta g_4 
\end{align*}
\vspace{-0.3cm}
\Until convergence
\State Output the optimal representation $\phi^*$ with weights $W_{\phi}^*$
\State Use $W_{\phi}^*$ to embed covariates in $\mathcal{D}$, and obtain $\hat{\mathcal{D}} =  \{(y_i, a_i, \hat{z}_i)\}_{i=1}^{n}$ with $\hat{z}_i = \phi^*(x_i)$
\vspace{0.3cm}
\For{$(y_i, a_i, \hat{z}_i) \in \hat{\mathcal{D}}$ } \Comment{\emph{Component 2: Counterfactual generator}}
    \State Randomly sample $m$ treatment values $\{ \tilde{a}_{ij} \}_{j=1}^m$ uniformly from the interval $[A_{\mathrm{min}}, A_{\mathrm{max}}]$
    \For{$j = 1, \ldots ,m$ }
        \State Compute $w_{ij}^*$ using Eq.~\eqref{eq:w_optim} for $x = \hat{z}_i$ and $\tilde{a} = \tilde{a}_{ij}$
        \State Compute $\tilde{y}_{ij} = Y_{\mathrm{vec}}^T \, w_{ij}^*$
    \EndFor
    \State \textbf{return} $\{ ( \tilde{y}_{ij}, \tilde{a}_{ij}) \}_{j=1}^m$
\EndFor
\State \textbf{return} counterfactual outcomes $
\big \{ (\{ \tilde{y}_{ij}, \tilde{a}_{ij} \}_{j=1}^m , \hat{z}_i) \big \}_{i=1}^{n}$
\State Add $
\big \{ (\{ \tilde{y}_{ij}, \tilde{a}_{ij} \}_{j=1}^m , \hat{z}_i) \big \}_{i=1}^{n}$ to $\hat{\mathcal{D}}$, and obtain $\tilde{\mathcal{D}} = \hat{\mathcal{D}} \cup \big \{ (\{ \tilde{y}_{ij}, \tilde{a}_{ij} \}_{j=1}^m , \hat{z}_i) \big \}_{i=1}^{n}$
\vspace{0.3cm}
\State Estimate the GPS model on $\tilde{\mathcal{D}}$ as in \cite{Hirano2004} \Comment{\emph{Component 3: Inference model}}
\State Output the estimated coefficients of the GPS model: $\hat{\beta}_{\mathrm{GPS}}$, $\hat{\sigma}_{\mathrm{GPS}}$, and $\hat{\alpha}_{\mathrm{GPS}}$
\end{algorithmic}
\end{algorithm}

\newpage


\section{Background on modeling aid effectiveness} 
\label{supp:lit_aid_effect} 

Extensive work has focused on analyzing the effectiveness of development aid on macroeconomic growth \cite{Lesink2000, Hansen2000, Doucouliagos2009}. Even other works study the link between development aid on specific SDG outcomes. For example, the authors in \cite{Birchler2016} examine the effect of education aid on primary education using generalized method of moments regression. \cite{Munyanyi2020} estimate the effectiveness of development aid on 8 SDG outcomes (i.e., poverty ratio, extreme hunger, school enrollment, HIV prevalence, gender parity at school, access to water, child mortality, and maternal mortality) using a panel fixed effects regression. Yet, as we detail below, a novel methodological approach is needed for our task. 

Previously, different methodological approaches have been applied in order to study the effectiveness of development aid. Several works draw upon field experiments (e.g., \cite{Duflo2003, Duflo2006}). Here, the objective is to provide confirmatory evidence whether there is indeed a causal effect (and of what magnitude) and thus to understand whether aid was generally effective. However, experiments involving development aid are costly and limited to micro-level analyses, that is, the sample is designed for a small geographic area and not across multiple countries, as needed for our task. Furthermore, standard randomized controlled trials with two treatment arms would be limited to binary treatments and not treatment-response curves and, without additional assumptions, estimates are limited to \emph{average} effects and \underline{not} \emph{heterogeneous} effects. 

Other works use observational data (e.g., \cite{Birchler2016,Munyanyi2020}). Yet, these works typically make strong modeling assumptions. First, they assume that the effect is linear in the aid volume. Second, the true treatment effect is typically underestimated due to the nature of the modeling approach, and, hence, the estimate is merely a lower bound. Third, the aforementioned works assume that the effectiveness of aid cannot vary across countries but that the effectiveness is instead identical for all countries. As such, these works estimate the \emph{average} effect of aid and \underline{not} the \emph{heterogeneous} effect of aid. To address this, we develop a tailored method for predicting between-country heterogeneity in treatment--response curves. Importantly, we use machine learning to avoid overfitting and thus to generate predictions that generalize well across countries.  

To support decision-making, one needs to answer the question \emph{``if volume $a$ is spent on development aid, what would be the most likely outcome on the SDGs?''}. To the best of our knowledge, one work \cite{Jakubik2022} uses traditional machine learning for that purpose, where the authors combine LASSO regression with mathematical optimization to inform the optimal allocation of development aid. We have included the method in \cite{Jakubik2022} as a baseline. Throughout our manuscript, we refer to it as linear model (LM) where we vary order and regularization term (i.e., linear regression, LASSO regression, or ridge regression). Different from the LM, our framework is designed as a more flexible, non-linear approach. On top of that, the LM suffers from an important limitation: it is based on traditional machine learning, and not causal machine learning. In particular, this method does not account for treatment selection bias, and, as a result, it may give unreliable predictions of treatment effect. 

To address the above limitations, a tailored machine learning framework is needed in order to provide more reliable predictions of heterogeneous aid--response curves. Across all of our experiments, we find robust evidence that this baseline is outperformed by our proposed machine learning framework. Thereby, we address that the question \emph{``if volume $a$ is spent on development aid, what would be the most likely outcome on the SDGs?''} is inherently linked to counterfactual inference where hypothetical questions are answered with regard to which outcome to expect. In order to generalize well across countries, our framework further makes extensive use of recent innovations in statistics and machine learning.

\section{Comparison with existing methods for treatment effect estimation} 
\label{supp:lit_methods} 

Here, we briefly review existing methods for treatment effect estimation and point to salient differences between them and our \framework. 

\vspace{0.25cm}
\noindent
\underline{Methods for binary treatments:} Prior literature on counterfactual inference makes nowadays increasing use of machine learning. This then allows to predict heterogeneous treatment effects. Here, prior literature has been strongly focused on the setting with binary treatments \cite{Johansson2016, Shalit2017, Yoon2018, Athey2019, Hatt2021}. However, this is different from our work where we have continuous treatments.  

\vspace{0.25cm}
\noindent
\underline{Methods for continuous treatments:} There are comparatively few methods that consider continuous treatments \cite{Hirano2004, Schwab2020, Bica2020}. Here, key methods are as follows: 
\begin{itemize}
\item \emph{Generalized propensity score (GPS)} \cite{Hirano2004}. The GPS is a two-step estimation procedure. In the first step, conditional distribution of treatment given covariates is modeled under Gaussian distribution assumption. The estimated parameters are then used to compute the estimate for GPS. In the second step, conditional distribution of outcome given treatment and GPS is modeled as 2nd-degree polynomial regression with an interaction term between treatment and GPS.
\item \emph{Dose-response network (DRNet)} \cite{Schwab2020}. To predict heterogeneous treatment--response curves, DRNet provides a multi-layer neural network with a shared representation, multiple treatment ``heads'' (for each possible treatment), and multiple dosage ``heads'' within each treatment ``head'' (for splitting respective dosage interval of each treatment). The aim of the architecture with multiple treatment and dosage ``heads'' is to avoid treatment and dosage information being lost within representation layers of the neural network.
\item \emph{SCIGAN} \cite{Bica2020}. SCIGAN is a method based on generative adversarial network (GAN). SCIGAN is motivated by the work of \cite{Yoon2018} where counterfactual generator is used to generate potential outcomes of different treatment assignments. In particular, SCIGAN uses the generator which takes observed data (i.e., observed outcome, treatment, dosage, and covariates), random treatment, random dosage, and random noise to generate a potential outcome for a random treatment and dosage (sampled uniformly from given treatments and dosages). The aim of the generator is to produce potential outcomes that `fool' the discriminator which discriminates between outcomes of observed treatment and dosage (i.e., the factual outcome), and outcomes of random treatments and dosages (i.e., counterfactual outcomes). Once the counterfactual generator is trained, an inference network (i.e., an artificial neural network) that predictes heterogeneous treatment--response curves is trained on combined original and generated data.
\item \emph{Methods based on (semiparametric) statistical theory} \cite{Kennedy.2017, Foster.2019, Colangelo.2020, Nie.2021}. There are several methods for obtaining \emph{asymptotic} (large-sample) convergence guarantees and valid confidence intervals. Here, methods roughly fall into two main categories: (i)~methods based on the efficient influence function of the dose-response curve, and (ii)~methods based on Neyman orthogonal losses (double machine learning). In the context of~(i), influence function-based methods aim at correcting for asymptotic plugin bias that occurs from estimation errors in the nuisance functions. Examples include the estimator proposed in \cite{Kennedy.2017}, which performs a pseudo-outcome regression, and VCNet \cite{Nie.2021}, which employs a functional targeted regularization. However, existing methods from (i) only consider \emph{average} but \underline{not} \emph{individualized} dose-response functions. In the context of~(ii), double machine learning-based methods such as \cite{Foster.2019} and \cite{Colangelo.2020} construct Neyman orthogonal loss functions, yet which are locally insensitive to small estimation errors in the nuisance functions. These methods rely on sample splitting to guarantee fast (asymptotic) convergence rates \cite{Chernozhukov.2018}. Due to these characteristics, we found them not to be effective for our work and instead opt for our machine learning framework called \framework.  
\end{itemize}
Note that both DRNet and SCIGAN are designed for multiple treatments and a dosage interval for each treatment. However, since we only have one treatment (i.e., development aid), we have implemented DRNet without multiple treatment ``heads'', and SCIGAN without treatment discriminator.

\vspace{0.25cm}
\noindent
\underline{Differences to our work:} In comparison with the above methods, our \framework offers three key advantages in our HIV setting. (1)~Our balancing autoencoder uses adversarial learning to learn a representation that is not predictive of the treatment (thereby addressing treatment selection bias) while simultaneously reducing the dimension of covariate space that we use to control for confounding. Removing selection bias is especially important in our high-dimensional \emph{and} low sample setting, which is not accounted for by other baselines. In particular, semiparametric theory-based methods rely on sample splitting to provide guarantees for \emph{asymptotic} convergence rates \cite{Chernozhukov.2018, Foster.2019}, but sample splitting may harm the estimation performance in low sample settings \cite{Curth.2021}. (2)~SCIGAN as a GAN-based method requires a comparatively large data sample to provide a good fit. This is evidenced by poor performance of SCIGAN in our HIV setting. Hence, our novel counterfactual generator offers a better solution for generating counterfactual outcomes in small sample settings. (3)~Our choice of the inference model to be GPS over more complex models (e.g., DRNet) has proven to be more robust in a small sample size setting. This is expected given that DRNet also has a rather large model architecture because of shared representation and multiple dosage ``heads'' that require more data for a good fit. 

\section{Hyperparameter tuning}
\label{supp:hyperpars} 

To ensure a fair comparison, we use the same training procedure and hyperparameter tuning (where applicable) for both \framework and the baselines. In particular, the GPS baseline is identical to the GPS inside our \framework; that is, they both build upon the same polynomial regression. Hence, this has an important implication: all performance improvements of \framework over GPS must be attributed to the fact that our \framework handles the data in a better way (i.e., by addressing high-dimensional covariates and treatment selection bias). This is further confirmed empirically in our ablation studies (see Supplement L below), where we combine the balancing autoencoder and the counterfactual generator with other inference models. Therein, we demonstrate that our \framework achieves consistent performance gains across all inference models.

Hyperparameters are tuned via cross-validation with a 80/20 split. Table~\ref{tab:baseline_hyp} shows the list of hyperparameters of both our \framework and the baselines. The hyperparameters and ranges for the baselines (i.e., a linear model (LM) and an artificial neural network (ANN)) are standard. For dose-response network (DRNet) and SCIGAN, we have used similar hyperparameters as proposed in \cite{Schwab2020, Bica2020}; however, we adjusted some of the ranges to accommodate the much smaller sample size in our study (i.e., the layer size and batch size are adjusted based on the number of observations $n$). Also, we vary additional hyperparameters such as learning rate and dropout rate. 

Importantly, when implementing \framework with baselines as inference models (in our main analysis and in supplements), we use the same hyperparameter tuning for the inference model as for the respective baseline. This ensures fair comparison between the \framework and the baselines. 



\begin{table}[H]
\begin{center}
\caption{Hyperparameters and search ranges for \framework and baselines.}
\label{tab:baseline_hyp}
\resizebox{0.45\textwidth}{!}{
\footnotesize
\begin{tabular}{l l l }
\hline \hline
 \textbf{Method} &  \textbf{Hyperparameter} & \textbf{Search range} \\
    \hline
 \multirow{2}{*}{Linear model (LM) } & Order & $0, 1, 2$ \\ 
 & Regularization & $0.05, 0.1, 0.5, 1, 5$ \\
 \hline 
  \multirow{5}{*}{Artificial neural network (ANN) } & Layer size & $ 53, 27, 14$ \\ 
 & Learning rate & $0.001, 0.0005, 0.0001$ \\
  & Dropout rate & $0, 0.1, 0.2$ \\
    & Number of epochs & $ 100, 200, 300$ \\
  & Batch size & $22, 11, 6$ \\
 \hline 
\multirow{1}{*}{Generalized propensity score (GPS) \cite{Hirano2004} } & --- & --- \\
   \hline
  \multirow{7}{*}{Dose--response network (DRNet) \cite{Schwab2020}} & Layer size & $ 53, 27, 14$ \\ 
   & Representation layer size & $22, 11, 6$ \\
 & Learning rate & $0.001, 0.0005, 0.0001$ \\
  & Dropout rate & $0, 0.1, 0.2$ \\
    & Number of epochs & $ 100, 200, 300$ \\
  & Batch size & $22, 11, 6$ \\
    & Number of dosage intervals (E) & $5$ \\
     \hline
  \multirow{7}{*}{SCIGAN \cite{Bica2020}} & Layer size & $ 53, 27, 14$ \\ 
 & Learning rate & $0.001, 0.0005, 0.0001$ \\
  & Dropout rate & $0, 0.1, 0.2$ \\
    & Number of epochs & $ 100, 200, 300$ \\
  & Batch size & $22, 11, 6$ \\
      & Number of dosage samples & $3, 5, 7$ \\
    & Factual loss trade-off ($\lambda$) & $1$ \\
 \hline
   \multirow{10}{*}{\framework } & Layer size & $ 14, 10, 7$ \\ 
    & Representation layer size & $ 10, 7, 4$ \\
    & Learning rate & $0.001, 0.0005, 0.0001$ \\
    & Dropout rate & $0, 0.1, 0.2$ \\
    & Number of epochs & $ 100, 200, 300$ \\
    & Batch size & $22, 11, 6$ \\
    & Balancing parameter ($\theta$) & $0.05, 0.1, 0.5, 1, 5$ \\
    & Regularization penalty ($\alpha$) & $0.05, 0.1, 0.5, 1, 5$ \\
    & Number of dosage samples ($m$) & $3, 5, 7$ \\
\hline \hline
\end{tabular}
}
\end{center}
\end{table}

\section{Covariate distributions for different development aid volumes}
\label{supp:covariate_shift_analysis} 

Here, we analyze the distribution of covariates (i.e., country characteristics that we control for) for different volumes of development aid in order to asses the presence of treatment-dependent covariate shifts in the observational data. Given that our treatment variable is continuous, we split our data in three groups (low, medium, and high) by using the empirical distribution of the observed development aid and the respective quantiles such that each group has the same probability mass. Hence, the group with (i)~\emph{low} volume of development aid contains all observations where development aid was below the $\frac{1}{3}-$quantile; the group with (ii)~\emph{medium} volume of development aid contains all observations where development aid was between the $\frac{1}{3}-$quantile and the $\frac{2}{3}-$quantile; and the group with (iii)~\emph{high} volume of development aid contains all observations where development aid was above the $\frac{2}{3}-$quantile. 

Table~\ref{tab:cov_shift} shows the mean and the standard deviation for each covariate across three treatment groups, i.e., low, medium, and high volume of development aid. We observe that treatment-dependent covariate shifts are present in the observational data, particularly among wealth indicators (e.g., GDP per capita, access to electricity) and health indicators (e.g., maternal mortality, infant mortality, life expectancy). Therefore, addressing the treatment selection bias through, e.g., the use of our balancing autoencoder, is crucial to ensure a reliable predictions of the heterogeneous treatment effect of development aid.

\begin{table}[H]
\begin{center}
\caption{Analysis of covariate distributions for different volumes of development aid. Reported are the summary statistics for the covariates in terms of the mean and the standard deviation (SD: standard deviation). The summary statistics are arranged across three treatment groups depending on the volume of development aid, i.e., low, medium, and high.}
\label{tab:cov_shift}
\resizebox{0.5\textwidth}{!}{
\renewcommand{\thinmuskip}{\,}
\begin{tabular}{l ccc}
\hline \hline
\textbf{Treatment group} & \textbf{Low} & \textbf{Medium} & \textbf{High} \\
\thinmuskip Development aid range (in USD millions) &   $[0.003-2.649]$ &  $[2.649-15.301]$ &   $[15.301-559.897]$ \\
\hline
\textbf{Covariates} & \textbf{Mean} $\pm$ \textbf{SD} & \textbf{Mean} $\pm$ \textbf{SD} & \textbf{Mean} $\pm$ \textbf{SD} \\
\thinmuskip GDP per capita PPP (in USD thousands)  & $14.13 \pm 6.60$ & $7.34 \pm 5.83$ & $5.15 \pm 4.27$ \\
\thinmuskip GDP growth (in \%)  & $3.69 \pm 4.77$ & $3.70 \pm 2.73$ & $4.05 \pm 2.79$ \\
\thinmuskip Foreign direct investment (in USD billions) & $3.40 \pm 6.26$ & $3.57 \pm 11.57$ & $2.90 \pm 7.78$ \\
\thinmuskip Consumer price index - inflation (in \%)  & $5.34 \pm 5.84$ & $5.20 \pm 5.68$ & $12.71 \pm 31.11$ \\
\thinmuskip Unemployment (in \%)  & $8.76 \pm 5.85$ & $7.23 \pm 5.48$ & $7.50 \pm 7.28$ \\
\thinmuskip Population (in millions)  & $18.98 \pm 28.40$ & $29.39 \pm 49.55$ & $76.61 \pm 226.20$ \\
\thinmuskip Fertility (number of births per woman)  & $2.46 \pm 0.76$ & $3.38 \pm 1.45$ & $3.92 \pm 1.27$ \\
\thinmuskip Maternal mortality (number of deaths per 100,000 live births)  & $80.77 \pm 72.43$ & $263.71 \pm 285.31$ & $379.14 \pm 268.43$ \\
\thinmuskip Infant mortality (number of deaths per 1,000 live births)  &  $17.91 \pm 12.54$ & $30.89 \pm 20.86$ & $44.04 \pm 18.01$ \\
\thinmuskip Life expectancy (in years)  & $73.49 \pm 4.25$ & $68.74 \pm 6.52$ & $63.25 \pm 5.59$ \\
\thinmuskip School enrolment ratio (primary education)  & $1.02 \pm 0.09$ & $0.99 \pm 0.12$ & $1.09 \pm 0.15$ \\
\thinmuskip Prevalence of undernourishment (in \% of population)  & $9.13 \pm 8.43$ & $14.64 \pm 13.13$ & $17.94 \pm 11.45$ \\
\thinmuskip Access to electricity (in \% of population)  & $94.55 \pm 9.58$ & $74.32 \pm 29.98$ & $52.62 \pm 28.76$ \\
\thinmuskip Incidence of tuberculosis (number of cases per 100,000 people)  & $92.58 \pm 133.42$ & $133.27 \pm 122.45$ & $261.03 \pm 169.08$ \\
\hline \hline
\end{tabular}
}
\end{center}
\end{table}

\newpage 

\section{Visualization of balancing representation}
\label{supp:balance_rep} 

We now provide explanatory insights into how the balancing representation in our \framework works. To this end, we plot the covariates for the country characteristics with vs. without balancing representation (i.e, with vs. without applying the balancing autoencoder in \framework). 

\begin{figure}[H]

\begin{minipage}[b]{0.45\textwidth}

{\raggedright\figletter{a}}
\vspace{0.1cm}

\centering
    \includegraphics[width=\textwidth]{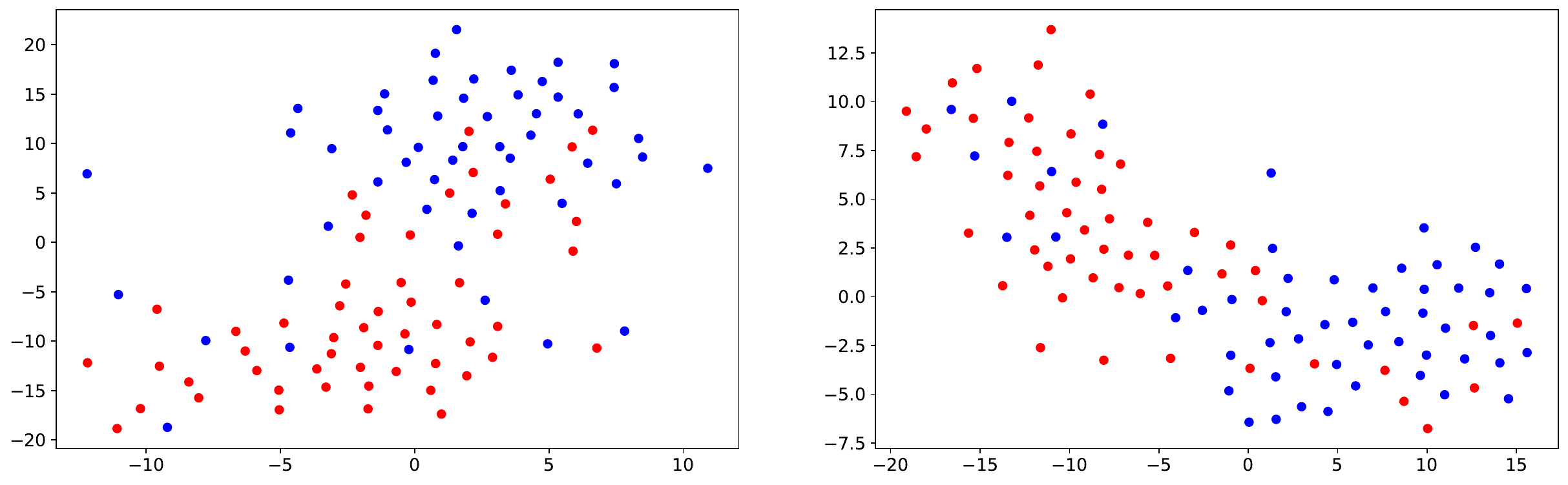}
    \label{fig:tsne}
\end{minipage}

\begin{minipage}[b]{0.45\textwidth}
 
{\raggedright\figletter{b}}
\vspace{0.1cm}

\centering
    \includegraphics[width=\textwidth]{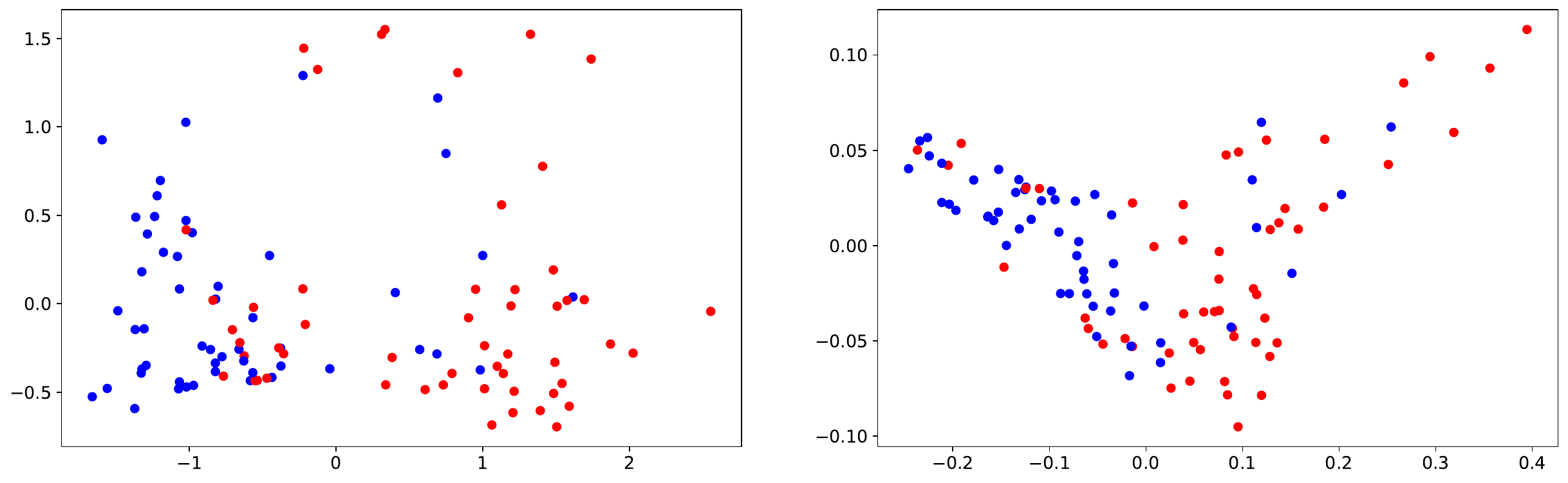}
    \label{fig:isomap}
\end{minipage}
\caption{Illustration of balancing representations for country characteristics (covariates). We visualize the representations of covariates in 2D across different treatment groups, which we color in red and blue (by splitting the original treatment according to whether the value is above or below the median). On the left, we show original covariates and, on the right, the balanced representation of covariates obtained using the balancing autoencoder. In \figletter{a}, we use t-SNE, and, in \figletter{b}, we use isomap for visualization. }
\label{fig:bal_rep}
\end{figure}

Figure~\ref{fig:bal_rep} shows 2D representations of covariates using t-SNE \cite{Maaten2008} (Figure~\ref{fig:bal_rep}a) and isomap \cite{Tanenbaum2000} (Figure~\ref{fig:bal_rep}b). We illustrate representations in 2D for the original covariates (left) and for the balanced representation of covariates learned by the balancing autoencoder (right). We examine whether the balancing autoencoder induces treatment balance in the covariate representations. For this, we split the data in two treatment groups by median (shown in Figure~\ref{fig:bal_rep} as red and blue). 

Overall, we observe that the balancing autoencoder produces more homogeneous covariate representation with respect to treatment as compared to original covariates, thereby successfully addressing treatment selection bias.

\newpage

\section{Examining the validity of underlying assumptions}
\label{supp:assump_robust} 

Here, we conduct additional analyses to examine the validity of the underlying assumptions for identifiability of treatment effects, namely, positivity and ignorability.

\begin{figure}[H]
 \centering
 \includegraphics[width=1\linewidth]{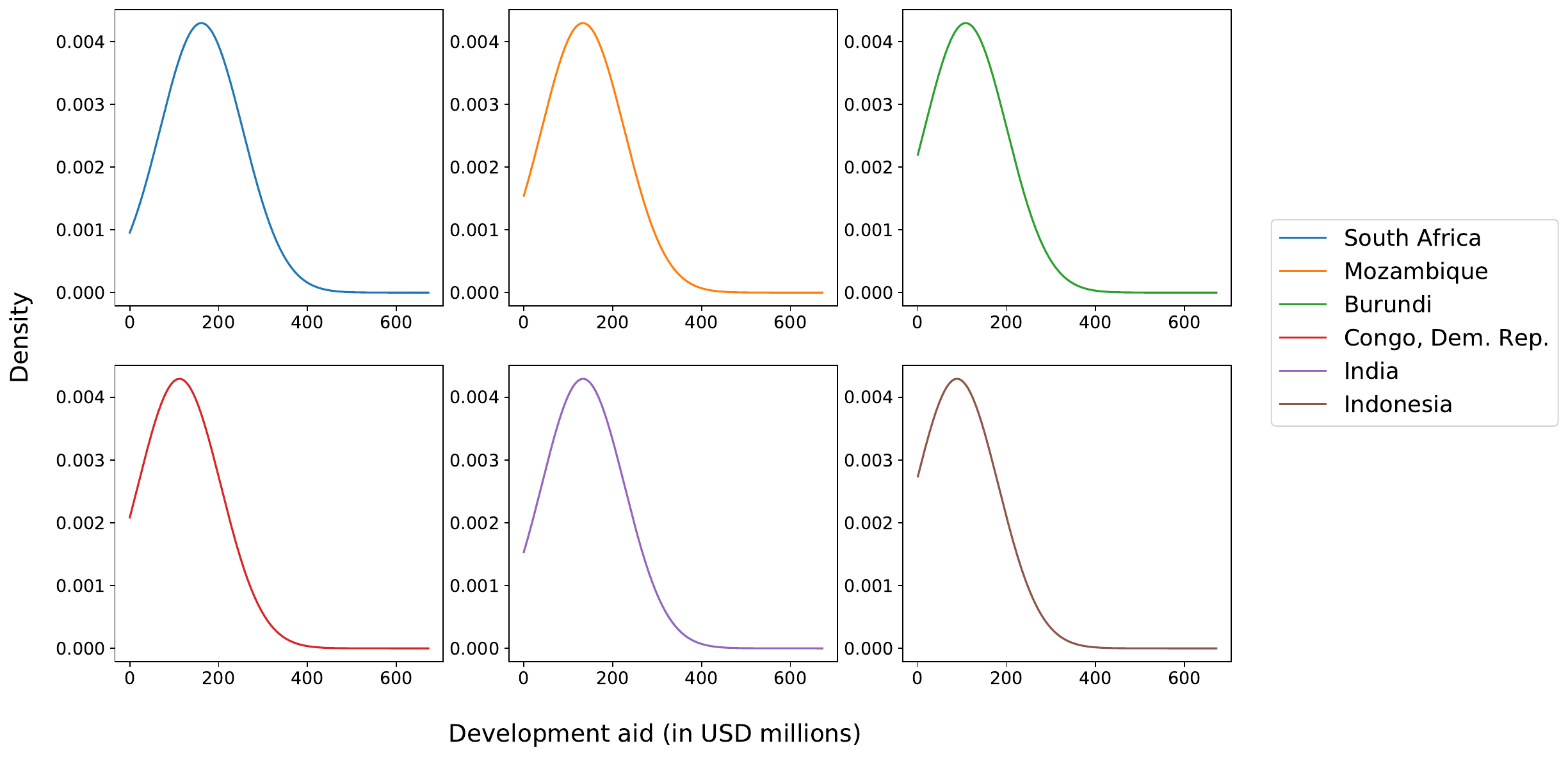}
\caption{Estimated probability curves of development aid given country characteristics for the six example countries as in the results of the main paper: South Africa, Mozambique, Burundi, Congo, India, and Indonesia. The $x$-axis is set to range $[0, A_\text{max}+\hat{\sigma}_{A}]$, where $A_\text{max}$ is the maximal observed development aid and $\hat{\sigma}_{A}$ is the estimated standard deviation of development aid in 2017.}
\label{fig:positivity}
\end{figure}

\vspace{0.25cm}
\noindent
\underline{Positivity assumption:} The positivity assumption states that, for any covariates $X = x$, the probability of treatment $A = a$ is positive for every $a \in \mathcal{A}$. Formally, we have, 
\vspace{0.1cm}
\begin{equation}
    0 < p(A = a \mid X = x) < 1, \forall a \in \mathcal{A}, \mathrm{if} \, \, p(x) > 0. 
\end{equation}
Validating whether the positivity assumption holds is generally impossible in real-world applications. However, we can use observational data to examine whether the probability estimates of a treatment given covariates lie in the range in which we predict the treatment effect, and should thus ensure reliable decisions. In our downstream decision-making problem where we optimize aid allocation, the range of aid volumes that we consider (i.e., the treatment) can vary from zero to $A_{\mathrm{max}}+\hat{\sigma}_{A}$, where $A_{\mathrm{max}}$ is the maximal value of observed aid for a single country, and $\hat{\sigma}_{A}$ is the estimated standard deviation for the aid variable $A$. Now, we examine the estimated probability of development aid over this range for six example countries that we have in the main paper. For this, we use a simple linear regression model to estimate the conditional mean of development aid given covariates and the standard deviation of the error term under a Gaussian distribution. We use the year 2016 for estimating the respective moments of the Gaussian distribution, and the year 2017 for evaluating the estimated probability curves. The estimated probability curves in Fig.~\ref{fig:positivity} show that aid values over the given range are plausible for different countries, meaning that severe extrapolation is unlikely when predicting aid effects over this range. Hence, it is reasonable to expect that the positivity assumption is fulfilled. 

\newpage 

\begin{figure}[H]

\begin{minipage}[b]{0.4\textwidth}
{\raggedright\figletter{a}}
\vspace{0.1cm}

\centering
    \includegraphics[width=\textwidth]{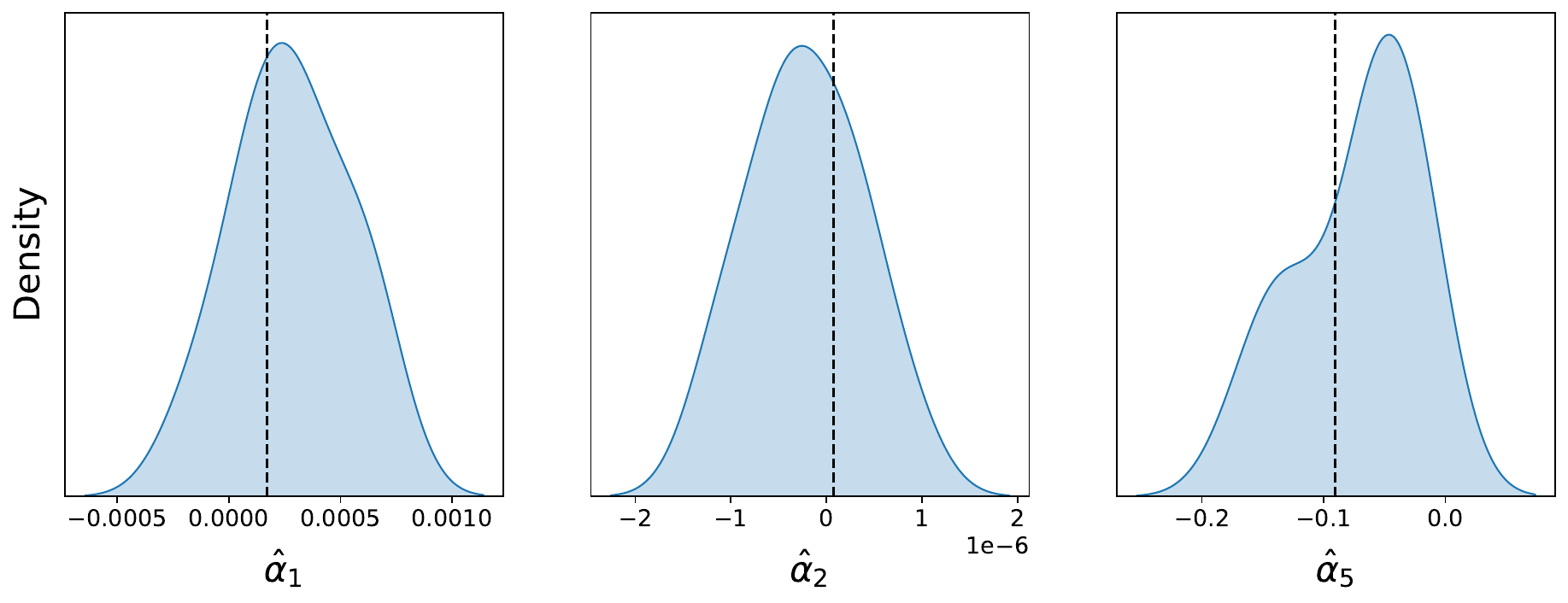}
    \label{fig:robustness_past_aid}
\end{minipage} \hfill

\begin{minipage}[b]{0.4\textwidth}
{\raggedright\figletter{b}}
\vspace{0.1cm}

\centering
    \includegraphics[width=\textwidth]{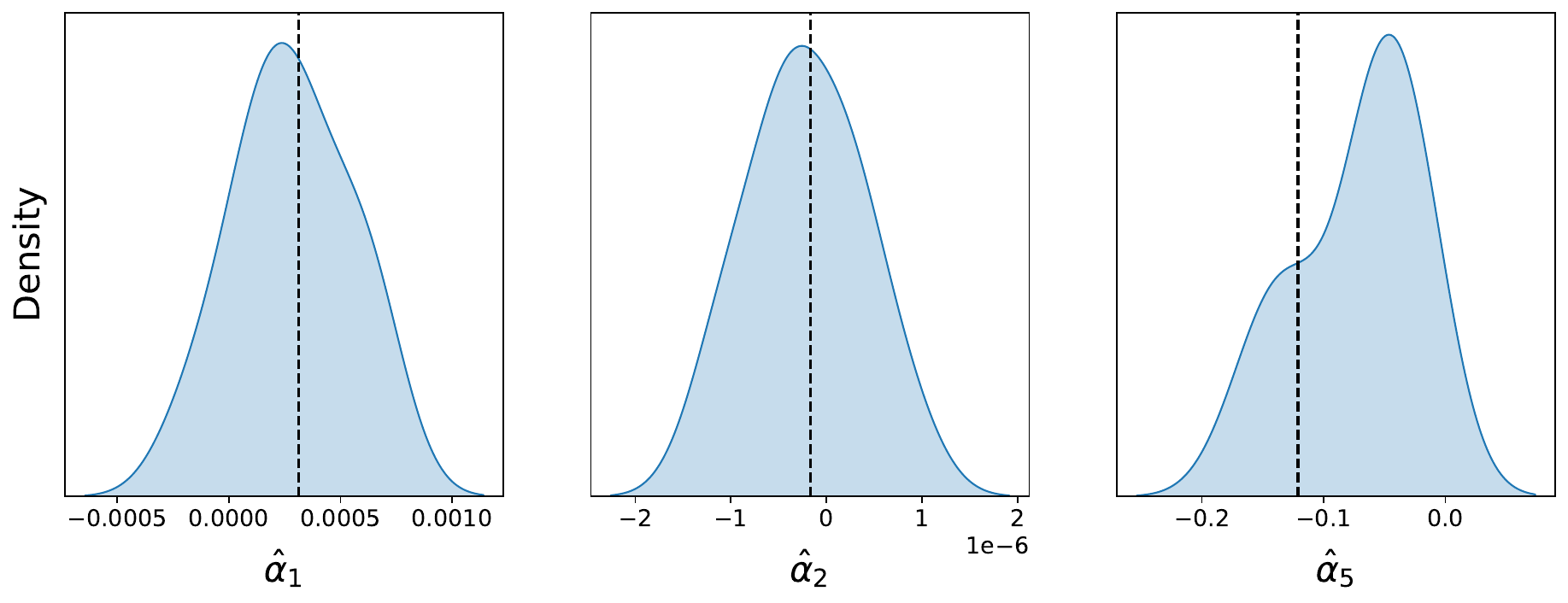}
    \label{fig:robustness_past_aid_neigh}
\end{minipage}

\begin{minipage}[b]{0.4\textwidth}
{\raggedright\figletter{c}}
\vspace{0.1cm}

\centering
    \includegraphics[width=\textwidth]{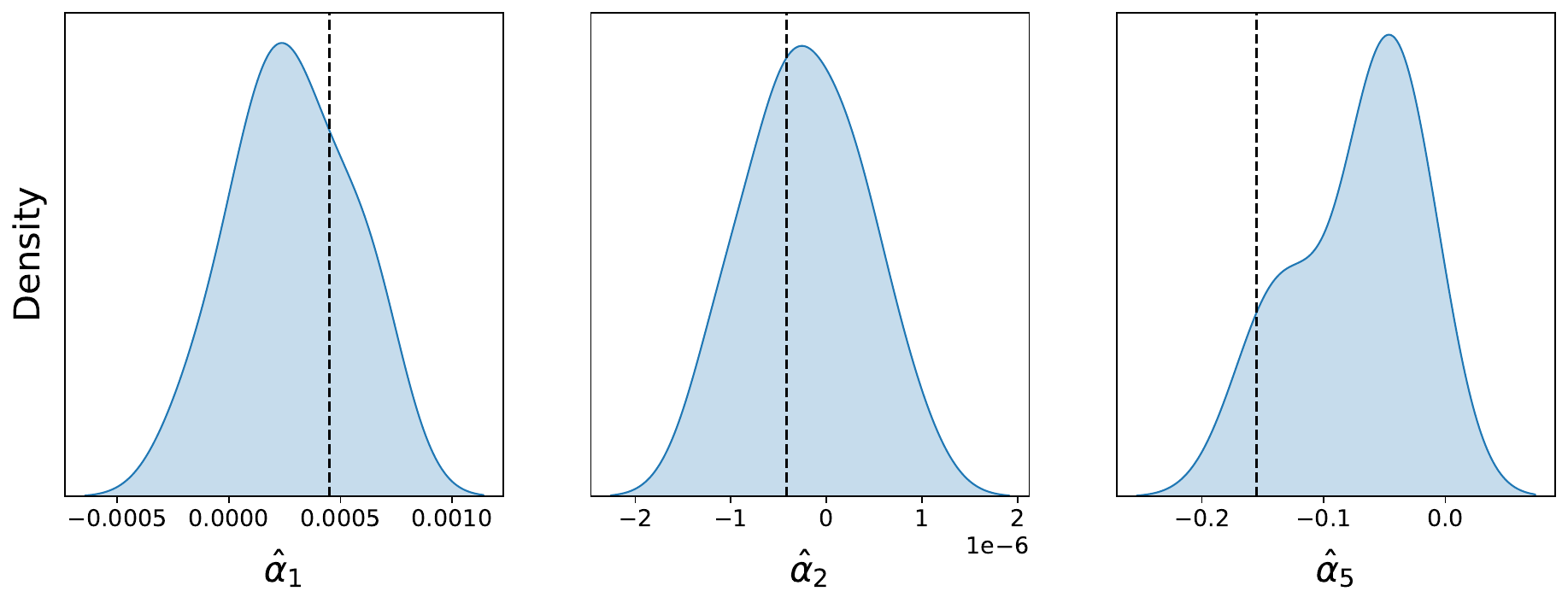}
    \label{fig:robustness_past_hiv}
\end{minipage}

\begin{minipage}[b]{0.4\textwidth}
{\raggedright\figletter{d}}
\vspace{0.1cm}

\centering
    \includegraphics[width=\textwidth]{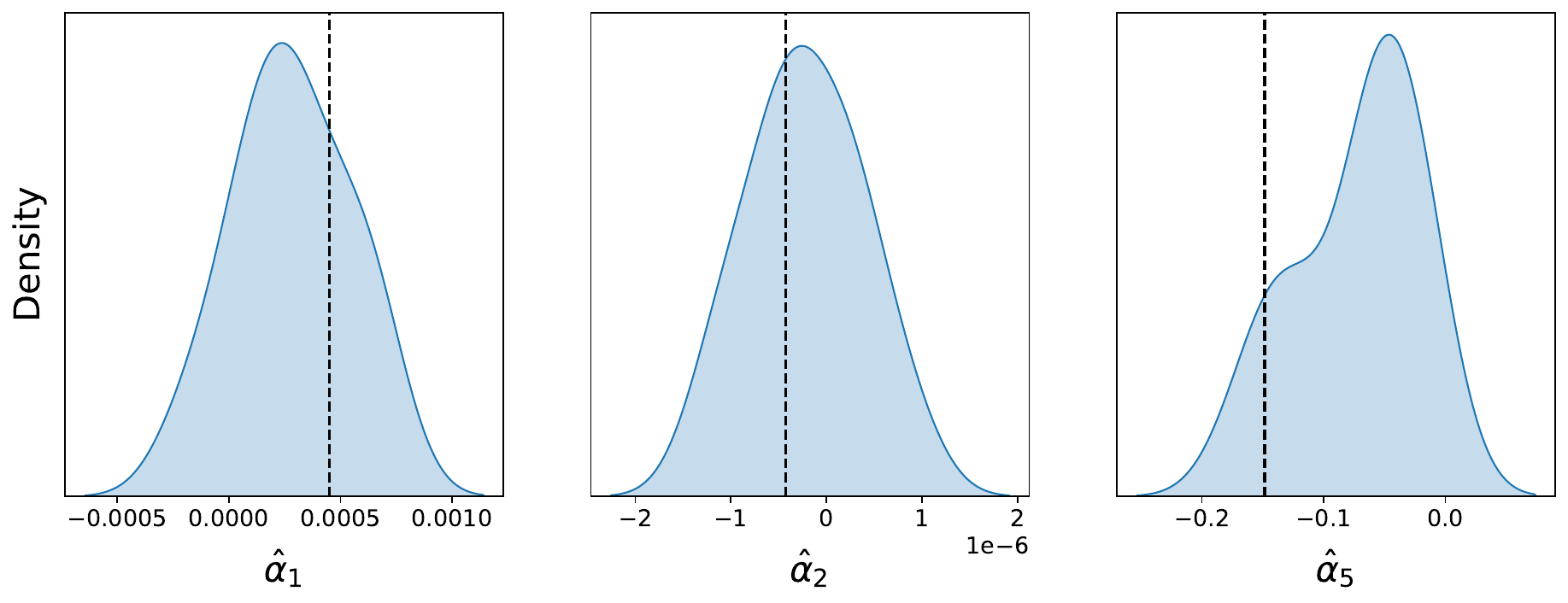}
    \label{fig:robustness_past_hiv_neigh}
\end{minipage}

\caption{Robustness of predicted aid effect when adding additional control variables. Shown are the estimated density of aid effect coefficients $\hat{\alpha_1}, \hat{\alpha_2}$ and $\hat{\alpha_5}$ from the main analysis (i.e., when using only country characteristics as covariates) over 10 runs (blue area), and the observed value of the respective coefficient when adding an additional control variable (dashed black line). Here, we compare: \figletter{a}~adding past aid volume of a country,  \figletter{b}~adding mean past aid volume of neighbor countries,  \figletter{c}~adding past HIV infection rate of a country, and \figletter{d}~adding mean past HIV infection rate of neighbor countries. }
\label{fig:ignorability}
\end{figure}

\noindent
\underline{Ignorability assumption:} The ignorability assumption states that potential outcomes $Y(a)$ are independent of the treatment variable $A$ given covariates $X$, for every $a \in \mathcal{A}$. In other words, all variables that affect both treatment $A$ and potential outcomes $Y(a)$ are measured as part of the covariates $X$. Therefore, ignorability is also referred to as ``no hidden confounders'' assumption. Formally, we have 
\begin{equation*}
    Y(a) \perp \!\!\! \perp A \mid X = x \, \, \forall a \in \mathcal{A} .
\end{equation*}
Validating whether ignorability assumption holds is generally impossible in practice. This is a widespread problem in causal inference when predicting treatment effects from observational data. However, we can examine the robustness of the treatment effect estimate when adding additional variables regarding country characteristics in order to observe whether these variables are potentially hidden confounders or whether the results are robust. Specifically, we examine whether the aid effect coefficients $\alpha_1, \alpha_2$, and $\alpha_5$ from the final stage of the GPS model change when a variable is added to country characteristics. To this end, we first estimate the probability density of estimates of these coefficients (i.e., $\hat{\alpha_1}, \hat{\alpha_2}$, and $\hat{\alpha_5}$) from our main analysis (i.e., when using only country characteristics as covariates) by performing the estimation over 10 runs. Then, we estimate the same coefficient when we add a new variable to the country characteristics and examine whether the estimated coefficients that represent the aid effect changes substantially. Here, we perform the analysis with the following additional variables: (i)~past aid volume of a country from a year before; (ii)~mean past aid volume of neighbor countries from a year before; (iii)~past HIV infection rate in a country from a year before; and (iv)~mean past HIV infection rate of neighbor countries from a year before. Thereby, we control for potential temporal dynamics and/or spillover effects across countries, which may be additionally relevant. The results in Fig.~\ref{fig:ignorability} show that predicted aid effect remain robust when adding any of these four covariates. This demonstrates that our set of country characteristics is sufficient to address potential confounding effects of these covariates. In the following section, we additionally provide a causal sensitivity analysis which shows that, even in the presence of unobserved confoudning, our results remain robust.  

\newpage

\newpage 

\section{Sensitivity analysis for unobserved confounding}
\label{supp:causal_sensitivity} 

We perform a sensitivity analysis of our results with respect to possible violations of the ignorability assumption. Thereby, we demonstrate that our results are robust to a certain level of unobserved confounding. To do so, we leverage recent advances in sensitivity analysis and use the method from Jesson et al.~\cite{Jesson.2022}, which is state-of-the-art for continuous treatment. Jesson et al. use a continuous marginal sensitivity model (CMSM) that relaxes the ignorability assumption and then derive bounds around the treatment--response curve. For potential unobserved confounders $U$, the CMSM \cite{Jesson.2022} assumes
\begin{equation}\label{eq:ratio_bound_cmsm}
\frac{1}{\Gamma} \leq \frac{f(a \mid x, u)}{f(a \mid x)}  \leq \Gamma,
\end{equation}
where $f(a \mid x, u)$ and $f(a \mid x)$ denote the conditional densities of the random variable $A$ given $X=x$, $U=u$, or $X=x$, respectively. Here, the sensitivity parameter $\Gamma$ controls the strength of allowed unobserved confounding. 

\begin{figure}[H]
 \centering
 \includegraphics[width=0.9\linewidth]{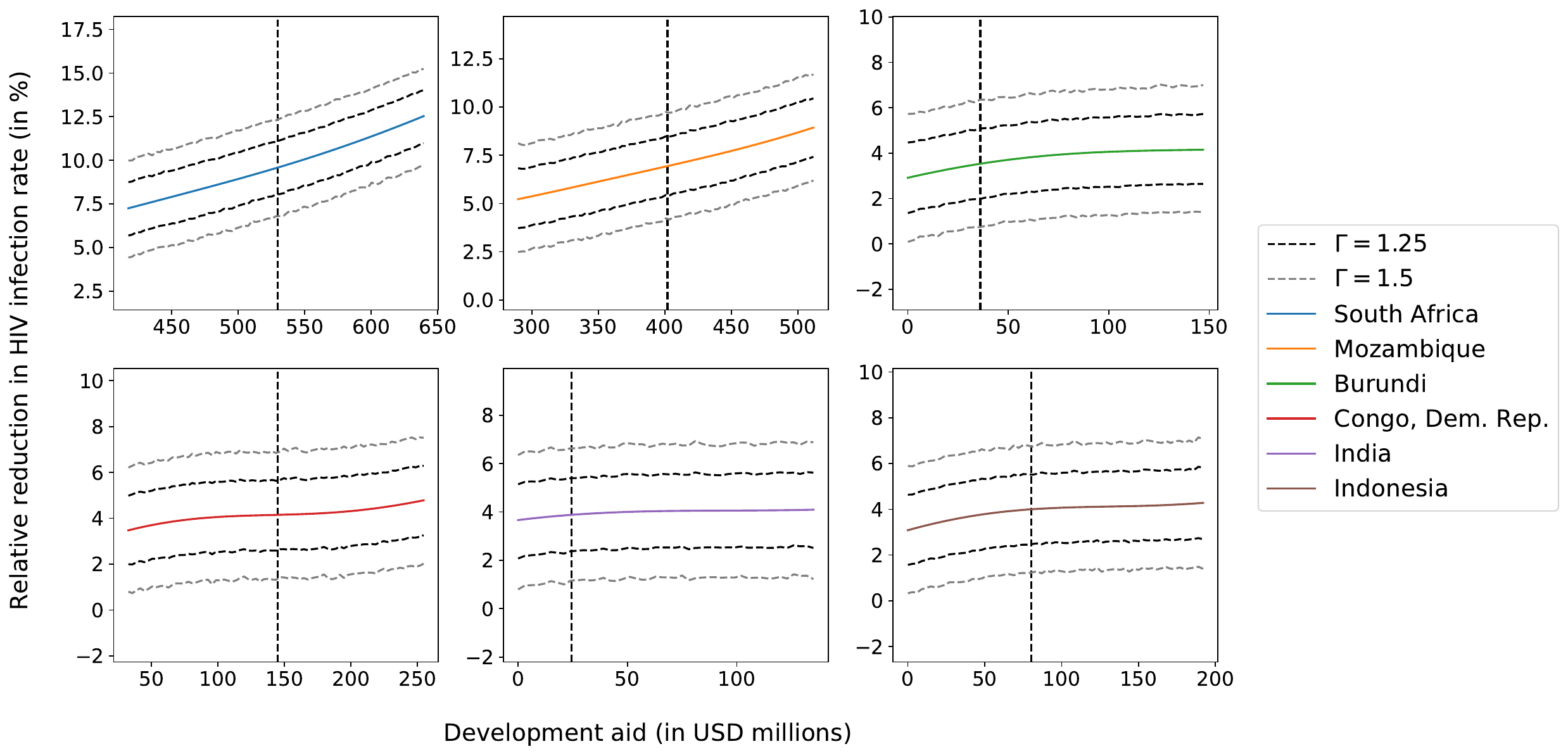}
\caption{Bounds on predicted treatment--response curves for the six example countries as in the results of the main paper: South Africa, Mozambique, Burundi, Congo, India, and Indonesia. Vertical dashed line denotes the actual volume of development aid as observed in 2017. The $x$-axis is set to $A_\text{obs}$ $\pm$ $\hat{\sigma}_{A}$ (with a cut-off at zero to prevent negative values for Burundi and India), where $A_\text{obs}$ is the observed development aid and $\hat{\sigma}_{A}$ is the estimated standard deviation of development aid in 2017. Shown are the predicted treatment--response curve (colored line) and sensitivity bounds for two levels of the strength of allowed unobserved confounding $\Gamma$ (dashed black lines). }
\label{fig:sensitivity_causal}
\end{figure}

To apply the method from \cite{Jesson.2022}, we need to predict the treatment--response curve $\mathbb{E}[Y \mid A = a, X = x]$ and the conditional outcome distribution $p(Y = y \mid A = a, X = x)$, which we model as a normal distribution with mean equal to $\mathbb{E}[Y \mid A = a, X = x]$ and constant variance. We use \framework to predict the treatment--response curve. We use different values of $\Gamma$ that are informed by domain knowledge in that already several of the observed country covariates such as GDP, population size, etc. are known, important drivers of HIV infection rates. The results are shown in Fig.~\ref{fig:sensitivity_causal} and confirm that our predicted treatment--response curves remain stable under small violations of the ignorability assumption. Crucially, unobserved confounding even up to $\Gamma = 1.25$ cannot explain away our predicted treatment effect. 

\section{Sensitivity analysis for hyperparameters}
\label{supp:sensitivity_hyperparameters} 

\begin{figure}[H]

\begin{minipage}[b]{0.5\textwidth}

{\raggedright\figletter{a}}
\vspace{0.1cm}

\centering
    \includegraphics[width=0.8\textwidth]{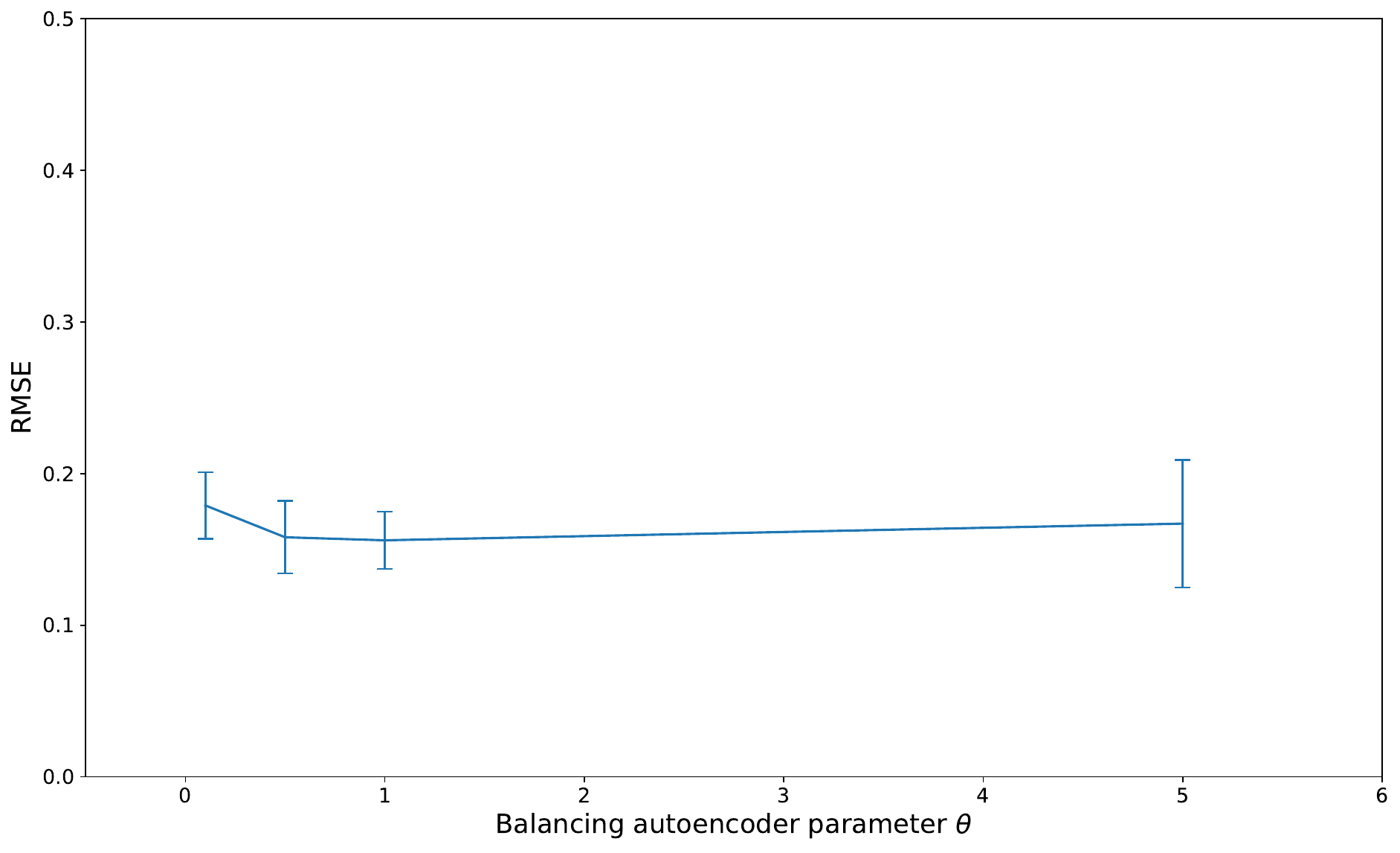}
    \label{fig:sen_theta}
\end{minipage}

\begin{minipage}[b]{0.5\textwidth}
 
{\raggedright\figletter{b}}
\vspace{0.1cm}

\centering
    \includegraphics[width=0.8\textwidth]{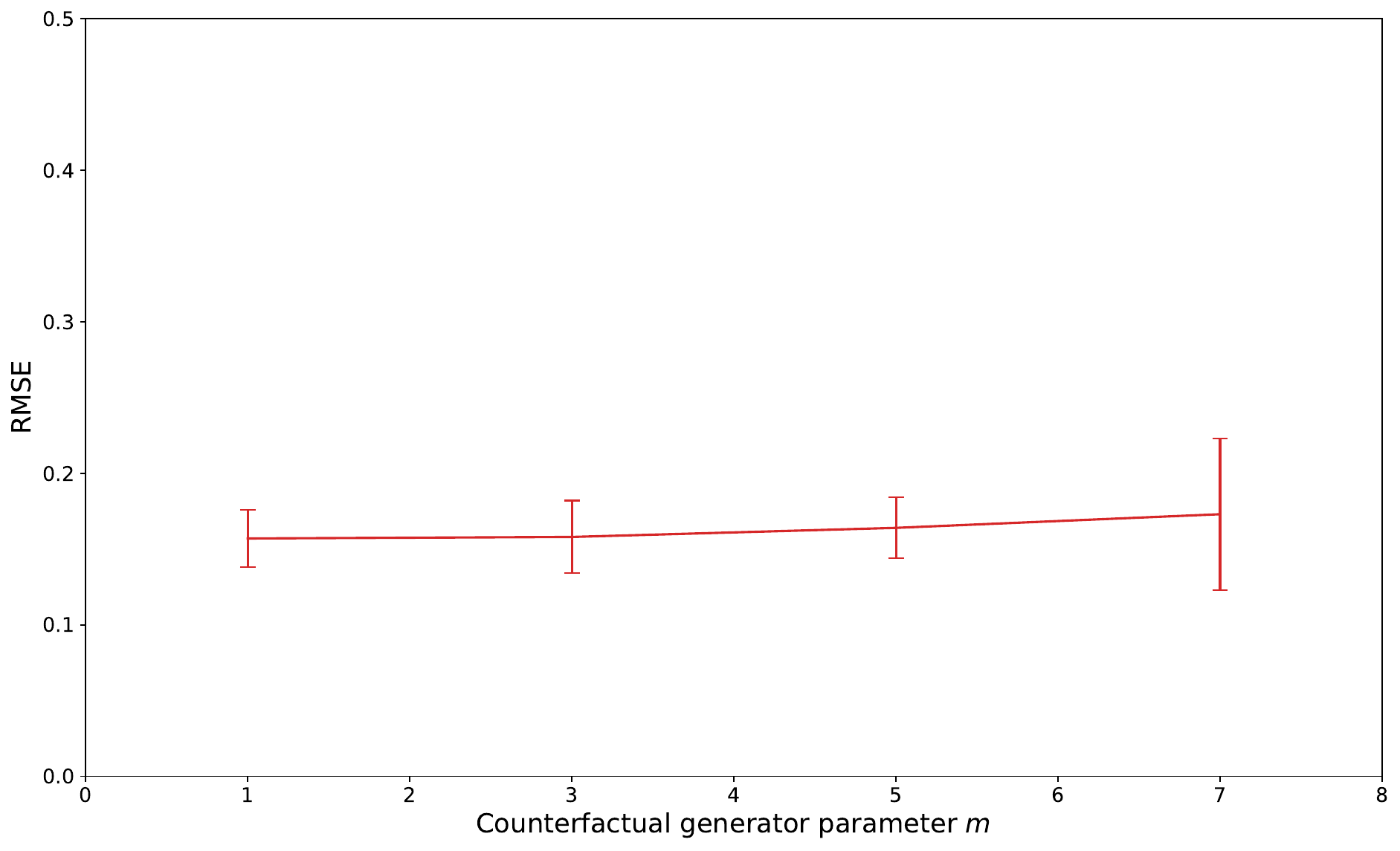}
    \label{fig:sen_m}
\end{minipage}
\caption{Sensitivity analysis to different hyperparameters. We visualize the performance of our \framework in predicting treatment--response curves using semi-synthetic data when varying the following two hyperparameters: in \figletter{a}~the balancing parameter $\theta$, and in \figletter{b}~the number of counterfactual outcomes generated per data point $m$. We observe that the performance remains robust with respect to these two hyperparameters. Whiskers show the standard deviation estimates over 10 runs.}
\label{fig:sensitivity}
\end{figure}

As a sensitivity analysis, we evaluate the performance of our \framework in predicting treatment--response curves when varying hyperparameters. Here, our intention is to show that the performance remains at a similar level across various choices, which thereby adds to the robustness of our proposed method. Specifically, we vary two hyperparameters: the balancing parameter $\theta$ (in Figure~\ref{fig:sensitivity}a), and the number of generated counterfactual outcomes $m$ (in Figure~\ref{fig:sensitivity}b). The results show that the performance of our \framework remains robust with respect to both hyperparameters.

\newpage

\section{Sensitivity analysis for \framework components}
\label{supp:ablation} 

We now assess the contribution of different components in our \framework to the overall performance. To this end, we perform experiments (i)~where we use other inference models, (ii)~where we toggle the balancing autoencoder and the counterfactual generator on/off, and (iii)~where we repeat the previous ablation study with other inference models. Hyperparameter tuning was done following the above for all experiments to allow for a fair comparison. The results are shown in Table~\ref{tab:ablation}. We discuss the findings in the following. 

For (i), we vary the inference models as follows: dose-response-network (DRNet) \cite{Schwab2020}, artificial neural network (ANN), linear model (LM), and generalized propensity score (GPS) \cite{Hirano2004}. The latter is used in our main analysis as it is the default in our \framework. The rationale was that it has a fairly parsimonious structure, which reduces the risk of overfitting while it offers the flexibility to handle various degrees of nonlinearities. We find that our \framework also works with other inference models. In fact, it consistently outperforms the baselines from the main paper, regardless of the underlying inference model. This thus confirms that our methodological innovations achieve consistent performance gains. Moreover, we find that our preferred choice for the inference model (GPS) as implemented in \framework is beneficial: it offers a superior performance with comparatively small variance. 

For (ii), we toggle the balancing autoencoder and the counterfactual generator on/off. Here, we find that the performance gains are a result of combining both the balancing autoencoder and the counterfactual generator (as implemented in \framework). We also performed ablation studies with both components separately (i.e., only the balancing autoencoder or only the counterfactual generator). However, the experiments with just one of the two components did not lead to performance improvements in our \framework. Hence, the combination of \emph{both} the balancing autoencoder \emph{and} the counterfactual generator is important for achieving a state-of-the-art performance. 

For (iii), we repeat the above ablation studies with other inference models. Here, we confirm that performance gains are achieved when combining \emph{both} the balancing autoencoder \emph{and} the counterfactual generator (whereas having only one of the two does not lead to performance improvements). Therefore, we find that our methodological innovations (i.e., the combination of both the balancing autoencoder and the counterfactual generator) leads to consistent improvements across all inference models. This corroborates the robustness of our contributions and shows the value of our methodological innovation for other inference methods. 

\begin{table}[H]
\begin{center}
\caption{Sensitivity analysis for model components. We show the results of experiments with semi-synthetic data with different inference models and with different model components. We report square root of the mean integrated squared error (mean $\pm$ standard deviation averaged over 10 runs). Specifically, we vary the model components as follows: (i)~without the balancing autoencoder and without the counterfactual generator (Base); (ii)~only with the balancing autoencoder (BAE); (iii)~only with the counterfactual generator (CF-GEN); and (iv)~with both the balancing autoencoder and the counterfactual generator (\framework).  }
\label{tab:ablation}
\resizebox{0.45\textwidth}{!}{
\footnotesize
\begin{tabular}{l  c c c c }
\hline \hline
 &  \multicolumn{4}{c}{\textbf{Inference models}} \\
\cline{2-5}
 & DRNet & ANN &  LM & GPS\\
\hline
Base & $0.281 \pm 0.043$  & $0.234 \pm 0.030$   &    $0.209 \pm 0.000 $ & $0.205 \pm 0.000 $  \\
\hline
BAE & $0.303\pm 0.062$  & $0.319 \pm 0.105$   &    $0.192 \pm 0.017 $ & $0.214 \pm 0.009 $  \\
\hline
CF-GEN & $0.230\pm 0.032$  & $0.176 \pm 0.021$   &    $0.246 \pm 0.003 $ & $0.210 \pm 0.003 $  \\
\hline
\framework & $\mathbf{0.181 \pm 0.048}$ & $\mathbf{0.173 \pm 0.018}$   &   $\mathbf{0.159 \pm 0.042}$ &  $\mathbf{0.158 \pm 0.024}$ \\

\hline \hline
\multicolumn{5}{l}{Lower = better}
\end{tabular}
}
\end{center}
\end{table}

\section{Experiments with other time frames}
\label{supp:hiv2016} 

In our main analysis, we use HIV data from the year 2016 for learning, and the data from the year 2017 for evaluation. Here, we show robustness of our results in a different time frame, where we use HIV data from the year 2015 for learning and data from the year 2016 for evaluation. We run experiments with both semi-synthetic and real-world data, with the same baselines, performance metrics, and semi-synthetic data generation procedure as in our main analysis. Our results (in Table~\ref{tab:res_time}) show that our conclusions from the main analysis remain unchanged, i.e., our \framework offers state-of-the-art performance in experiments with both semi-synthetic and real-world data. 

Reassuringly, we remind that we also experimented with multi-year data for learning; however, this did not led to significant out-of-sample performance improvements. Here, one explanation is that allocation practices are subject to changes over time (e.g., changes in priorities, changes in programs, funding mix) \cite{Avila2013}.  
\vspace{-0.25cm}
\begin{table}[H]
\begin{center}
\caption{{Results of experiments with other time frames.} We show the results of experiments with semi-synthetic and real-world data when using the year 2015 for learning, and the year 2016 for evaluation. For experiments with semi-synthetic data, we report the square root of the mean integrated squared error, and for experiments with real-world data, we report the root mean squared error (mean $\pm$ standard deviation averaged over 10 runs). }
\label{tab:res_time}
\vspace{-0.3cm}
\resizebox{0.45\textwidth}{!}{
\begin{tabular}{l  c c c c c c }
\hline \hline
\multirow{2}{*}{\textbf{Data}} &  \multicolumn{6}{c}{\textbf{Method}} \\
\cline{2-7} 
 & SCIGAN  & DRNet & GPS & ANN & LM & \framework \\
\hline
\textbf{Semi-synthetic}  & $8.425 \pm 3.629$ & $0.255 \pm 0.046$  & $0.187 \pm 0.000$ & $0.221 \pm 0.042$  &  $0.237 \pm 0.000$  & $\mathbf{0.173 \pm 0.023}$    \\
\hline
\textbf{Real-world} & $2.397 \pm 2.336$ & $0.109 \pm 0.019$  & $0.090 \pm 0.000$ & $0.098 \pm 0.010$ &  $0.087 \pm 0.000$  & $\mathbf{0.087 \pm 0.002}$    \\
\hline \hline
\multicolumn{7}{l}{Lower = better}
\end{tabular}
}
\end{center}
\end{table}

\vspace{-0.6cm}

\newpage

\section{Predicted aid--response curves for different countries in 2017}
\label{supp:country_plots} 

Here, we show the predicted aid--response curves for all remaining countries that are not shown in the main paper. The vertical dashed line denotes the actual volume of development aid as observed in 2017. The $x$-axis is set to $A_\text{obs}$ $\pm$ $\hat{\sigma}_{A}$ (with a cut-off at zero to prevent negative values), where $A_\text{obs}$ is the observed development aid and $\hat{\sigma}_{A}$ is the estimated standard deviation of development aid in 2017. Shown are both point predictions (thick line) and the standard deviation over 10 runs (shaded area).

\begin{figure}[H]
\centering
\includegraphics[width=0.43\textwidth]{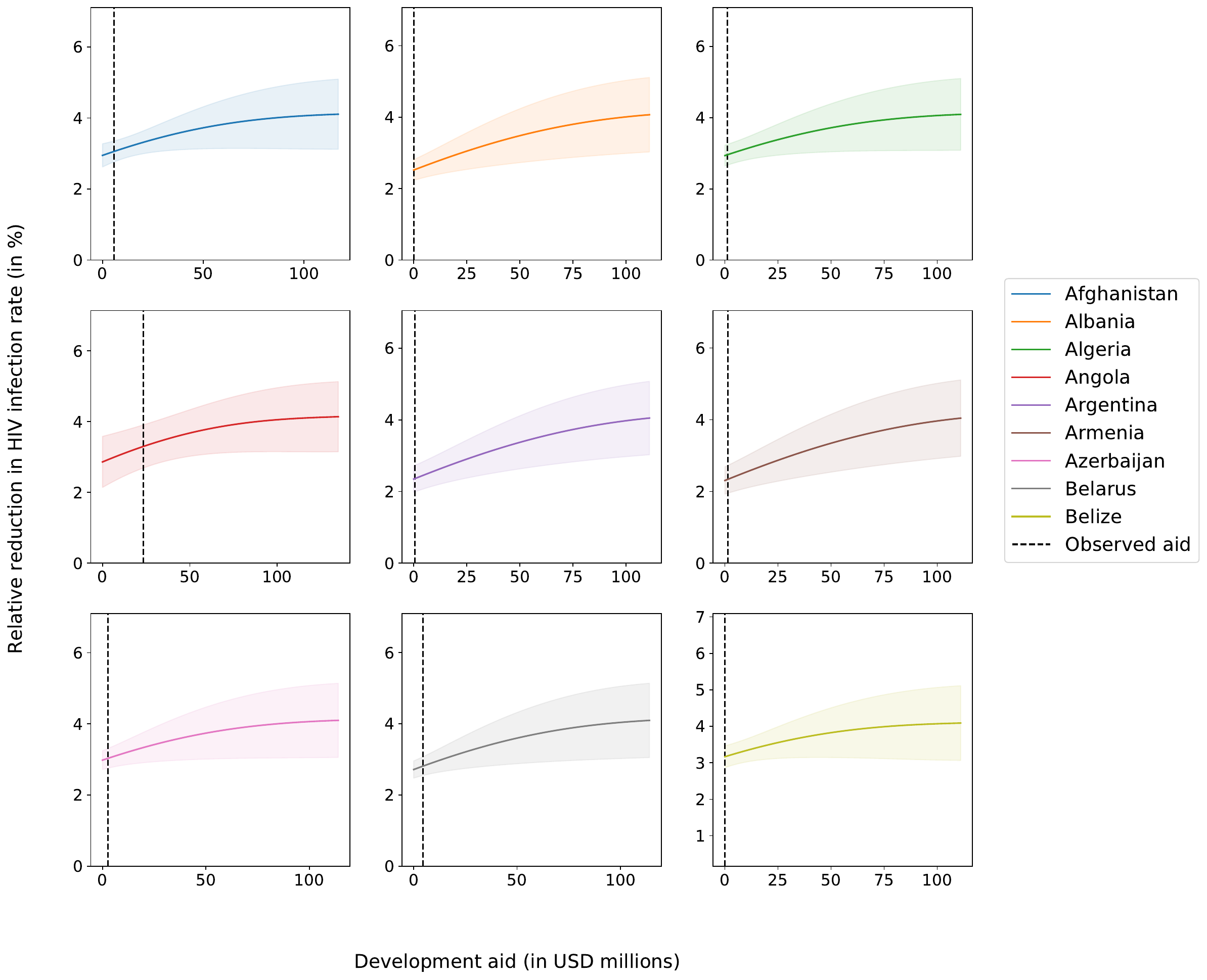}
\label{fig:results0}
\end{figure}

\begin{figure}[H]
\centering
\includegraphics[width=0.43\textwidth]{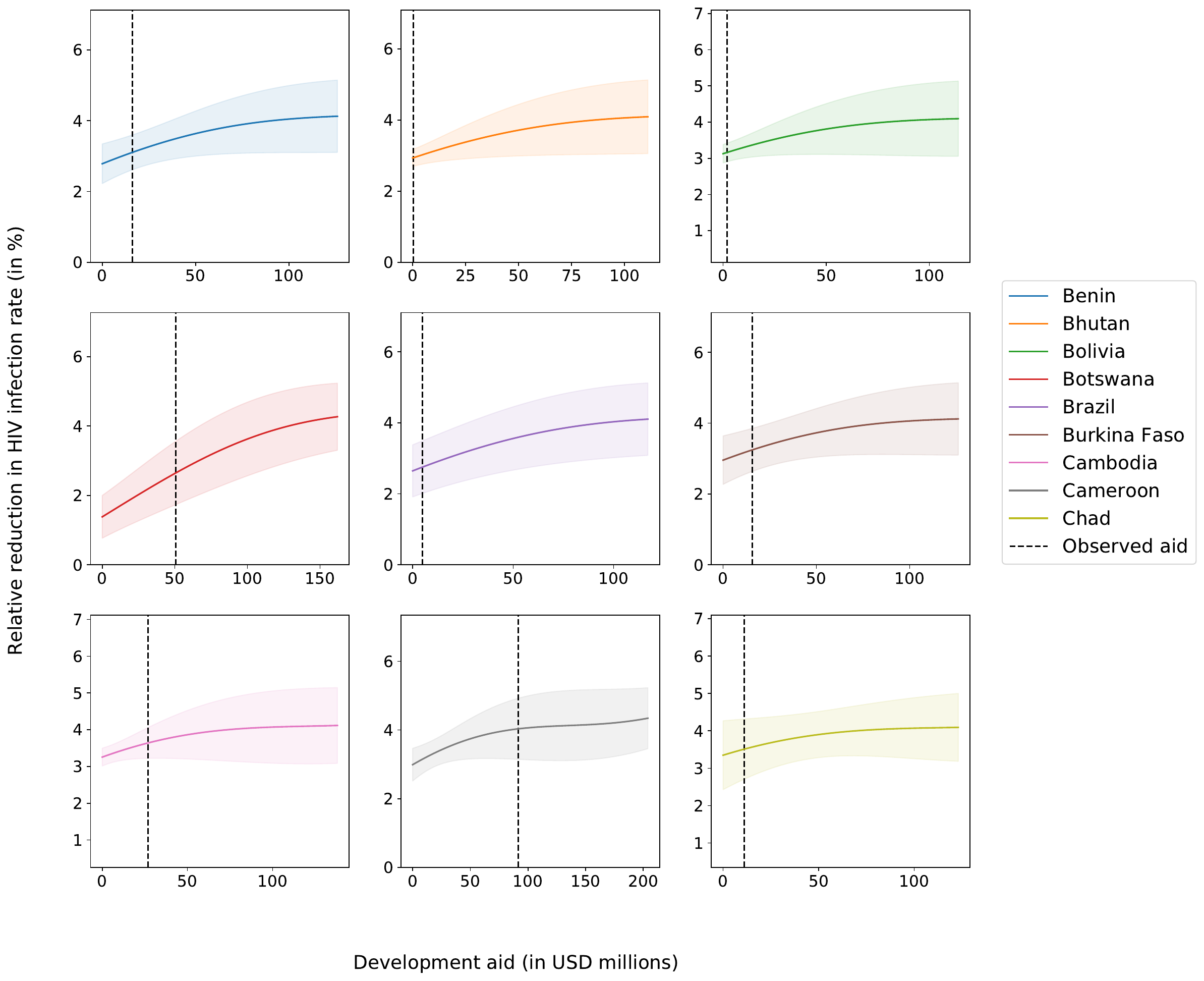}
\label{fig:results1}
\end{figure}

\begin{figure}[H]
\centering
\includegraphics[width=0.43\textwidth]{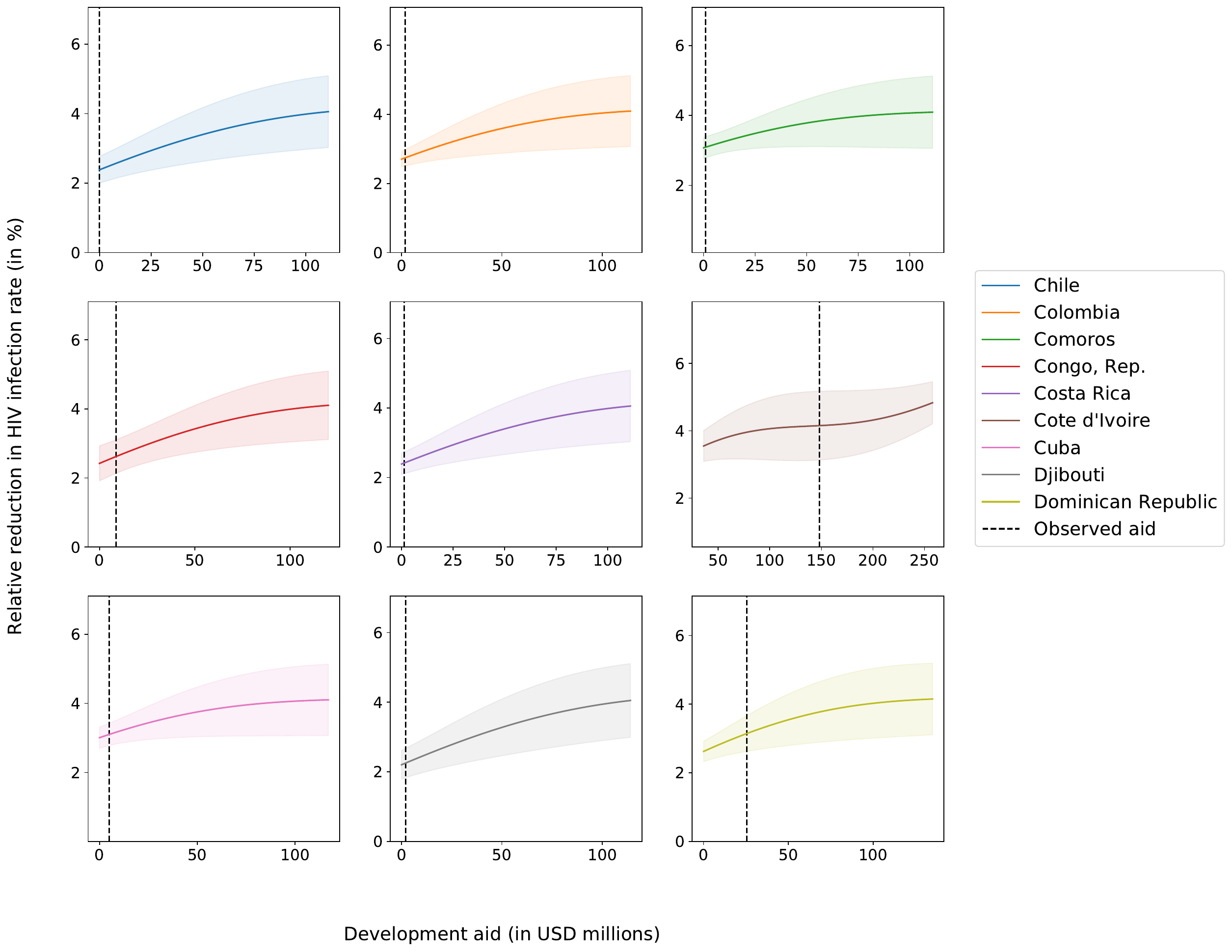}
\label{fig:results2}
\end{figure}

\begin{figure}[H]
\centering
\includegraphics[width=0.43\textwidth]{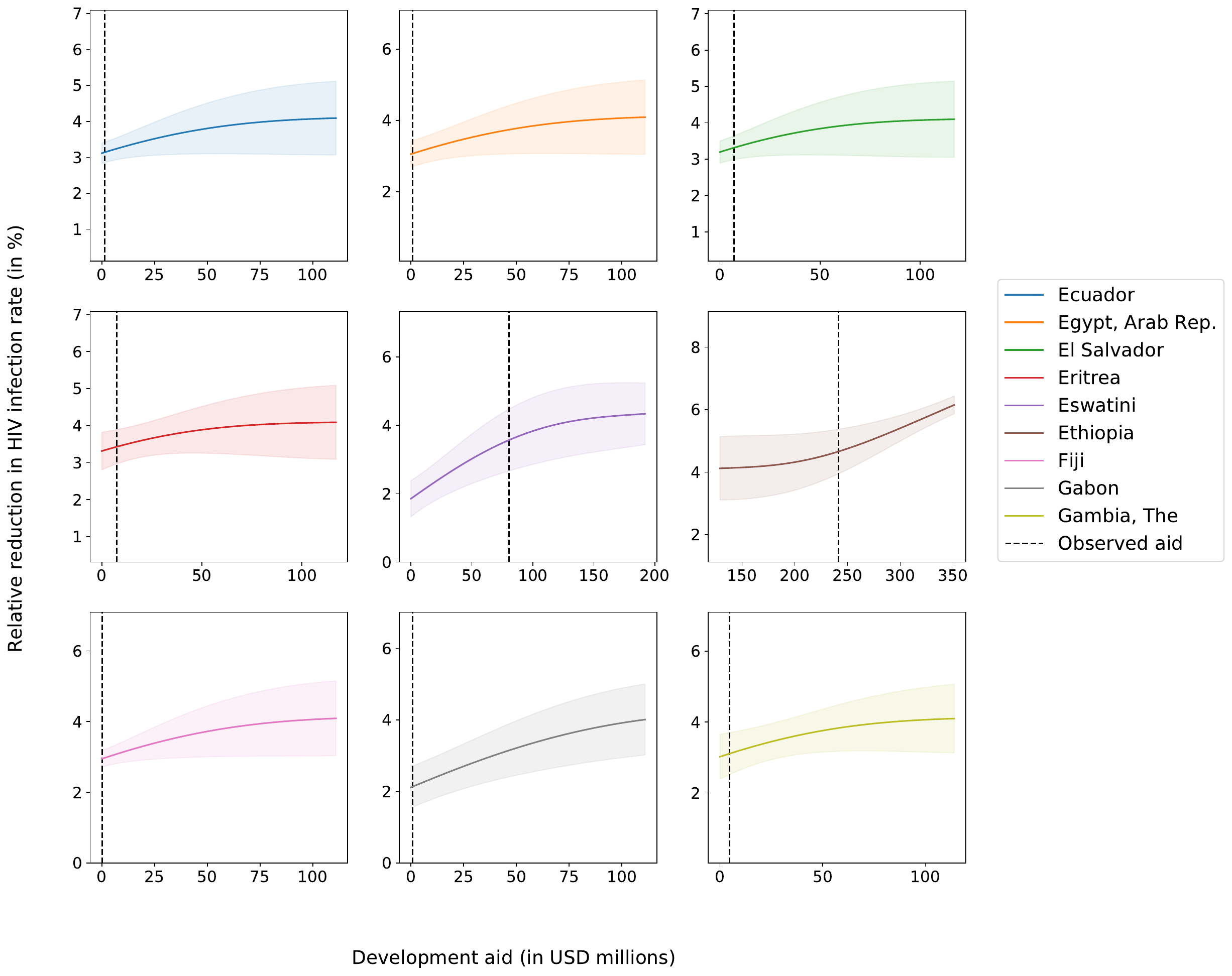}
\label{fig:results3}
\end{figure}

\begin{figure}[H]
\centering
\includegraphics[width=0.43\textwidth]{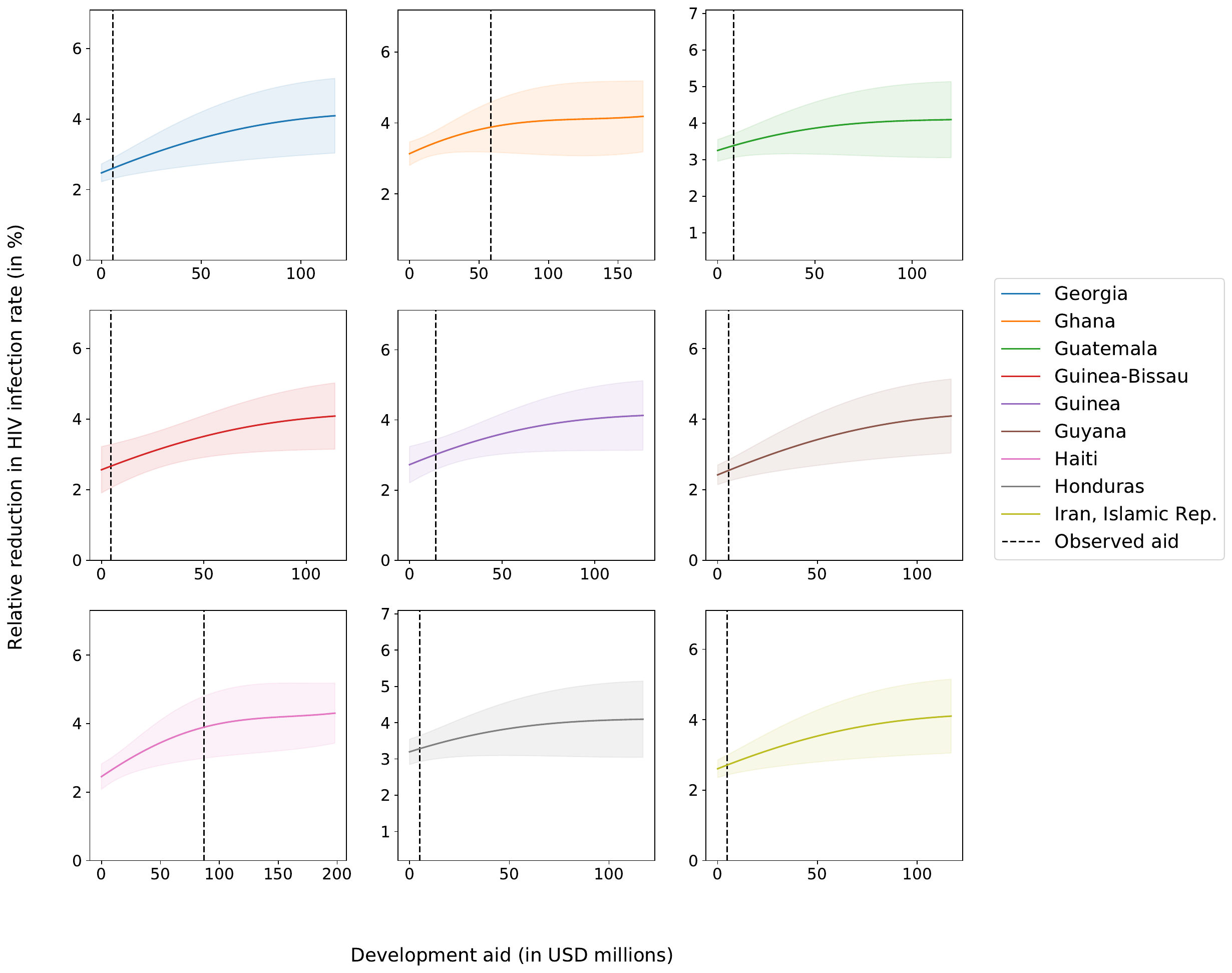}
\label{fig:results4}
\end{figure}

\begin{figure}[H]
\centering
\includegraphics[width=0.43\textwidth]{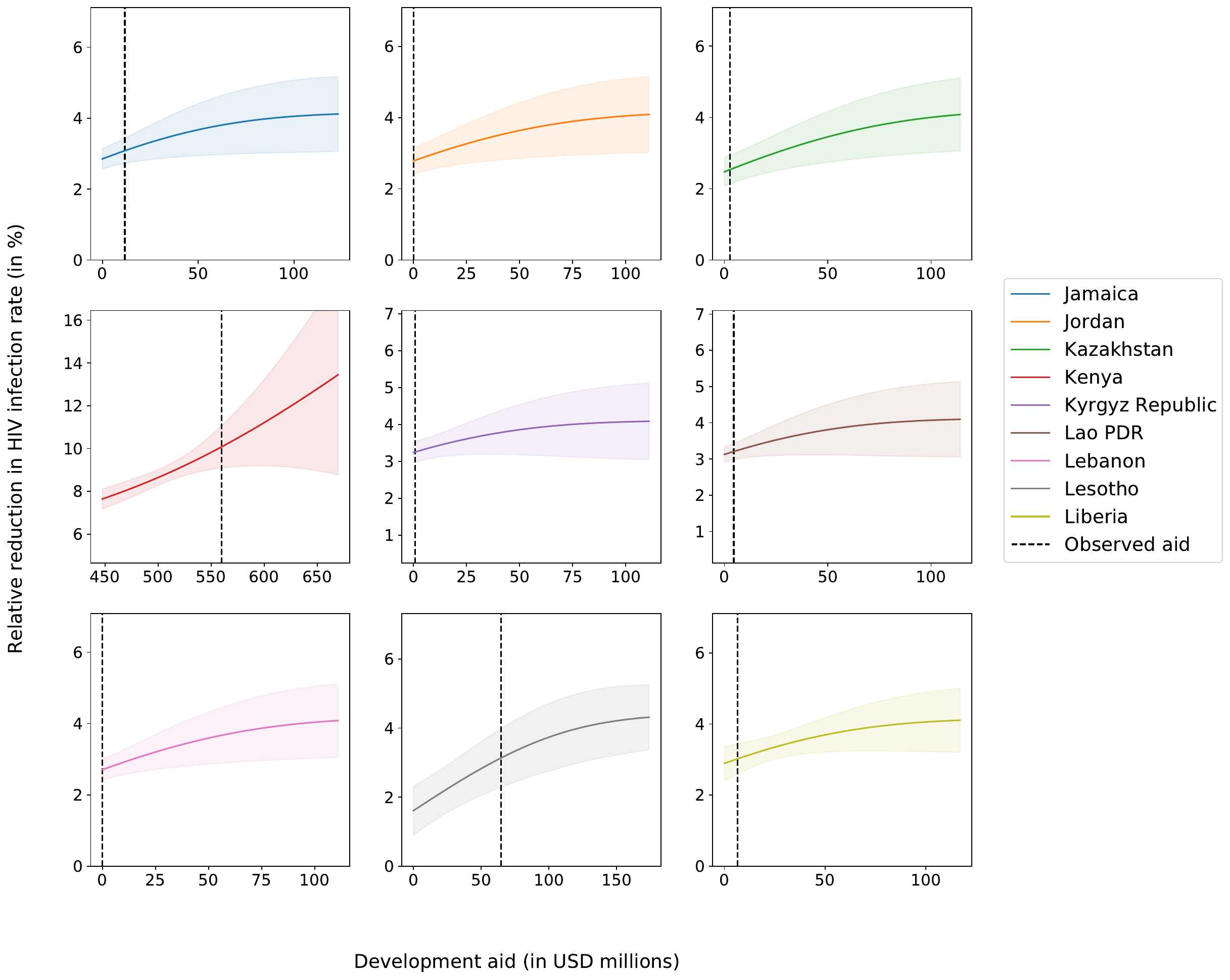}
\label{fig:results5}
\end{figure}

\begin{figure}[H]
\centering
\includegraphics[width=0.43\textwidth]{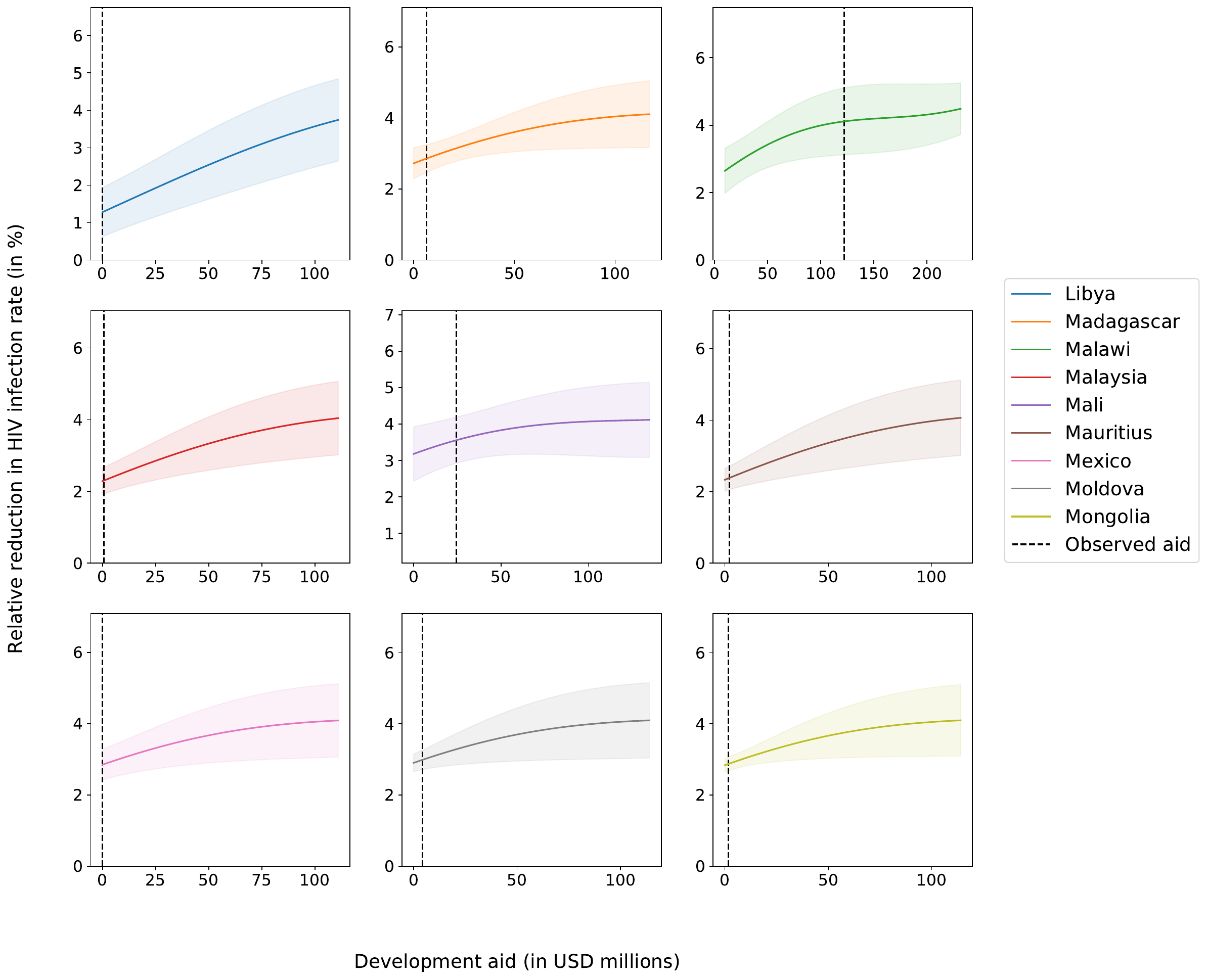}
\label{fig:results6}
\end{figure}

\begin{figure}[H]
\centering
\includegraphics[width=0.43\textwidth]{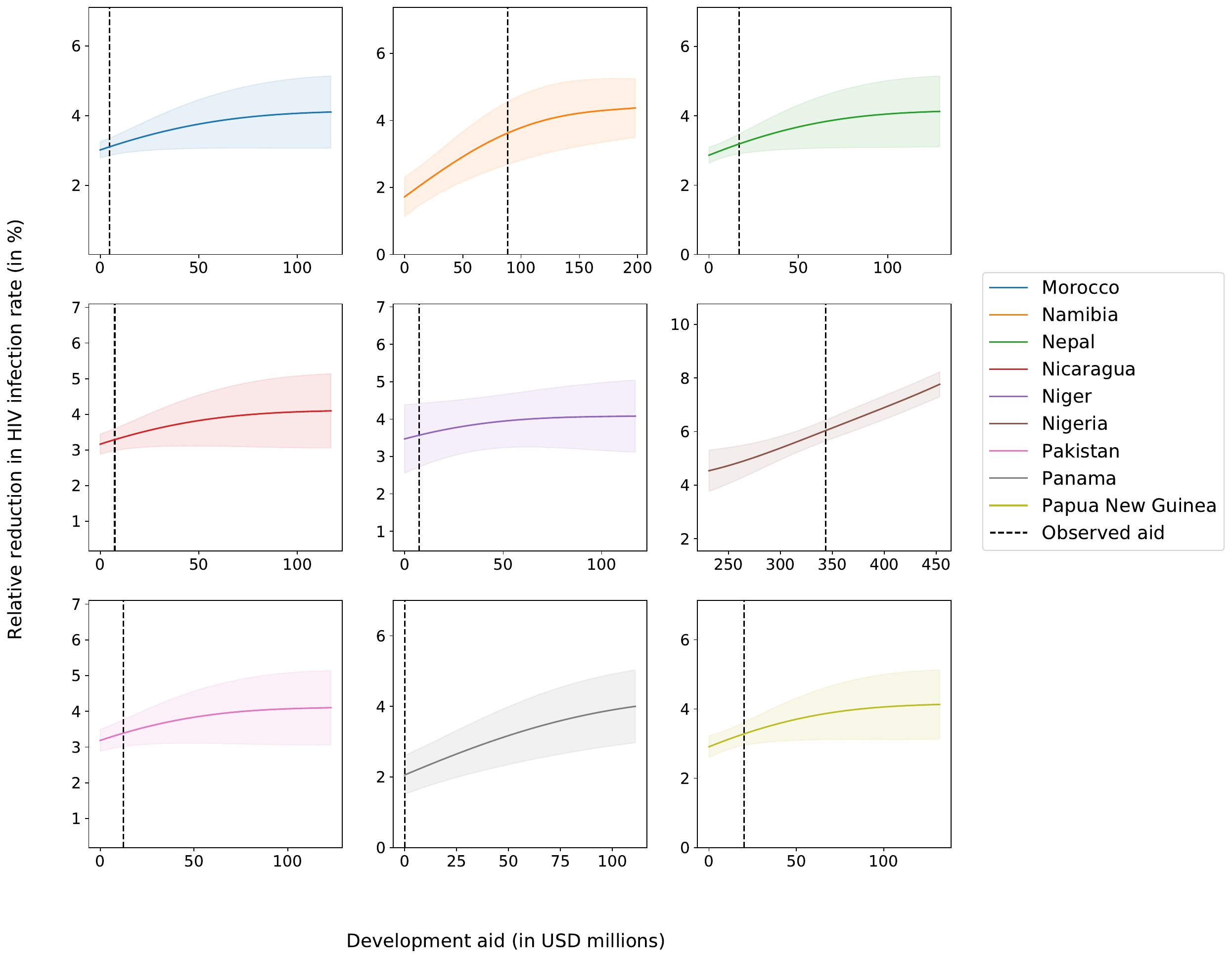}
\label{fig:results7}
\end{figure}

\begin{figure}[H]
\centering
\includegraphics[width=0.43\textwidth]{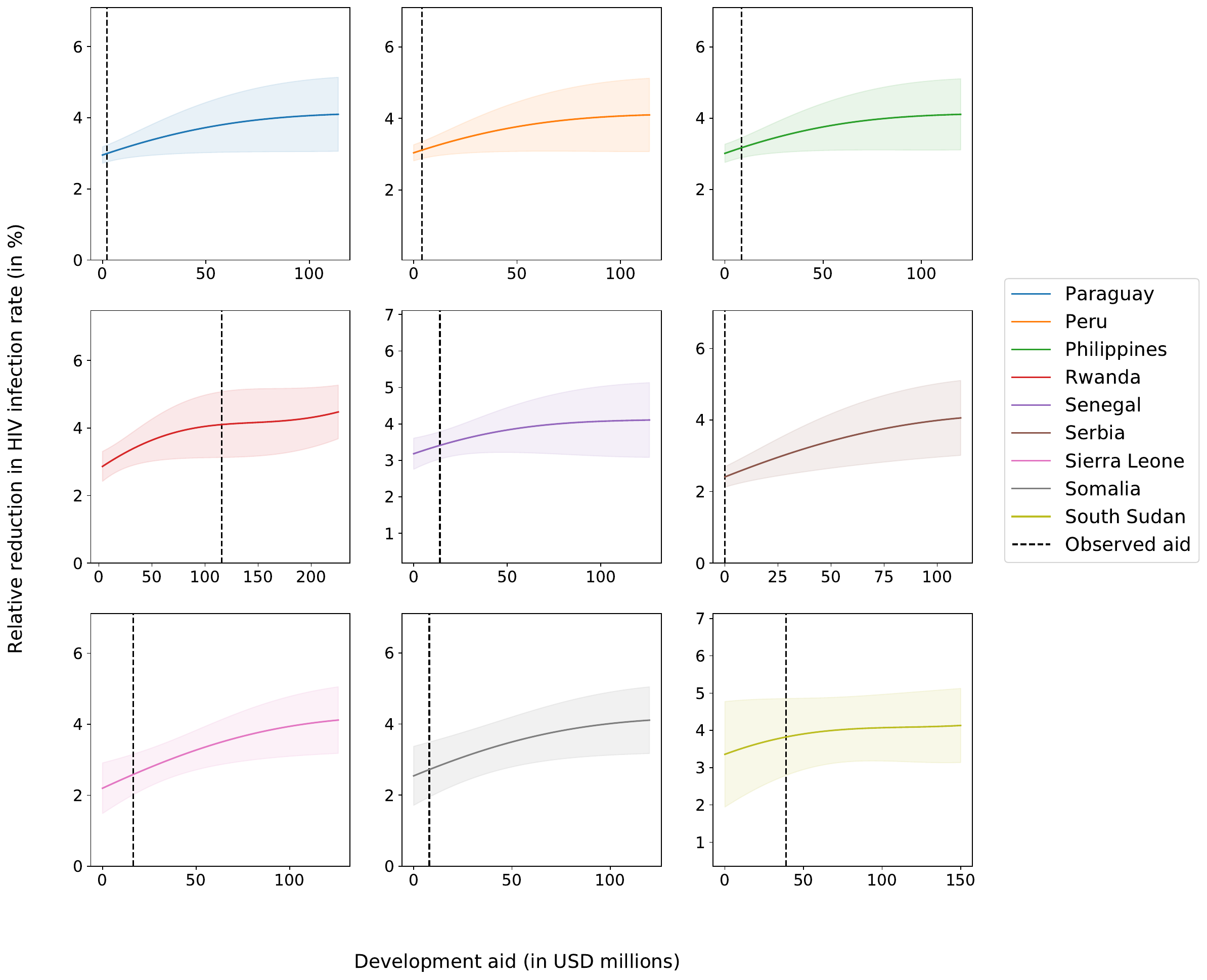}
\label{fig:results8}
\end{figure}

\begin{figure}[H]
\centering
\includegraphics[width=0.43\textwidth]{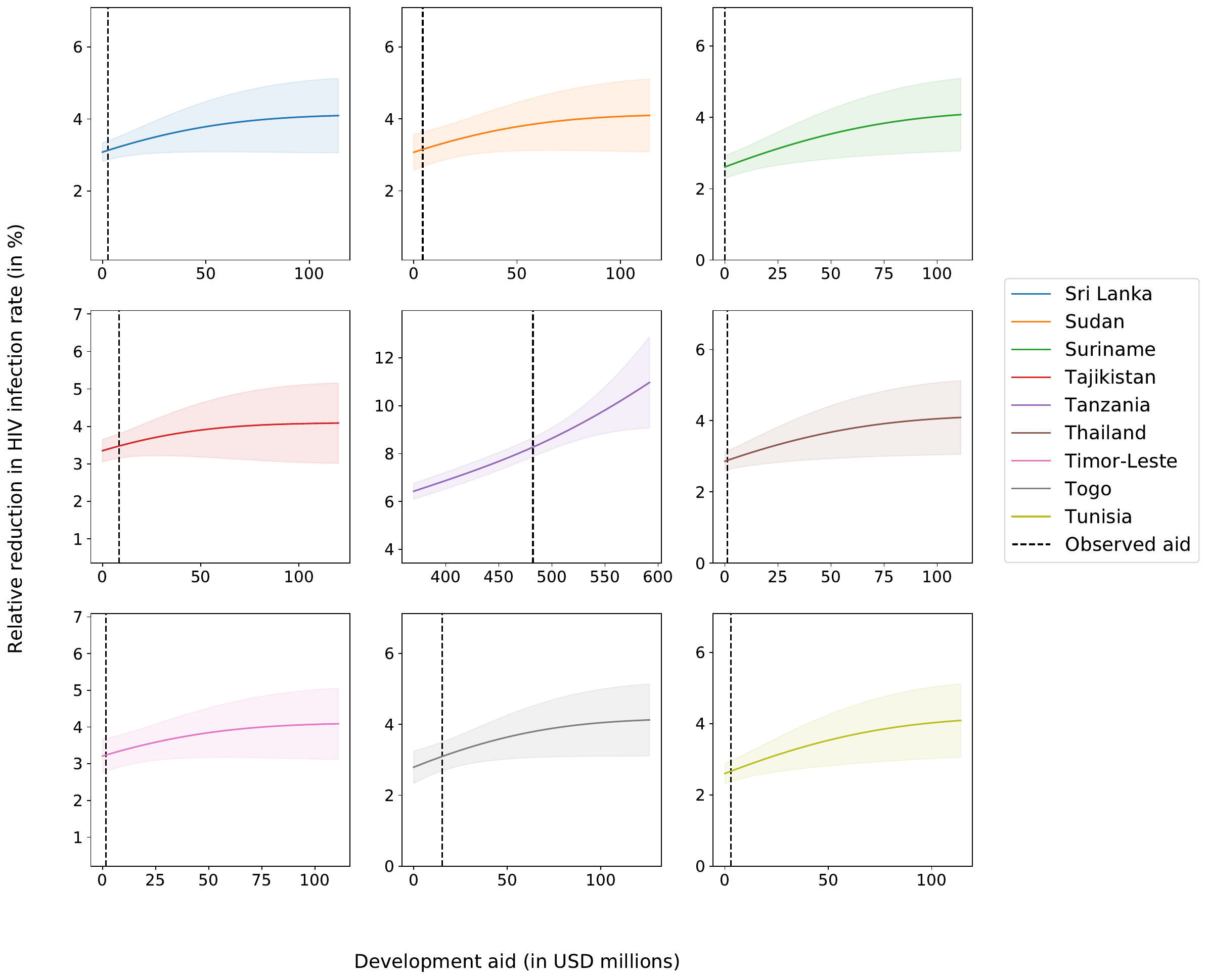}
\label{fig:results9}
\end{figure}

\begin{figure}[H]
\centering
\includegraphics[width=0.43\textwidth]{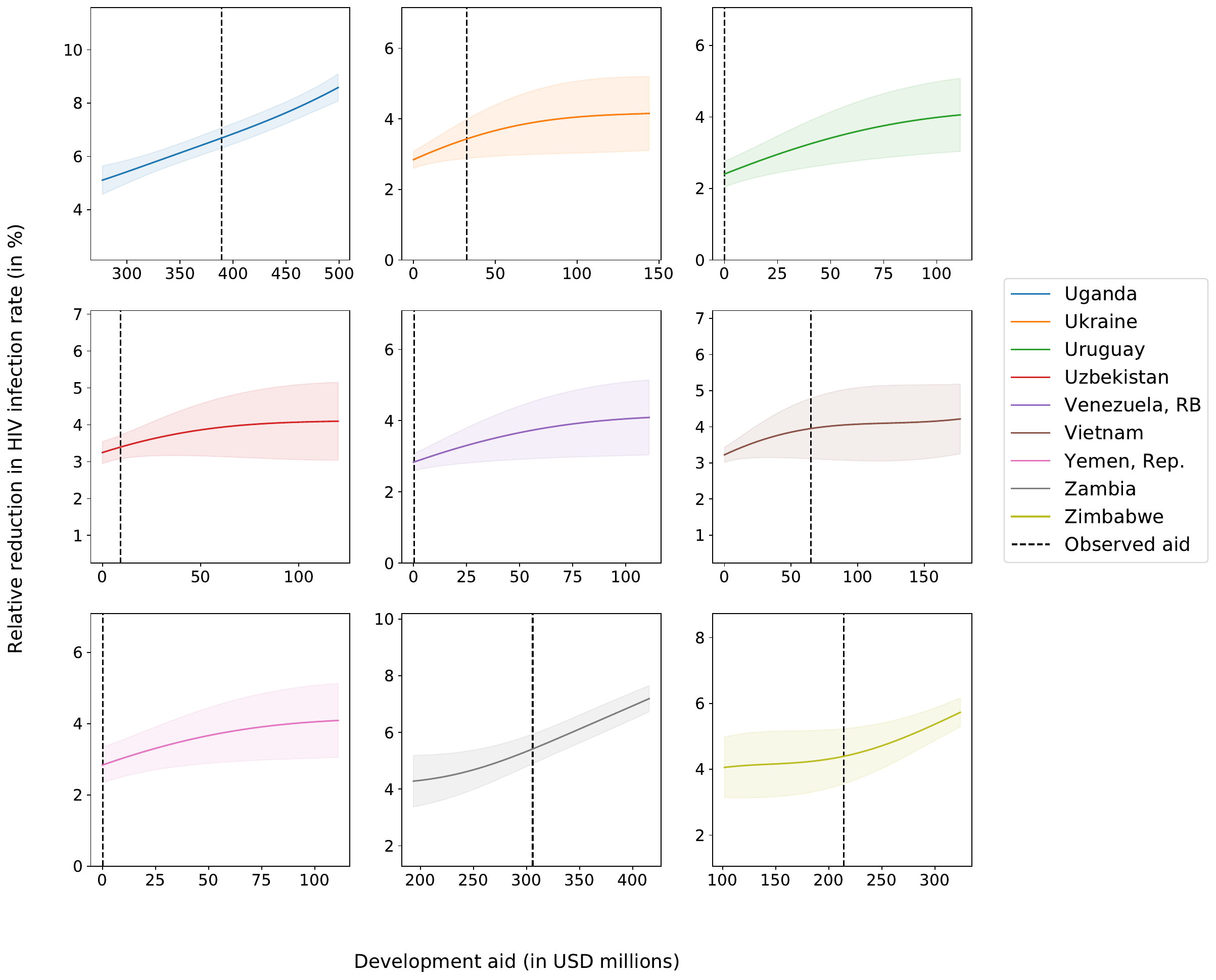}
\label{fig:results10}
\end{figure}

\end{document}